\documentclass[10pt,twocolumn,letterpaper]{article}

\usepackage[pagenumbers]{cvpr} 

\DeclareMathOperator*{\argmax}{arg\,max}
\DeclareMathOperator*{\softmax}{softmax}

\definecolor{cvprblue}{rgb}{0.21,0.49,0.74}
\usepackage[pagebackref,breaklinks,colorlinks,allcolors=cvprblue]{hyperref}

\usepackage[accsupp]{axessibility}

\usepackage{url}
\usepackage{hyperref}

\usepackage{multirow} 
\usepackage{makecell} 

\usepackage{adjustbox}
\usepackage{amsmath}

\usepackage{graphicx} 

\usepackage{colortbl}
\usepackage{xcolor}
\usepackage{color}
\usepackage{array}   

\usepackage{float}

\usepackage{nicematrix}  
\usepackage{tabularx}

\usepackage{algorithmic}
\usepackage{algorithm}
\usepackage{caption}

\usepackage{fontawesome}

\usepackage{pifont}

\usepackage{tocloft}
\usepackage{etoc}

\usepackage[most]{tcolorbox} 

\usepackage{scrextend}

\definecolor{removed}{RGB}{142, 8, 7}

\definecolor{method_orange}{RGB}{211, 91, 28}
\definecolor{method_blue}{RGB}{67, 128, 186} 
\definecolor{method_green}{RGB}{96, 130, 66}
\definecolor{method_purple}{RGB}{183, 108, 209}

\definecolor{table_orange}{RGB}{253, 243, 239}
\definecolor{table_blue}{RGB}{239, 245, 251}
\definecolor{table_green}{RGB}{243, 250, 238}

\definecolor{fig_purple}{RGB}{153, 120, 191}

\definecolor{uclagold}{RGB}{232,237,205}

\definecolor{grayred}{RGB}{254,180,167}

\definecolor{prompt_blue}{RGB}{86, 143, 193}
\definecolor{prompt_green}{RGB}{142, 167, 119}

\definecolor{speed_up_color}{RGB}{211, 91, 28}

\definecolor{highlight_result}{RGB}{206, 206, 250}
\newcommand{\highlight}[1]{{\cellcolor[RGB]{206, 206, 250}}{#1}}

\newcommand{\HighTableOrange}[1]{{\cellcolor[RGB]{253, 243, 239}}{#1}}
\newcommand{\HighTableBlue}[1]{{\cellcolor[rgb]{0.925,0.957,1}}{#1}}

\newcommand{\HighTableGreen}[1]{{\cellcolor[RGB]{243, 250, 238}}{#1}}

\newtcolorbox{promptbox_blue}[1]{ 
  colback=prompt_blue!10!white,
  colframe=prompt_blue,
  boxrule=0.8pt,
  arc=4pt,
  left=6pt, right=6pt, top=6pt, bottom=6pt,
  fonttitle=\bfseries\large,
  coltitle=white,
  title=#1 
}

\newtcolorbox{promptbox_green}[1]{ 
  colback=prompt_green!10!white,
  colframe=prompt_green,
  boxrule=0.8pt,
  arc=4pt,
  left=6pt, right=6pt, top=6pt, bottom=6pt,
  fonttitle=\bfseries\large,
  coltitle=white,
  title=#1 
}

\def\paperID{269} 
\def\confName{CVPR}
\def\confYear{2026}

\title{From Scale to Speed: Adaptive Test-Time Scaling for Image Editing}

\author{%
XXX
}

\author{
\hspace{-0.9cm}
\textbf{Xiangyan Qu}\textsuperscript{12}$^{*}$\thanks{Work done during the internship at AMAP, Alibaba Group.},
\textbf{Zhenlong Yuan}\textsuperscript{3}$^{*}$,
\textbf{Jing Tang}\textsuperscript{3}\textsuperscript{$\ddagger$},
\textbf{Rui Chen}\textsuperscript{3}, 
\textbf{Datao Tang}\textsuperscript{3},
\textbf{Meng Yu}\textsuperscript{3},\\
\textbf{Lei Sun}\textsuperscript{3},
\textbf{Yancheng Bai}\textsuperscript{3},
\textbf{Xiangxiang Chu}\textsuperscript{3}, 
\textbf{Gaopeng Gou}\textsuperscript{12}\textsuperscript{\textsection}, 
\textbf{Gang Xiong}\textsuperscript{12}, 
\textbf{Yujun Cai}\textsuperscript{4},
\vspace{0.5em}
\\
{\small
\textsuperscript{1}Institute of Information Engineering, Chinese Academy of Sciences\quad
\textsuperscript{2}School of Cyber Security, University of Chinese Academy of Sciences\quad} \\
{
\small
\textsuperscript{3}AMAP, Alibaba Group\quad
\textsuperscript{4}University of Queensland
}\\
\vspace{0.5em}
{\small \textsuperscript{$*$} Equal contribution. \quad \textsuperscript{$\ddagger$} Project lead. \quad \textsuperscript{\textsection} Corresponding author.} \\
\vspace{-2em}
}

\begin{document}
\maketitle

\begin{abstract}
Image Chain-of-Thought (Image-CoT) is a test-time scaling paradigm that improves image generation by extending inference time. Most Image-CoT methods focus on text-to-image (T2I) generation. Unlike T2I generation, image editing is goal-directed: the solution space is constrained by the source image and instruction. This mismatch causes three challenges when applying Image-CoT to editing: inefficient resource allocation with fixed sampling budgets, unreliable early-stage verification using general MLLM scores, and redundant edited results from large-scale sampling. To address this, we propose \textbf{AD}aptive \textbf{E}dit-\textbf{CoT} (ADE-CoT), an on-demand test-time scaling framework to enhance editing efficiency and performance. It incorporates three key strategies: (1) a difficulty-aware resource allocation that assigns dynamic budgets based on estimated edit difficulty; (2) edit-specific verification in early pruning that uses region localization and caption consistency to select promising candidates; and (3) depth-first opportunistic stopping, guided by an instance-specific verifier, that terminates when intent-aligned results are found. Extensive experiments on three SOTA editing models (Step1X-Edit, BAGEL, FLUX.1 Kontext) across three benchmarks show that ADE-CoT achieves superior performance-efficiency trade-offs. With comparable sampling budgets, ADE-CoT obtains better performance with more than 2× speedup over Best-of-N.
\end{abstract}


\section{Introduction}
\label{sec:intro}

Recent image-editing methods have shown impressive progress by latent-level fusion between multimodal large language models (MLLMs) and diffusion decoders, such as Step1X-Edit~\cite{Step1X_Edit_arxiv2025}, FLUX.1 Kontext~\cite{FLUX_Kontext_arxiv2025}, BAGEL~\cite{Bagel_2025_arxiv}, and Qwen-Image~\cite{qwen_image_arxiv2025}. However, their performance remains challenging on complex edits, such as large pose changes, multi-object edits, or multi-turn edits. Image Chain-of-Thought (Image-CoT)~\cite{TTS_Baseline, Image_CoT, TTS_SANA}, a test-time scaling strategy, offers a promising approach to tackle this challenge. It provides a plug-and-play, training-free solution that extends the inference time to boost the generation quality.

%
Most Image-CoT studies focus on text-to-image (T2I) generation.
The standard approach samples multiple candidates via noise perturbations~\cite{TTS_Baseline, TTS_SANA} and uses Best-of-N (BoN) selection.
However, the computational cost scales linearly with the number of samples, making it challenging to generate higher-quality images under a limited inference budget~\cite{TTS_survey}.
To address this, some methods incorporate prompt-level intervention through rewriting or reflective updates~\cite{Reflection_DiT, GenRef_CoT_ICCV2025} to increase candidate diversity and image-text alignment. Other work involves path search and pruning during the generation process. These methods utilize MLLMs as verifiers~\cite {Video_TTS, TTS_Baseline, Image_CoT, TTS_VAR, UniGen_TTS, Video_TTS_beam_search} to score intermediate denoising states and select promising candidates. By pruning low-potential samples early, they reduce computational cost without degrading final image quality.


{
\begin{figure}[t]
\begin{center}
\includegraphics[width=0.95\linewidth]{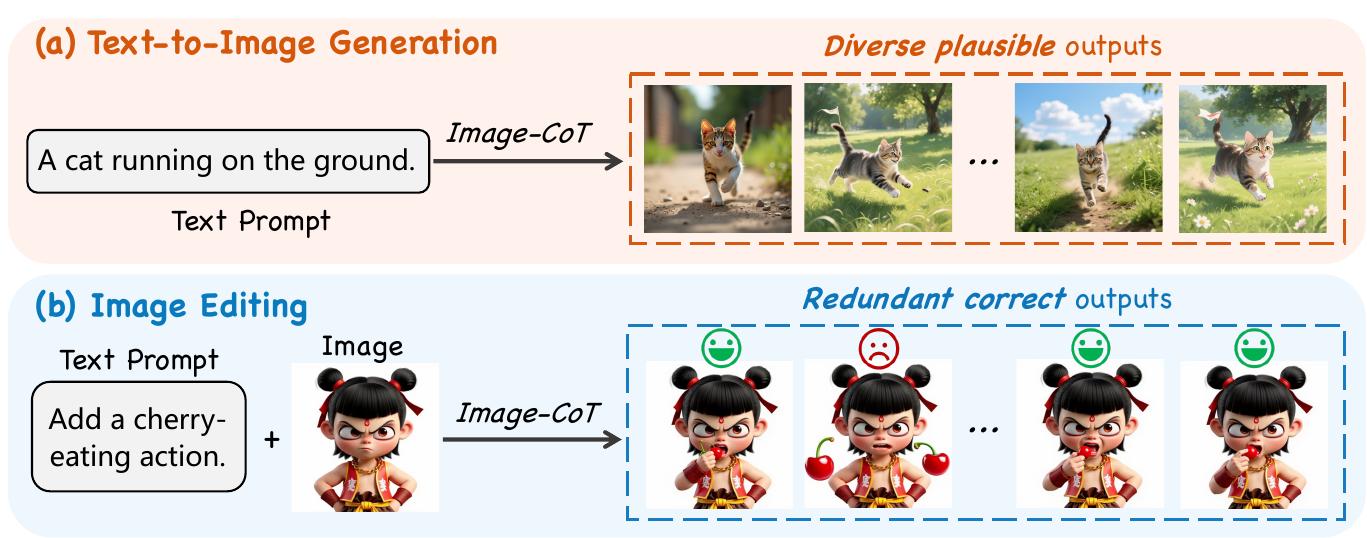}
\end{center}
\vspace{-1.7em}
\caption{\textbf{Impact of Image-CoT on different generative tasks}. 
\textbf{(a)} T2I generation is an open-ended task benefiting from large-scale sampling.
\textbf{(b)} Image editing is a goal-directed task where outputs are constrained by the prompt and source image, leading to redundant correct outputs after large-scale sampling.}
\vspace{-1.8em}
\label{fig: motivation_task_diff}
\end{figure}
}


{
\begin{figure*}[t]
\begin{center}
\includegraphics[width=0.99\linewidth]{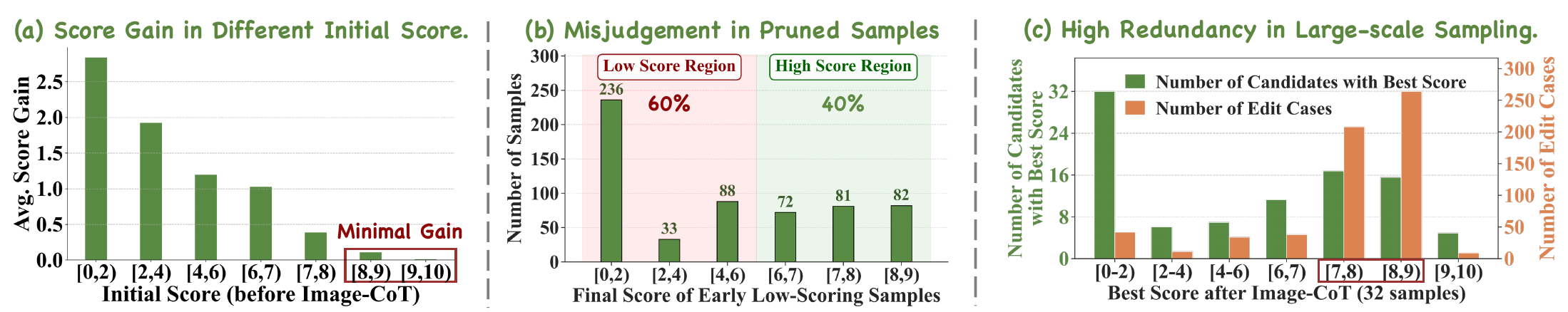}
\end{center}
\vspace{-1.6em}
\caption{\textbf{Why existing Image-CoT methods are suboptimal for editing.}
\textbf{(a) Inefficient resource allocation}: 
Fixed budgets waste computation on simple edits (high initial scores, \textbf{\textcolor{removed}{red box}}) that show minimal improvement. 
\textbf{(b) Unreliable early-stage verification}: 40\% of samples with low early scores achieve high final scores, but are pruned incorrectly by general MLLM scores. 
\textbf{(c) Redundant edited results}: 
Large-scale sampling produces redundant correct outputs with identical best scores, though only one is sufficient.}
\vspace{-1.6em}
\label{fig: motivation}
\end{figure*}
}


However, directly applying these T2I-centric Image-CoT methods to image editing is suboptimal due to fundamental differences between the tasks.
T2I generation is an \textbf{open-ended} task that benefits from large-scale sampling and post hoc selection (see~\cref{fig: motivation_task_diff}(a)).
In contrast, image editing is a \textbf{goal-directed} task. The solution space is constrained by the source image and instruction, even when varying noise or rewriting prompts (see~\cref{fig: motivation_task_diff}(b)).
This mismatch exposes three issues when transferring Image-CoT to editing:
\textbf{(1) Inefficient resource allocation}. Existing methods~\cite{TTS_Baseline, TTS_SANA, Image_CoT, ICEdit_arxiv2025} use a fixed sampling budget for all edits (\eg, 32 samples). However, \cref{fig: motivation}(a) shows that simple edits (high initial scores, obtained by MLLM evaluation before applying Image-CoT) see minimal improvement, while difficult edits (low initial scores) benefit more from Image-CoT. This fixed budget wastes computation on simple cases.
\textbf{(2) Unreliable early-stage verification}. Current methods~\cite{Image_CoT, ICEdit_arxiv2025} rely on general MLLM scores to evaluate intermediate denoising states for early pruning. However, editing often modifies subtle, localized regions of the source image, making these changes hard to distinguish in early denoising stages.
This causes general scores to misjudge sample quality: 40\% of samples scoring low at early stages achieve high final scores(see~\cref{fig: motivation}(b)). Such misjudgement leads to incorrect pruning of high-potential candidates, degrading final performance.
\textbf{(3) Redundant edited results}. 
Large-scale sampling in editing often generates multiple correct results with identical best scores. For instance, most edit cases with best scores in $[7, 9)$ obtain over 15 candidates that share the same best score (see ~\cref{fig: motivation}(c)).
However, one intent-aligned result is sufficient for the editing task. 
Existing pruning strategies~\cite{ICEdit_arxiv2025, TTS_Baseline, Image_CoT} typically adopt breadth-first search, generating all candidates in parallel before a final best-of-N selection. 
This leads to unnecessary computational cost on redundant correct results.


 
To address these issues, we propose \textbf{AD}aptive \textbf{E}dit-\textbf{CoT} (ADE-CoT), an on-demand test-time scaling framework that shifts focus from \textit{scale} to \textit{speed}. It improves efficiency while preserving editing correctness through three core strategies:
\textbf{(1) Difficulty-aware resource allocation}. Instead of a fixed sampling budget, ADE-CoT dynamically adjusts the budget based on estimated edit difficulty. Simple edits receive a minimal budget, whereas complex ones expand the search. This allocates computation to challenging edits. 
\textbf{(2) Edit-specific verification in early pruning}. To mitigate the misjudgement of general scores, we introduce edit-specific metrics to discover potentially successful edited images. We preview partially denoised images at intermediate timesteps to check edited-region localization accuracy and instruction–caption consistency. Moreover, samples with high visual similarity are discarded to avoid redundancy.
\textbf{(3) Depth-first opportunistic stopping}. To reduce redundant correct results, we incorporate a depth-first generation process. Candidates are generated sequentially by their early-stage scores. The process terminates when sufficient intent-aligned images are found. A two-stage instance-specific verifier guides this decision to confirm fine-grained correctness. Our key contributions are: 
\begin{itemize}[leftmargin=*,nosep]
    \item We identify three issues when applying Image-CoT to editing: inefficient resource allocation, unreliable early-stage verification, and redundant edited results. Based on this analysis, we propose ADE-CoT, an on-demand test-time scaling algorithm that enhances editing efficiency and correctness during large-scale sampling.
    \item We address these challenges with three mechanisms: (1) difficulty-aware resource allocation that assigns a dynamic budget for each instance; (2) edit-specific verification using region localization and caption consistency for accurate early pruning; and (3) depth-first opportunistic stopping guided by an instance-specific verifier.
    \item Extensive experiments on three SOTA editing models (Step1X-Edit~\cite{Step1X_Edit_arxiv2025}, BAGEL~\cite{Bagel_2025_arxiv}, and FLUX.1 Kontext~\cite{FLUX_Kontext_arxiv2025}) demonstrate that ADE-CoT achieves superior performance-efficiency trade-offs across three benchmarks. With comparable sampling budgets, it delivers better performance with more than 2$\times$ speedup over BoN. 
\end{itemize}


 
\section{Related Work}
\label{sec:related_work}



\textbf{Image editing} with diffusion models has advanced rapidly. Early methods~\cite{train_free_stable_flow, train_free_RF_Solver, train_free_BrushEdit, train_free_KV_Edit, train_free_Plug_and_Play, train_free_Null_text, train_free_FlowEdit, train_free_prompt_to_prompt, train_free_MasaCtrl} are often training-free, relying on prompt guidance, attention modulation, or inversion editing. However, they lack precise control in fidelity and controllability. 
To address this issue, later work~\cite{fine_tune_dataset_OmniEdit, fine_tune_dataset_emu_edit, fine_tune_UltraEdit, MagicBrush_NIPS2023, AnyEdit_CVPR2025, fine_tune_dataset_hq_edit, fine_tune_dataset_prompt_fix, fine_tune_dataset_InsightEdit, InstructPix2Pix_CVPR2023, fine_tune_dataset_seed_data} fine-tunes on high-quality, large-scale datasets with modified architectures. Recent approaches~\cite{MLLM_add_DiT_X-Fusion, MLLM_add_DiT_OminiControl, MLLM_add_DiT_Lumina_OmniLV, MLLM_add_DiT_ACE++, MLLM_add_DiT_ACE, MLLM_add_DiT_master_t2i_diffusion, MLLM_add_DiT_X2I, MLLM_add_DiT_SmartEdit, MLLM_add_DiT_MGIE, Step1X_Edit_arxiv2025, MLLM_and_DiT_SmartFreeEdit, FLUX_Kontext_arxiv2025, qwen_image_arxiv2025, lu2025end} combine multimodal large language models with diffusion decoders.
This enables instruction-following edits by fusing modalities at the latent level. 
Inspired by Gemini~\cite{gemini2_20250312} and GPT-4o~\cite{openai_2025_chatgpt4o}, subsequent work~\cite{Bagel_2025_arxiv, chu2025usp, MLLM_add_DiT_UniWorld, MLLM_add_DiT_Ming-Lite-Uni, MLLM_add_DiT_ILLUME+, MLLM_add_DiT_SEED-X, MLLM_add_DiT_VARGPT, MLLM_add_DiT_Nexus_Gen, MLLM_add_DiT_MetaQueries, Hunyuan3} has improved generation quality by jointly training understanding and generation tasks.
However, their performance in complex edits still faces challenges. In this paper, we adapt Image-CoT to editing to enhance generation while improving efficiency.




\noindent
\textbf{Test-time scaling in image generation} enhances quality by extending inference time. 
Inspired by Chain-of-Thought (CoT) in LLMs~\cite{NLP_CoT, NLP_CoT_ORM, NLP_CoT_PRM_1, NLP_CoT_PRM_2}, Image-CoT has emerged as a promising paradigm. Noise scaling~\cite{TTS_Baseline, TTS_SANA} generates multiple samples by perturbing the noise and selects the best as the final result. 
However, its computational cost scales linearly. Subsequent work aims to improve this trade-off.
Some methods enhance sample diversity through prompt-level intervention, such as rewriting~\cite{prompt_TTS_Yoonjin, T2IR1} or reflective updates~\cite{Reflection_DiT, GenRef_CoT_ICCV2025, Image_CoT}. Others adapt search algorithms~\cite{TTS_Baseline, classical_search_TTS, Tree_Sample_TTS, Video_TTS_EVO, TTS_search_ICML, TTS_greedy} (\eg, MCTS) to treat the reverse diffusion chain as a search trajectory and change the noise based on verifier scores.
Recent methods~\cite{Video_TTS, Image_CoT, Video_TTS_beam_search, UniGen_TTS, ICEdit_arxiv2025} use MLLMs as verifiers to prune low-potential trajectories early. However, most Image-CoT methods target T2I generation and are suboptimal for editing. To this end, we propose ADE-CoT, an edit-specific test-time scaling method, to address issues of inefficient resource allocation, unreliable early-stage verification, and redundant edited results during Image-CoT.










\noindent 
\textbf{Path pruning in Image-CoT} aims to remove low-potential samples by scoring intermediate states. Existing methods typically adopt breadth-first search. PRM [83] evaluates whether each denoising step meets quality requirements to prune candidates. PARM~\cite{Image_CoT} introduces two verifiers to judge which step is clear and assess whether it has high-quality potential. VideoTTS~\cite{Video_TTS} proposes Tree-of-Frames to leverage the feedback from multi-verifiers to guide the generation. 
ICEdit~\cite{ICEdit_arxiv2025} is the first work to incorporate Image-CoT into editing. It proposes the early filter strategy to generate preliminary images with a few additional denoising steps and select the optimal initial noise by a general MLLM verifier (\textit{i.e.}, VIE-Score~\cite{VIE_Score}). However, general scores may incorrectly remove high-potential candidates.
In contrast, we propose edit-specific metrics, including edited-region and caption verification, to mitigate this issue.
Moreover, an early preview mechanism is also introduced to obtain preliminary images without extra denoising steps. Besides, we incorporate depth-first opportunistic stopping to terminate the search early.




{
\begin{figure*}[t]
\begin{center}
\includegraphics[width=0.99\linewidth]{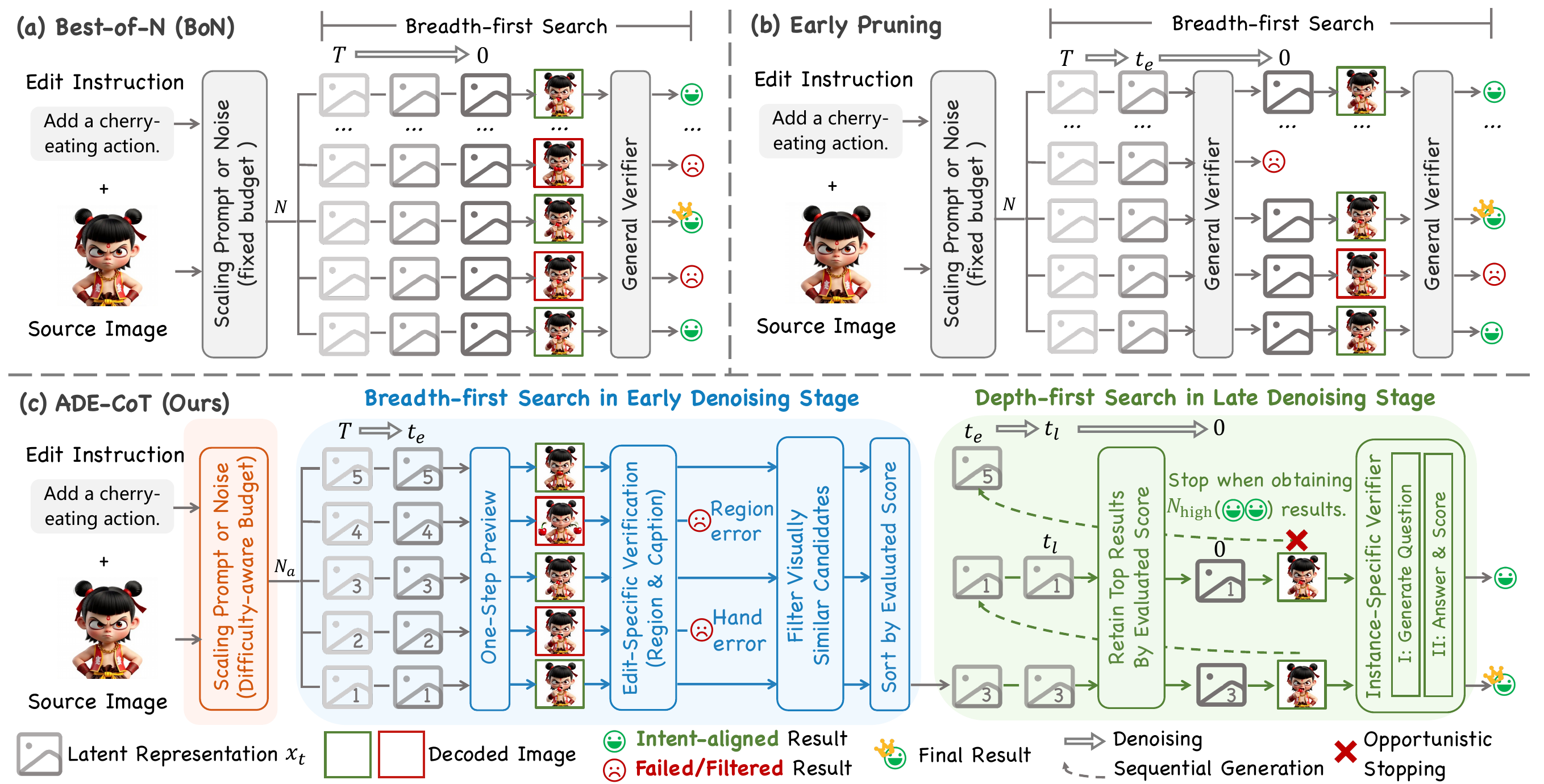}
\end{center}
\vspace{-1.7em}
\caption{\textbf{Pipeline comparison of Image-CoT methods for editing}. 
(a) \textbf{Best-of-N} employs a breadth-first search with a fixed sampling budget, producing all $N$ candidates before verification. 
(b) \textbf{Previous pruning strategies} improve efficiency by pruning with general MLLM scores.
(c) \textbf{Our ADE-CoT} employs three strategies for improved quality and efficiency: difficulty-aware resource allocation to dynamically adjust the sampling budget (\textcolor{method_orange}{\textbf{orange}}, \cref{sec: adaptive_sample}), edit-specific verification to identify promising candidates in early denoising stage (\textcolor{method_blue}{\textbf{blue}}, \cref{sec: prune_and_rank}), and depth-first opportunistic stopping to terminate search when intent-aligned results are obtained (\textcolor{method_green}{\textbf{green}}, \cref{sec: adapt_stop}).
}
\label{fig: method}
\vspace{-1.6em}
\end{figure*}
}

\section{Method}
\label{sec: method}

Our \textbf{AD}aptive \textbf{E}dit-\textbf{CoT} (ADE-CoT) framework is shown in~\cref{fig: method}(c) and detailed in Supp. Alg.3. We first propose a difficulty-aware resource allocation strategy to dynamically adjust sampling budgets based on edit difficulty (\cref{sec: adaptive_sample}). Then, an edit-specific verification is introduced to search promising candidates and discard errors in the early pruning stage (\cref{sec: prune_and_rank}). Subsequently, we adopt a depth-first sequential generation that stops adaptively when sufficient intent-aligned results are found. (\cref{sec: adapt_stop}).

\noindent \textbf{Preliminaries}. 
Given a source image $I_{\text{src}}$ and an edit instruction $c$, the goal of image editing is to generate an edited image $I$ that semantically aligns with the guidance provided by $c$. 
In Image-CoT, the standard Best-of-$N$ method (see~\cref{fig: method}(a) and Supp. Alg.1) consists of two stages: (1) Generation: A set of $N$ candidate images $\{ I_1, I_2, \cdots, I_N \}$ is generated by varying initial noise or rewriting prompts. 
(2) Selection: A verifier assigns a scalar score $S$ to each candidate, reflecting its semantic alignment with $c$:
\setlength{\abovedisplayskip}{2pt}
\setlength{\belowdisplayskip}{2pt}
\begin{equation}
    \texttt{Vrf}: I_{\text{src}} \times I \times c \rightarrow \mathbb{R}. 
\end{equation}
The common verifier is the general score $S_{\mathrm{gen}}$, which uses an MLLM with prompts (\eg, VIE-Score~\cite{VIE_Score}) to evaluate images on instruction adherence and aesthetic quality. 
The final output $I^*$ is the image with the highest score:
\begin{equation}
    I^* = \argmax_{{I_i} \in \{I_1, I_2, \cdots, I_N\} }  \texttt{Vrf}(I_{\text{src}}, I_i, c).
\end{equation}
Early Pruning~\cite{ICEdit_arxiv2025, Image_CoT, Video_TTS} is a standard method to improve efficiency (see~\cref{fig: method}(b) and Supp. Alg.2). It generates intermediate previews by sampling at timestep $t_e$ instead of the full denoising steps $T$, where $t_e < T$. Candidates scoring below the rejection threshold $S_{\mathrm{rj}}$ are pruned, avoiding the full generation of low-potential samples.

\subsection{Difficulty-aware Resource Allocation}
\label{sec: adaptive_sample}

To mitigate the inefficiency of fixed sampling budgets in previous methods~\cite{TTS_Baseline, Image_CoT, Video_TTS, GenRef_CoT_ICCV2025, ICEdit_arxiv2025}, we introduce a difficulty-aware resource allocation strategy (summarized in Supp. Alg.4), which dynamically adjusts sampling budgets based on a preliminary estimation of edit difficulty.

Given the source image $I_{\text{src}}$ and instruction $c$, we generate a single candidate and evaluate it with the verifier \texttt{Vrf}. This yields an initial score $S$, which acts as a proxy for edit difficulty. The adaptive budget $N_a$ is formulated as: 
\begin{equation}
    N_a = N_{\mathrm{min}} + \lceil (N - N_{\mathrm{min}}) \times (1 -  S / S_{\mathrm{max}})^\gamma \rceil,
    \label{eq: adapt_num}
\end{equation}
where $N_{\mathrm{min}}$ and $N$ respectively denote the minimal and original budgets, $S_{\mathrm{max}}$ is the maximum score, and $\gamma$ is a hyperparameter to control sensitivity. 
This formulation ensures that for easy edits where $S \rightarrow S_{\mathrm{max}}$, $N_a$ converges to $N_{\mathrm{min}}$. Conversely, for difficult edits where $S \rightarrow 0$, $N_a$ approaches $N$.
As a result, it allocates more computation to difficult cases and saves resources on easy ones. 



\subsection{Edit-specific Verification in Early Pruning}
\label{sec: prune_and_rank}

To address the misjudgement by general scores at early denoising stage, we introduce three key components for early pruning (see blue in~\cref{fig: method}(c) and Supp. Alg.5): (1) a one-step preview mechanism to obtain approximate previews of the final output; (2) a unified score using edit-specific verifiers to find high-potential candidates; and (3) a visual similarity filter to remove redundant results.
Finally, the retained candidates are ranked to guide the subsequent generation.


\textbf{One-step preview mechanism}. 
Directly evaluating the noisy latent $x_{t_e}$ at an early timestep $t_e$ is challenging, where $t_e \ll T$. 
Since recent editing models~\cite{Step1X_Edit_arxiv2025, Bagel_2025_arxiv, FLUX_Kontext_arxiv2025} are mainly trained with flow matching~\cite{flow_matching}, we estimate the approximate clean latent $x_{0|t_e}$ from $x_{t_e}$ in a single step:
\begin{equation}
    x_{0|{t_e}} = x_{t_e} - \sigma_{t_e} \epsilon_\theta(x_{t_e}, T_{t_e}).
\end{equation}
Here, $\sigma_{t_e}$ is the noise scale and $\epsilon_\theta (x_{t_e}, T_{t_e})$ is the predicted noise. 
$x_{0|{t_e}}$ is the predicted clean latent, which is then decoded into the preview image $I_{0|{t_e}}$. 
In Supp. B.2.1, we show that this preview reflects the correctness of the final output, which provides the basis for edit-specific verifiers. 

\textbf{Edit-specific verifiers}. To complement the general score $S_{\text{gen}}$, we introduce two verifiers that assess edited-region correctness and instruction-caption consistency:

\noindent 
\textbf{(1) Edited-region correctness}. 
A primary failure in image editing is the mislocalization of edits. To assess this, we first prompt the MLLM with $P_{\text{reg}}$ to identify the object to modify or keep unchanged. This object is then fed into Grounded SAM2~\cite{GroundedSAM} to generate a binary mask $M \in \{0, 1\}^{H\times W}$ for the expected edit region. We hypothesize that correct edits primarily affect pixels within this region. Given the edited image $I$ and source image $I_{\text{src}}$, we compute the per-pixel change map $\Delta \in \mathbb{R}^{H \times W}$ by averaging the absolute RGB differences across image channel dimension $C$: 
\begin{equation} 
\Delta = \frac{1}{C} \sum_{c=1}^{C} | I^{(c)} - I_{\text{src}}^{(c)} |.
\end{equation} 
We normalize $\Delta$ using a pixel-wise softmax to weight pixels by relative change magnitude. The score $S_{\text{reg}}$ is computed by aggregating changes within the mask $M$: 
\begin{equation} 
    S_{\mathrm{reg}} = \sum_{H,W} M  \odot \softmax_{H,W} (\Delta), 
\end{equation} 
where $\odot$ denotes the element-wise product. A higher $S_{\mathrm{reg}}$ indicates that the changes are better concentrated within the intended region. 
More details are in Supp. Sec. B.2.3.


\noindent \textbf{(2) Instruction-caption consistency}.
A standard metric in image editing is the semantic similarity between the edited image and a ground-truth caption. However, a key challenge is that such captions are unavailable at test time.
To address this, we instruct an MLLM with prompt $P_{\text{cap}}$,  conditioned on the source image $I_{\mathrm{src}}$ and the instruction $c$, to generate a targeted caption $c_{\text{cap}}$. This allows us to compute an image-caption consistency score using CLIP~\cite{CLIP}: 
\begin{equation}
    S_{\text{cap}} = \mathrm{CLIPScore}(I, c_{\text{cap}}), 
\end{equation}
where a higher $S_{\text{cap}}$ indicates better semantic alignment to the instruction. Detailed prompts are in Supp. B.2.4.

\textbf{Filtering error by evaluated score}. 
We combine the above metrics into a unified evaluation score $S$:
\begin{equation}
    S = S_{\text{gen}} + \lambda_{\text{reg}} S_{\text{reg}} + \lambda_{\text{cap}} S_{\text{cap}}, 
\end{equation}
where $\lambda_{\text{reg}}$ and $\lambda_{\text{cap}}$ are weighting factors. We compute the score for each preview image $I_{0|{t_e}}$ and prune low-potential candidates using a rejection threshold $S_{\text{rj}}$. In~\cref{sec: exp_efficiency_analysis}, we show that this strategy 
significantly reduces misjudgement of high-potential candidates in early pruning. 
As a result, our method improves efficiency while maintaining performance. Moreover, computing $S_{\text{reg}}$ and $S_{\text{cap}}$ requires only a single MLLM query per edit case. This design minimizes the additional MLLM overhead. 

\textbf{Filtering visually similar candidates}. 
Goal-directed image editing often yields multiple redundant edited results during large-scale sampling. Notably, this redundancy has been apparent in the early preview images (see Supp.B.2.5). To remove similar images, we extract visual embeddings from each preview using DINOv2~\cite{DINO_v2} and compute the pairwise similarity between them. If the similarity score between two preview images exceeds a threshold $\tau_{\text{sim}}$, we discard the one with the lower evaluation score. This step ensures that visually distinct and high-potential candidates are retained for the subsequent generation stage.

\textbf{Sorting by evaluated score}. 
Finally, the remaining candidates are sorted by $S$ in descending order. We empirically find that candidates with higher early scores tend to achieve higher final scores (see Supp. C.3). This enables the opportunistic stopping stage to terminate search earlier.

\subsection{Depth-first Opportunistic Stopping}
\label{sec: adapt_stop}
To avoid unnecessary computation on redundant yet correct results, we introduce a depth-first opportunistic stopping mechanism. 
The remaining candidates are processed sequentially based on early-stage scores. The search stops upon finding intent-aligned results (see green in Fig.~\ref{fig: method}(c) and Supp. Alg.6). It consists of two components: (1) a late-stage filter to retain the most promising candidates; (2) an instance-specific verifier to guide the stopping decision. 


\textbf{Retaining top results}.  
Motivated by the stronger correlation between final image quality and preview scores at later denoising stages (see Supp. C.3), we use an additional check at a later timestep $t_l$, where $t_e < t_l < T$. Similar to the early pruning stage, we generate a preview for each candidate and compute its unified score at timestep $t_l$. Instead of a fixed threshold, we apply an adaptive filter to retain candidates with scores comparable to the current highest score. 
This dynamically prunes suboptimal samples.

\textbf{Instance-specific verifier}.
While general scores $S_{\text{gen}}$ are effective for coarse-grained rank, they often assign the same top scores to many candidates (see~\cref{fig: motivation}(c)), even when some contain editing errors. This makes the final selection unreliable. 
To address this, we introduce an instance-specific verifier for fine-grained assessment. 
We found that a two-stage inquiry effectively guides the MLLM to notice critical details (see~\cref{sec: exp_efficiency_analysis}).
Given $I_{\text{src}}$ and $c$, it first generates a set of \emph{yes-no} questions about the current edit via prompt $P_{q}$, covering aspects such as instruction adherence and aesthetics. Then, it answers them using prompt $P_{a}$ to produce an instance-specific score $S_{\text{spec}}$ by counting \textit{yes} responses, where \textit{yes} indicates a correct edit of one aspect. 
We combine $S_\text{spec}$ with the unified score to penalize error candidates. The search process stops after $N_{\text{high}}$ candidates are intent-aligned (\ie, receiving \textit{yes} answers). Finally, the highest-scoring candidate is selected as the output.

\section{Experiments}
\label{sec: exp}


%



{
\setlength{\tabcolsep}{3.6pt}
\renewcommand{\arraystretch}{1.02} 
\begin{table*}[t]
    \caption{\textbf{Comparison of ADE-CoT with SOTA Image-CoT methods}. $\eta$ measures performance-efficiency trade-off and $\xi$ measures generation redundancy. Except for LPIPS, higher values are better for other metrics. The best results within a model are in \textbf{bold}. Non-degraded performances compared to the BoN method after pruning strategies are \colorbox{highlight_result}{highlighted}. All results are the average of three runs.}
    \label{tab: main_result}
    \vspace{-0.6em}
    \centering
    \resizebox{1.0\linewidth}{!}
    {
    \scriptsize
\begin{tabular}{l | c | ccccc | ccccc | ccccc }
\toprule
\multirow{2}{*}{\textbf{Model}} & \multirow{2}{*}{$N$} &  
\multicolumn{5}{c|}{\textbf{GEdit-Bench-EN~\cite{Step1X_Edit_arxiv2025} (Full set)}} 
& \multicolumn{5}{c|}{\textbf{AnyEdit-Test~\cite{AnyEdit_CVPR2025}}} & \multicolumn{5}{c}{\textbf{Reason-Edit~\cite{MLLM_add_DiT_SmartEdit}}} \\
 \cmidrule(lr){3-7} \cmidrule(lr){8-12} \cmidrule(lr){13-17}
&  & G\_SC & G\_PQ & G\_O & $\eta$ & $\xi$ 
& $\text{CLIP}_\text{im}$ & $\text{CLIP}_\text{out}$ & DINO &  $\eta$ & $\xi$ & PSNR & LPIPS$\downarrow$ & CLIP & $\eta$ & $\xi$  \\
\midrule

FLUX.1 Kontext~\cite{FLUX_Kontext_arxiv2025} & 1 & 6.517 & 7.548 & 6.021 & - & - & 0.874 & 0.302 & 0.774 & - & - & 25.135 & 0.066 & 21.361 & - & -   \\ 
w/ BoN~\cite{TTS_Baseline} & 32 & \highlight{7.132} & \highlight{\textbf{7.721}} & \highlight{6.641} & 0.66 & 0.12 & \highlight{0.882} & \highlight{0.307} & \highlight{0.784} & 0.66 & 0.21 & \highlight{25.657} & \highlight{0.054} & \highlight{21.635} & 0.24 & 0.24  \\ 
w/ PRM~\cite{Image_CoT} & 32 & 7.018 & 7.713 & 6.517 & 1.13 & 0.27 & 0.880 & 0.306 & 0.782 & 0.99 & 0.37 & 25.633 & 0.058 & 21.603 & 0.31 & 0.44  \\ 
w/ PARM~\cite{Image_CoT} & 32 & 7.087 & 7.716 & 6.563 & 0.77 & 0.29 & 0.881 & 0.307 & 0.783 & 0.85 & 0.36 & 25.498 & 0.056 & \highlight{21.656} & 0.26 & 0.39  \\
w/ TTS-EF~\cite{ICEdit_arxiv2025} & 32 & 6.866 & 7.657 & 6.376 & 0.98 & 0.57 & 0.878 & 0.305 & 0.779 & 0.72 & 0.33 & 25.509 & 0.057 & \highlight{21.639} & 0.44 & 0.46  \\ 
w/ TTS-EF (modified) & 32 & \highlight{7.142} & 7.699 & \highlight{6.643} & 0.79 & 0.51 & \highlight{0.882} & \highlight{0.307} & \highlight{\textbf{0.785}} & 0.62 & 0.35 & \highlight{25.657} & \highlight{0.054} & \highlight{21.637} & 0.25 & 0.49  \\
{w/ \textbf{ADE-CoT} (ours)} & {32}  & \highlight{\textbf{7.225}} & {7.719} & \highlight{\textbf{6.695}} & \textbf{1.47} & \textbf{0.66} & \highlight{\textbf{0.883}} & \highlight{\textbf{0.308}} & \highlight{0.784} & \textbf{1.61} & \textbf{0.58}  & \highlight{\textbf{25.755}} & \highlight{\textbf{0.053}} & \highlight{\textbf{21.663}} & \textbf{0.50} & \textbf{0.70}  \\
\textcolor{speed_up_color}{\textbf{Speedup (vs. BoN)}} & - & - & - & - & \textcolor{speed_up_color}{${\uparrow \boldsymbol{2.2}\times}$} & \textcolor{speed_up_color}{${\uparrow \boldsymbol{5.5}\times}$} & - & - & - & \textcolor{speed_up_color}{${\uparrow \boldsymbol{2.4}\times}$} & \textcolor{speed_up_color}{${\uparrow \boldsymbol{2.8}\times}$} & - & - & - & \textcolor{speed_up_color}{${\uparrow \boldsymbol{2.1}\times}$} & \textcolor{speed_up_color}{${\uparrow \boldsymbol{2.9}\times}$} \\
\midrule

BAGEL~\cite{Bagel_2025_arxiv} & 1 & 7.124 & 6.664 & 6.372 & - & - & 0.881 & 0.305 & 0.778 & - & - & 26.820 & 0.061 & 23.241 & - & -   \\
w/ BoN~\cite{TTS_Baseline} & 32 & \highlight{7.725} & \highlight{7.016} & \highlight{6.908} & 0.69 & 0.14 & \highlight{0.891} & \highlight{0.310} & \highlight{0.794} & 0.67 & 0.21 & \highlight{27.668} & \highlight{0.050} & \highlight{23.393} & 0.26 &  0.22  \\ 
w/ PRM~\cite{Image_CoT} & 32 & 7.496 & 6.854 & 6.685 & 1.17 & 0.33 & 0.888 & 0.308 & 0.785 & 1.12 & 0.36 & 27.231 & 0.051 & \highlight{23.408} & 0.37 & 0.41 \\ 
w/ PARM~\cite{Image_CoT} & 32 & 7.566 & 6.899 & 6.765 & 1.21 & 0.22 & 0.890 & 0.309 & 0.789 & 1.08 & 0.40 & 27.483 & \highlight{0.050} & 23.324 & 0.44 & 0.56 \\
w/ TTS-EF~\cite{ICEdit_arxiv2025}  & 32 & 7.402 & 6.834 & 6.660 & 1.15 & 0.43 & 0.886 & 0.307 & 0.788 & 1.12 & 0.30 & 27.387 & 0.054 & 23.230 & 0.30 & 0.27   \\  
w/ TTS-EF (modified) & 32 & \highlight{7.749} & 6.986 & \highlight{6.910} & 1.04 & 0.43 & \highlight{0.891} & \highlight{0.310} & 0.793 & 0.88 & 0.40 & \highlight{27.673} & \highlight{0.049} & \highlight{\textbf{23.409}} & 0.39 & 0.53   \\

{w/ \textbf{ADE-CoT} (ours)} & {32}  & \highlight{\textbf{7.823}} & {6.987} & \highlight{\textbf{6.972}} & \textbf{1.27} & \textbf{0.62}  & \highlight{\textbf{{0.893}}} & \highlight{\textbf{{0.311}}} & \highlight{\textbf{{0.796}}} & \textbf{1.64} & \textbf{0.53}  & \highlight{\textbf{{27.849}}} & \highlight{\textbf{{0.045}}} & \highlight{{23.399}} &\textbf{0.58} & \textbf{0.62}  \\
\textcolor{speed_up_color}{\textbf{Speedup (vs. BoN)}}  & - & - & - & - & \textcolor{speed_up_color}{${\uparrow \boldsymbol{1.8}\times}$}  & \textcolor{speed_up_color}{${\uparrow \boldsymbol{4.4}\times}$}  & - & - & - & \textcolor{speed_up_color}{${\uparrow \boldsymbol{2.4}\times}$} & \textcolor{speed_up_color}{${\uparrow \boldsymbol{2.5}\times}$} & - & - & - & \textcolor{speed_up_color}{${\uparrow \boldsymbol{2.2}\times}$} & \textcolor{speed_up_color}{${\uparrow \boldsymbol{2.8}\times}$} \\

\midrule

Step1X-Edit~\cite{Step1X_Edit_arxiv2025} & 1 &  7.002 & 7.085 & 6.403 & - & - & 0.865 & 0.302 & 0.742 & - & - & 21.443 & 0.106 & 22.463 & - & - \\ 
w/ BoN~\cite{TTS_Baseline} & 32 & \highlight{7.732} & \highlight{7.485} & \highlight{7.157} & 0.72 & 0.13 & \highlight{0.877} & \highlight{0.308} & \highlight{0.765} & 0.65 & 0.21 & \highlight{23.301} & \highlight{0.087} & \highlight{22.750} & 0.23 & 0.22  \\ 
w/ PRM~\cite{Image_CoT} & 32 & 7.647 & 7.405 & 7.031 & 0.94 & 0.22 & 0.874 & 0.307 & 0.762 & 1.03 & 0.40 & 23.118 & 0.090 & \highlight{22.791} & 0.31 & 0.41   \\ 
w/ PARM~\cite{Image_CoT} & 32 & 7.692 & 7.446 & 7.072 & 0.94 & 0.23 & 0.876 & 0.307 & 0.764 & 0.84 & 0.36 & 23.299 & 0.088 & \highlight{22.803} & 0.29 & 0.44  \\
w/ TTS-EF~\cite{ICEdit_arxiv2025}  & 32 & 7.296 & 7.301 & 6.777 & 0.96 & 0.51 & 0.873 & 0.306 & 0.758 & 0.70 & 0.33 & 22.273 & 0.095 & 22.502 & 0.34 & 0.24  \\ 
w/ TTS-EF (modified)  & 32 & \highlight{7.743} & 7.478 & \highlight{7.162} & 0.93 & 0.54 & \highlight{0.877} & \highlight{0.308} & \highlight{\textbf{0.766}} & 0.61 & 0.35 & 23.297 & \highlight{0.087} & \highlight{22.760} & 0.23 & 0.49  \\
{w/ \textbf{ADE-CoT} (ours)} & {32} & \highlight{\textbf{7.821}} & {7.465} & \highlight{\textbf{{7.196}}} & \textbf{1.45} & \textbf{0.62}  & \highlight{\textbf{0.878}} & \highlight{\textbf{0.309}} & \highlight{\textbf{0.766}} & \textbf{1.34} & \textbf{0.56}  & \textbf{\highlight{23.405}} & \highlight{\textbf{0.086}} & \highlight{\textbf{22.834}} & \textbf{0.46} & \textbf{0.63}  \\
\textcolor{speed_up_color}{\textbf{Speedup (vs. BoN)}} & - & - & - & - & \textcolor{speed_up_color}{${\uparrow \boldsymbol{2.0}\times}$} & \textcolor{speed_up_color}{${\uparrow \boldsymbol{4.8}\times}$} & - & - & - & \textcolor{speed_up_color}{${\uparrow \boldsymbol{2.1}\times}$} & \textcolor{speed_up_color}{${\uparrow \boldsymbol{2.7}\times}$} & - & - & - & \textcolor{speed_up_color}{${\uparrow \boldsymbol{2.0}\times}$} & \textcolor{speed_up_color}{${\uparrow \boldsymbol{2.9}\times}$} \\

\bottomrule
\end{tabular}
}
\vspace{-0.5 em}
\end{table*}
}

{
\begin{figure*}[t]
\centering
{
    \hfill
    \subfloat[FLUX.1 Kontext.]{\includegraphics[height=0.163\textwidth]{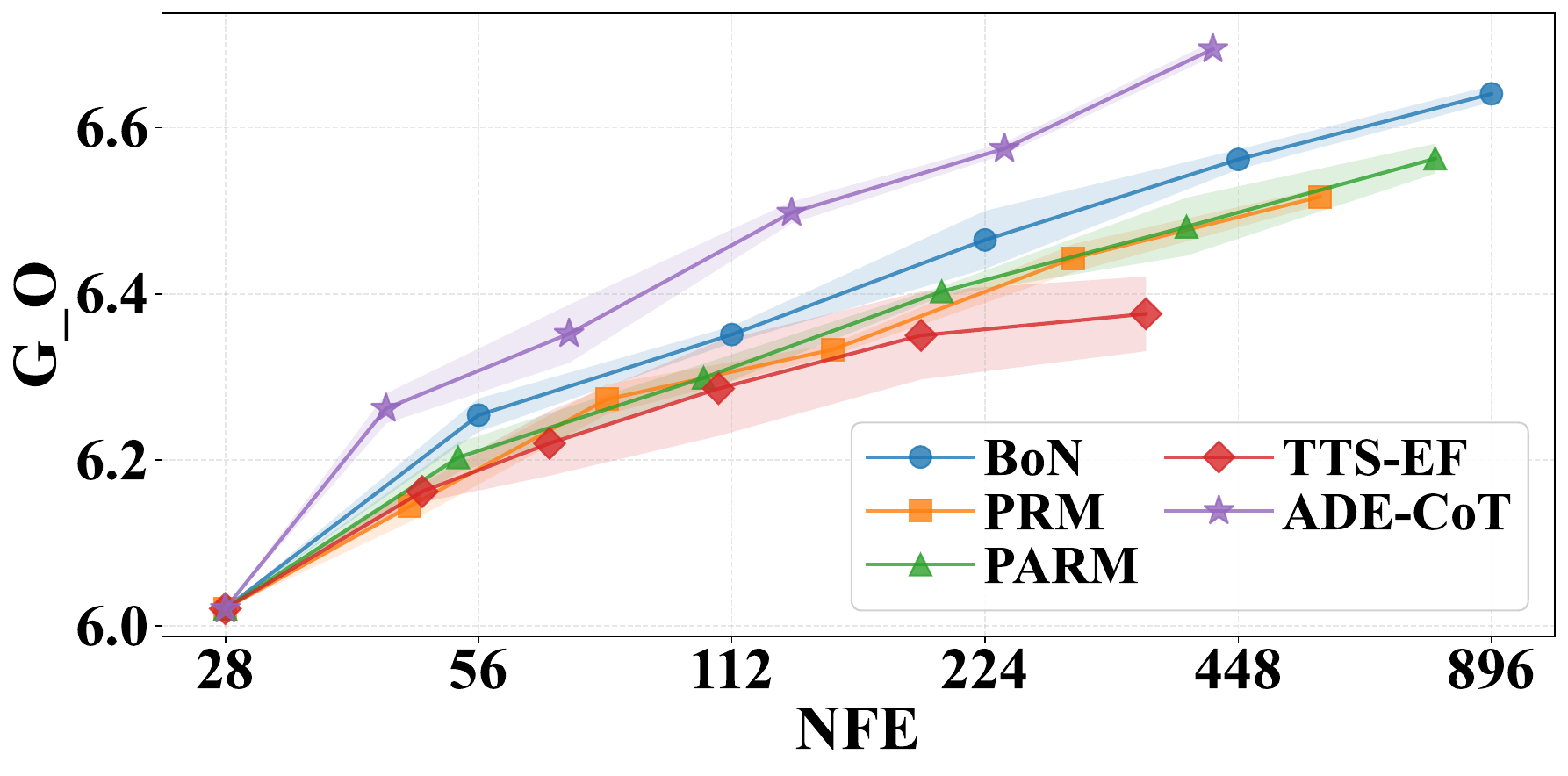}}
    \hfill
    \subfloat[BAGEL.]{\includegraphics[height=0.163\textwidth]{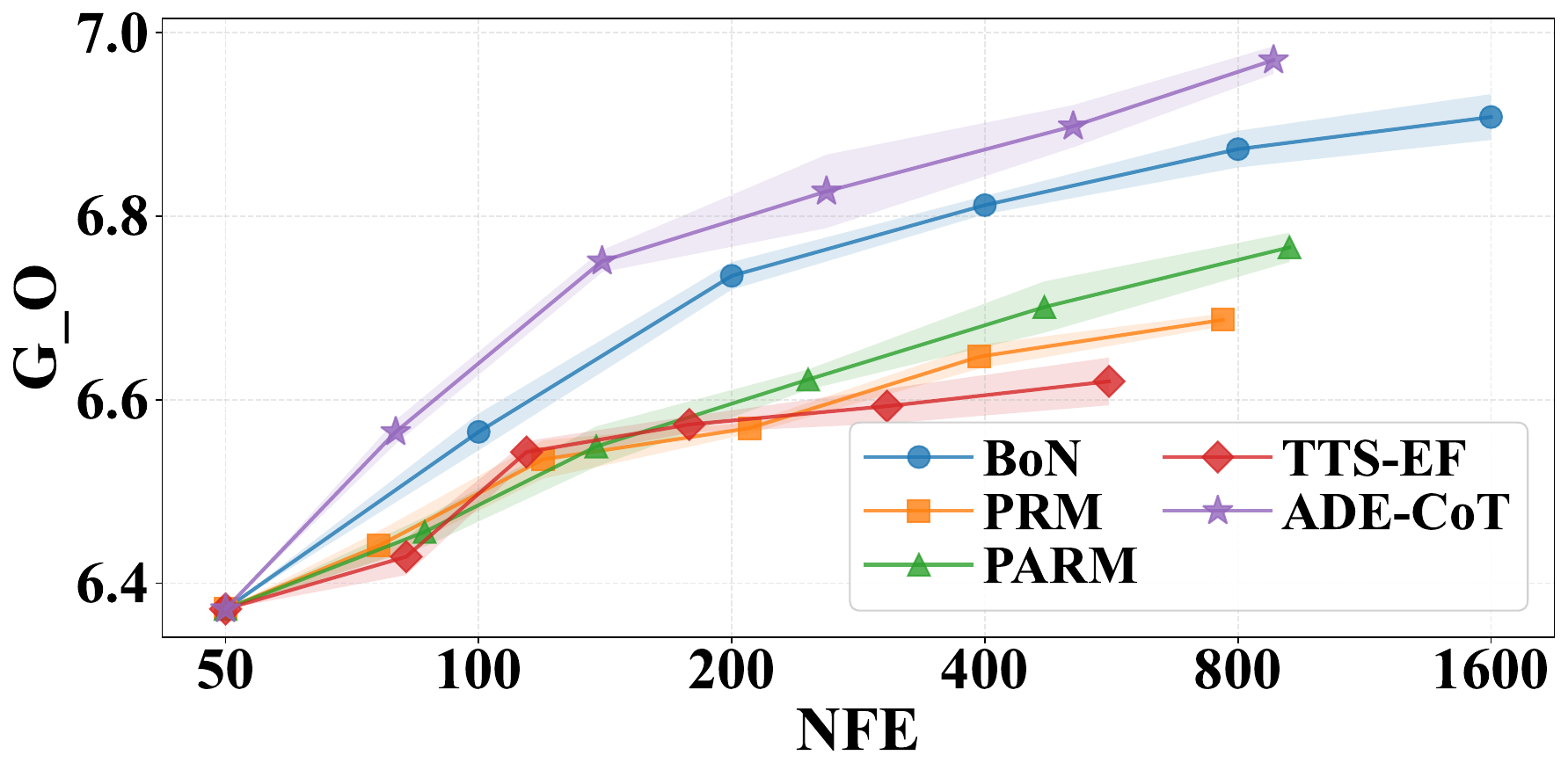}}
    \hfill
    \subfloat[Step1X-Edit.]{\includegraphics[height=0.163\textwidth]{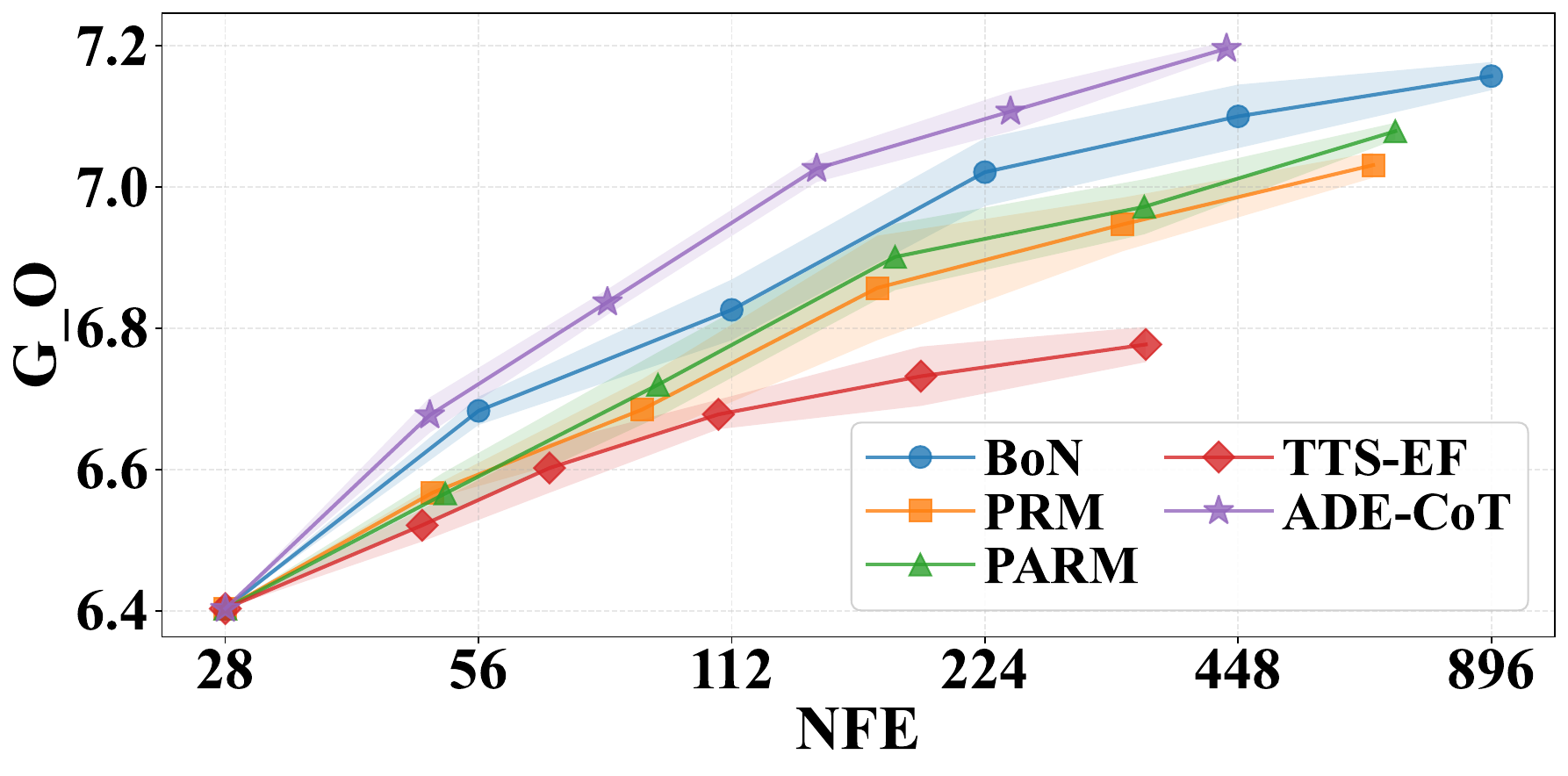}}
}
\vspace{-8pt}
\caption{\textbf{Scaling curves on GEdit-Bench across different editing models.} We show overall performance (G\_O, y-axis) versus computational cost (NFE, x-axis) for sampling budgets of $N = 1, 2, 4, 8, 16, 32$. The shaded regions indicate error bars. Our ADE-CoT (\textcolor{fig_purple}{\textbf{purple star}}) consistently surpasses SOTA Image-CoT methods across all models and budgets, achieving a better performance-efficiency trade-off.
}
\label{fig: scaling_laws}
\vspace{-1.3em}
\end{figure*}
}

\noindent \textbf{Evaluation settings}. 
We evaluate on three popular benchmarks:  
(1) GEdit-Bench contains real-world user edits. 
We use GPT4.1~\cite{GPT4} with VIE-Score~\cite{VIE_Score} to measure semantic consistency (G\_SC), perceptual quality (G\_PQ), and overall score (G\_O).
(2) AnyEdit-Test covers a range of tasks, such as local, global, and implicit editing tasks. Following ~\cite{AnyEdit_CVPR2025},  we report average semantic similarity ($\text{CLIP}_{\text{im}}$ and $\text{CLIP}_{\text{out}}$~\cite{CLIP, CLIPScore}) and visual similarity (DINO distance~\cite{DINO, DINO_v2}). 
(3) Reason-Edit involves complex understanding and reasoning scenarios. We follow~\cite{MLLM_add_DiT_SmartEdit} to evaluate with PSNR (dB)~\cite{metric_PSNR}, LPIPS~\cite{metric_LPIPS}, and CLIP Score.

\noindent \textbf{Metrics for efficiency}. 
Following~\cite{TTS_Baseline, Video_TTS, ICEdit_arxiv2025}, we measure computational cost via the Number of Function Evaluations (NFE), \textit{i.e.}, the total denoising steps in generation. 
To evaluate the balance between quality and computational cost after applying pruning strategies, we introduce the reasoning efficiency, $\eta = \frac{1}{M} \sum^{M}_{i=1} \sigma_i \cdot \frac{S^{(i)}}{S_{\text{max}}} \cdot \frac{NT}{\text{NFE}^{(i)}} $. Here, $\sigma_i =1$ if the final result achieves non-degraded performance compared to BoN, and $\sigma_i = 0$ otherwise. $S^{(i)}$ is the final score for instance $i$, and $S_{\text{max}}$ is the maximum score. $M$ is the number of test instances. A higher $\eta$ means a better trade-off between performance and efficiency. To assess generation redundancy, we propose outcome efficiency $\xi = \frac{1}{M} \sum^M_{i=1} \sigma_i \frac{\text{NFE}^{(i)}_{\text{min}}}{\text{NFE}^{(i)}}$. Here, $\text{NFE}_{\text{min}}^{(i)}$ denotes the NFE to reach the first image that achieves a non-degraded result compared to BoN. A higher $\xi$ means less redundancy.


\noindent \textbf{Implementation details}. 
Our ADE-CoT is evaluated on three SOTA, open-sourced image editing models: Step1X- Edit~\cite{Step1X_Edit_arxiv2025}, FLUX.1 Kontext~\cite{FLUX_Kontext_arxiv2025}, and BAGEL~\cite{Bagel_2025_arxiv}. We use their default denoising steps, \ie, $T = 28, 28, 50$. In the default setting, the early step $t_e$ is set to $8, 8, 16$  and the retain step $t_l$ is set to $16, 16, 36$. We adopt Qwen-VL-MAX~\cite{qwen25vl} for MLLM queries and VIE-Score~\cite{VIE_Score} as the general score $S_{\text{gen}}$. 
We generate five instance-specific \emph{yes-no} questions per edit. To ensure robustness, all results are averages of three runs. More details are in Supp. C.1.



\subsection{Comparison with SOTA Methods}

\textbf{Compared methods}. 
We compare ADE-CoT with SOTA Image-CoT methods.
Best-of-N (BoN)~\cite{TTS_Baseline} serves as the baseline. PRM~\cite{Image_CoT} and PARM~\cite{Image_CoT} assess intermediate denoised images using general MLLM scores to prune low-potential samples. 
TTS-EF~\cite{ICEdit_arxiv2025} generates early preview images by additional denoising steps and selects the best initial noise to continue generation. For fair comparison, we modify TTS-EF by increasing the number of retained samples to maintain performance comparable to BoN.

\textbf{Results under fixed sampling budget}. 
We first evaluate all methods under a fixed sampling budget ($N=32$) across three editing models and three benchmarks. Our analysis reveals three key findings.
\ding{182} ADE-CoT achieves comparable or superior performance to BoN while providing significant speedups. Specifically, it improves reasoning efficiency $\eta$ by over \textbf{2}$\times$ compared to BoN, indicating a superior performance–efficiency trade-off. Moreover, ADE-CoT increases outcome efficiency $\xi$ by an average of \textbf{4.9}$\times$, \textbf{2.7}$\times$, and \textbf{2.9}$\times$ on GEdit-Bench, AnyEdit, and Reason-Edit, respectively. This shows low redundancy through our proposed strategies. The scaling curves in~\cref{fig: scaling_laws} further demonstrate this advantage across different sampling budgets ($N=2, 4, 8, 16, 32$). ADE-CoT consistently achieves higher performance with lower computational cost than all baselines at every budget level.
\ding{183} PRM and PARM show limited performance compared to BoN. This is primarily because general scores struggle to accurately judge early preview images, causing many high-potential samples to be incorrectly discarded.
\ding{184} TTS-EF shows high efficiency scores but suffers from poor performance. This is because selecting only a single best sample from early previews becomes less reliable when scaling the number of samples.



\textbf{Results under comparable performance.}
We next compare methods that achieve non-degraded quality relative to BoN. We observe three key findings.
\ding{182} BoN, TTS-EF (modified), and ADE-CoT achieve similar results across datasets. Notably, \cref{tab: main_result} shows that ADE-CoT achieves the highest $\eta$ and $\xi$, demonstrating superior efficiency. 
\ding{183} Although TTS-EF (modified) maintains BoN-level performance, it shows lower efficiency than our method, even lower than the original TTS-EF. This is due to an increase in retained samples and additional denoising steps for early previews.
\ding{184} The scaling curves in~\cref{fig: scaling_laws} further confirm this advantage. When performance is comparable (same y-axis values), ADE-CoT requires less computation (lower x-axis values). This validates that our three strategies effectively allocate computation to challenging edits.

\subsection{Performance and Efficiency Analysis}
\label{sec: exp_efficiency_analysis}
In this section, we analyze the key reasons for the performance and efficiency improvement of our ADE-CoT.

{
\begin{figure}[t]
\centering
{
    \hfill
    \subfloat[NFE and performance versus $\gamma$.]{\includegraphics[height=0.156\textwidth]{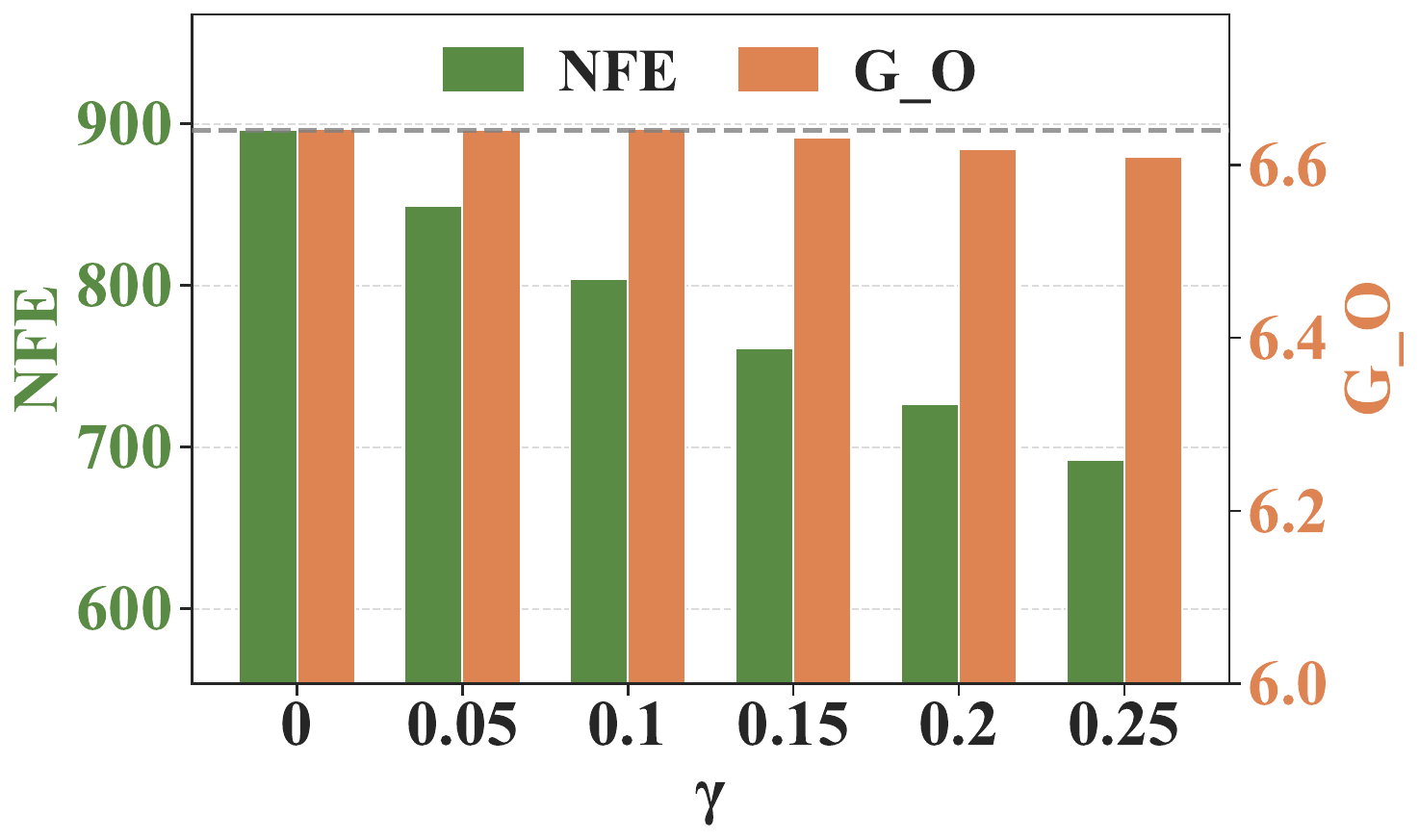}}
    \hfill
    \subfloat[Efficiency metrics versus $\gamma$.]{\includegraphics[height=0.156\textwidth]{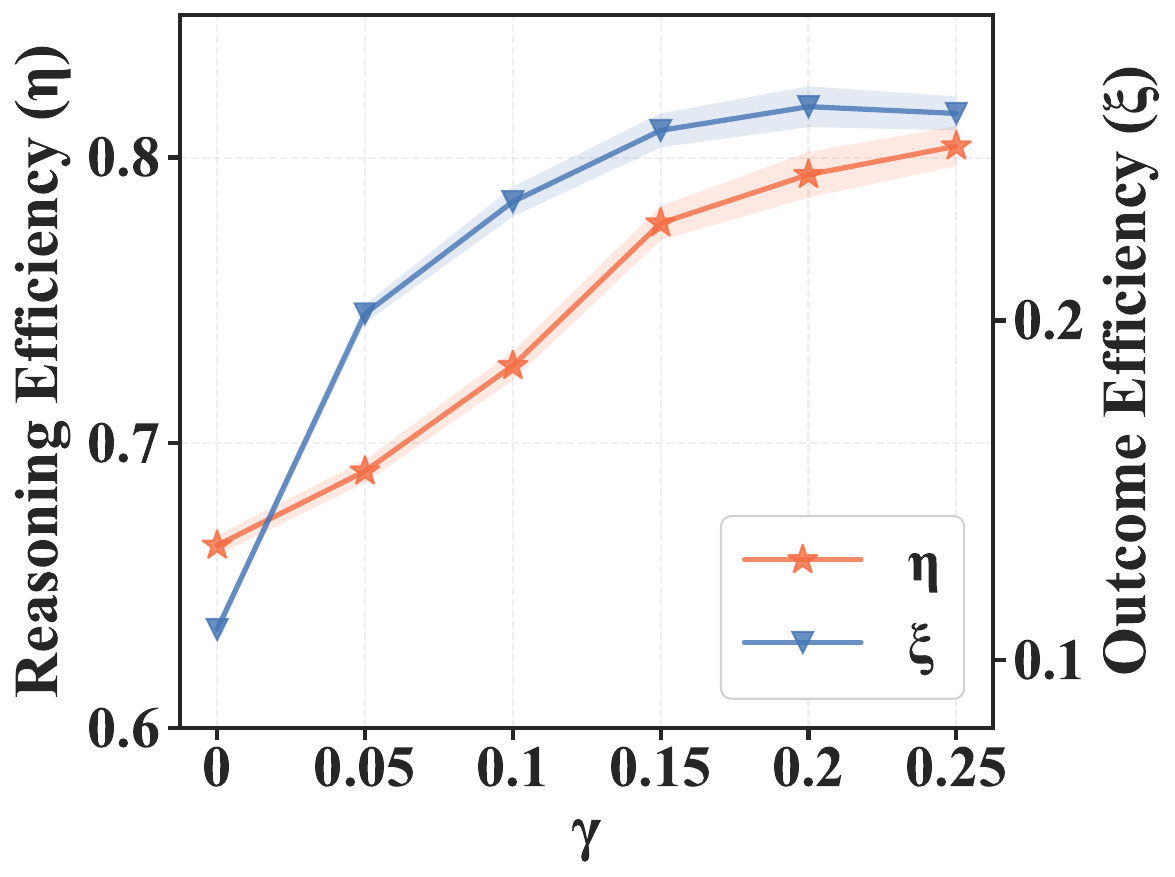}}
}
\vspace{-10pt}
\caption{\textbf{Effect of $\gamma$ in difficulty-aware resource allocation}.
}
\label{fig: analysis_adapt_sample}
\vspace{-9pt}
\end{figure}
}

{
\begin{figure}[t]
\centering
{
    \hfill
    \subfloat[Distribution of pruned candidates.]{\includegraphics[height=0.156\textwidth]{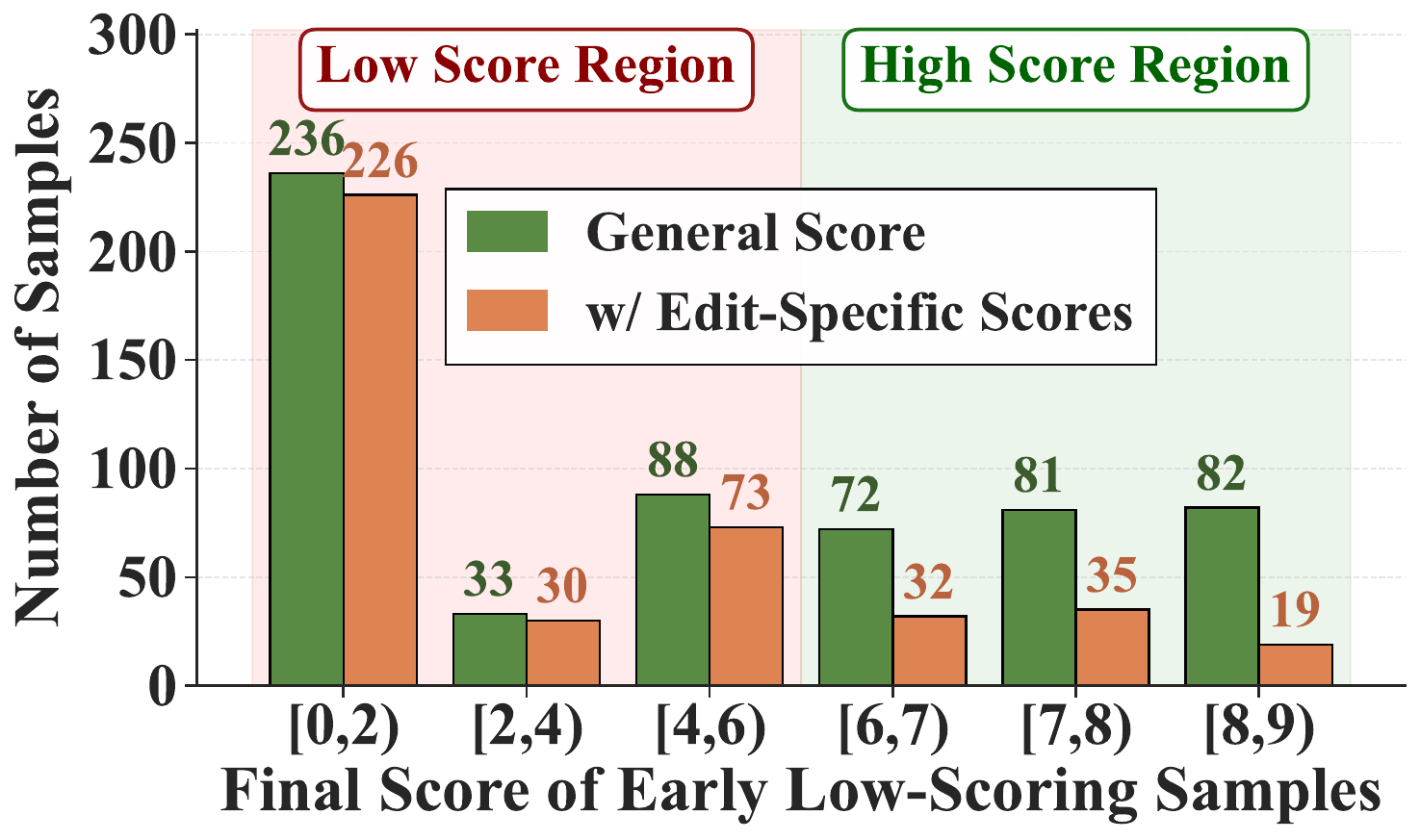}}
    \hfill
    \subfloat[Scaling curves.]{\includegraphics[height=0.156\textwidth]{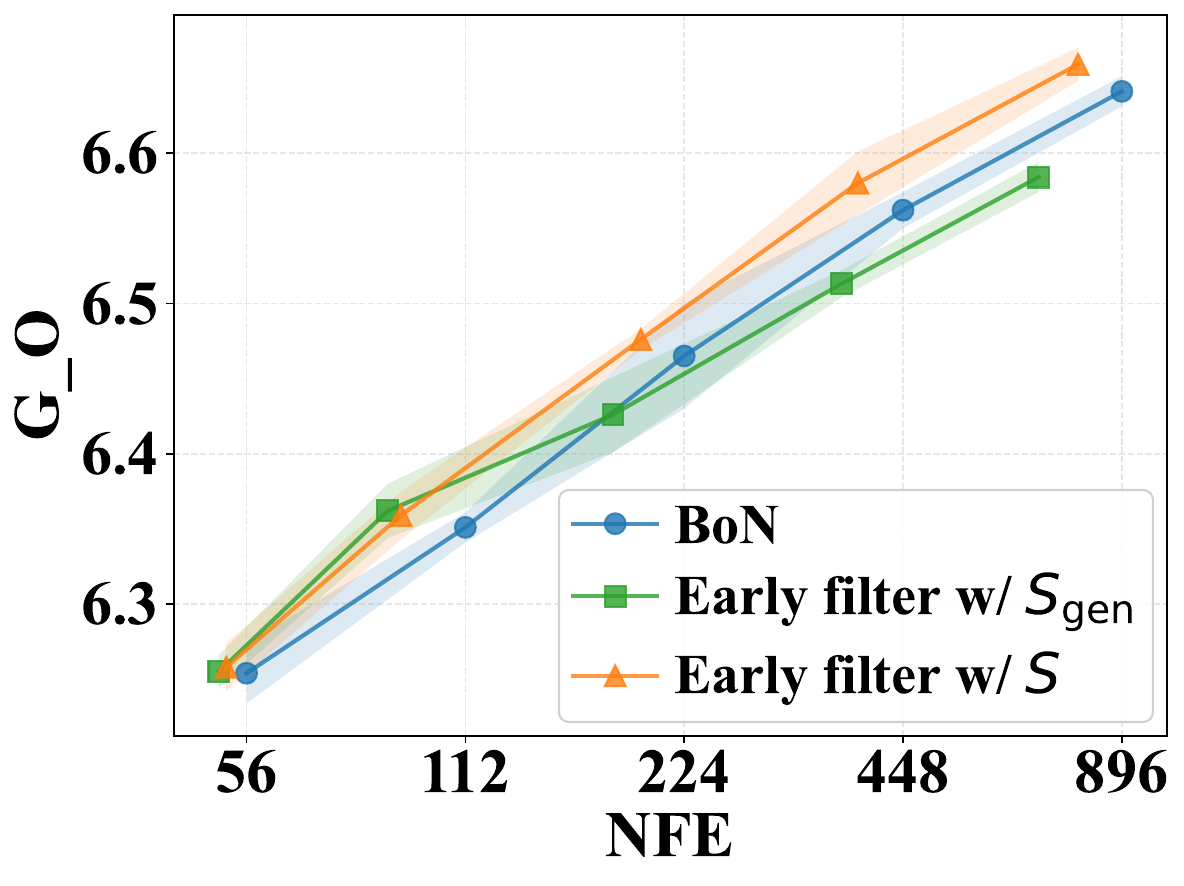}}
}
\vspace{-10pt}
\caption{\textbf{Effect of edit-specific verification on early pruning.}
}
\label{fig: edit_specific_compare_general}
\vspace{-1.3em}
\end{figure}
}


\textbf{Difficulty-aware resource allocation reduces costs on easy edits.} To validate this, we study the impact of the threshold $\gamma$ in~\cref{eq: adapt_num} on both performance and efficiency: 
\ding{182} In~\cref{fig: analysis_adapt_sample}(a), we show that increasing $\gamma$ from $0$ (equivalent to BoN Baseline) steadily reduces NFE while performance remains nearly unchanged until $\gamma$ exceeds $0.15$.  
This confirms that reducing budgets for simple edits enhances efficiency while preserving image quality. Row a) in~\cref{tab: efficiency_improve} further confirms this finding. 
\ding{183} We further examine the efficiency gains in~\cref{fig: analysis_adapt_sample}(b). Both efficiency metrics ($\eta$ and $\xi$) increase as $\gamma$ grows. Growth slows beyond $\gamma = 0.15$, mainly because the performance starts to decline. To balance efficiency and quality, we set the default $\gamma$ to $0.15$.


\textbf{Edit-specific verifiers reduce misjudgement in early pruning}. We demonstrate this with three key findings: 
\ding{182} \cref{fig: edit_specific_compare_general}(a) shows that using the general score alone causes significant misjudgement in the high score region $[6, 9)$. 
After utilizing edit-specific scores, misjudgement drops from $235$ to $86$ (a $63$\% reduction). 
Meanwhile, pruned low‑score samples remain nearly unchanged ($357 \rightarrow 329$). 
This confirms that assessing edited-region correctness and caption consistency effectively improves pruning accuracy. 
\ding{183} \cref{fig: edit_specific_compare_general}(b) shows performance across different budgets.
Under an identical pruning threshold, the general score achieves lower performance as sampling increases.
When $N > 8$, accumulated misjudgement causes its performance to drop below BoN. In contrast, our unified score $S$ with edit-specific verifiers maintains superior results. \ding{184} Rows b) and c) in~\cref{tab: efficiency_improve} quantify the NFE reduction when maintaining comparable performance to BoN. By mitigating misjudgement, early pruning with $S$ enables a higher rejection threshold, achieving greater NFE reduction than $S_\text{gen}$.



{
\begin{figure}[t]
\begin{center}
\includegraphics[width=0.95\linewidth]{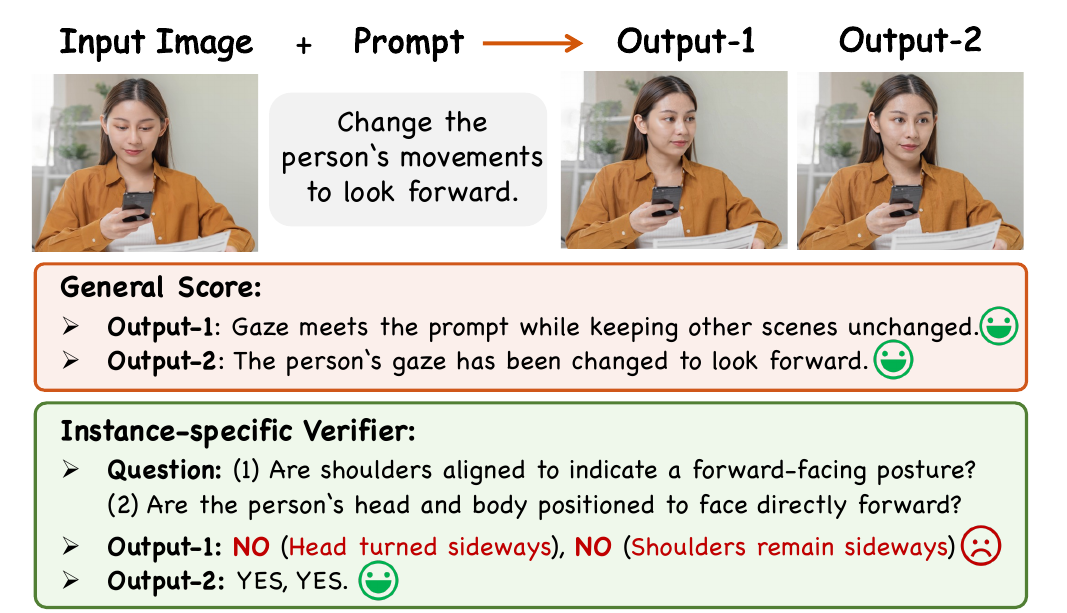}
\end{center}
\vspace{-1.6em}
\caption{
\textbf{Instance-specific verification detects subtle errors}. Both candidates receive similarly high general scores. In contrast, our instance-specific verifier reveals the critical flaw in the left image (``\textit{head turned sideways}''), improving final selection accuracy.
}
\label{fig: visualization_edit_specific}
\vspace{-0.9em}
\end{figure}
}

{
\begin{figure}[t]
\centering
{
    \hfill
    \subfloat[Scaling curves. Shaded regions are efficiency gains from opportunistic stopping.]{\includegraphics[height=0.165\textwidth]{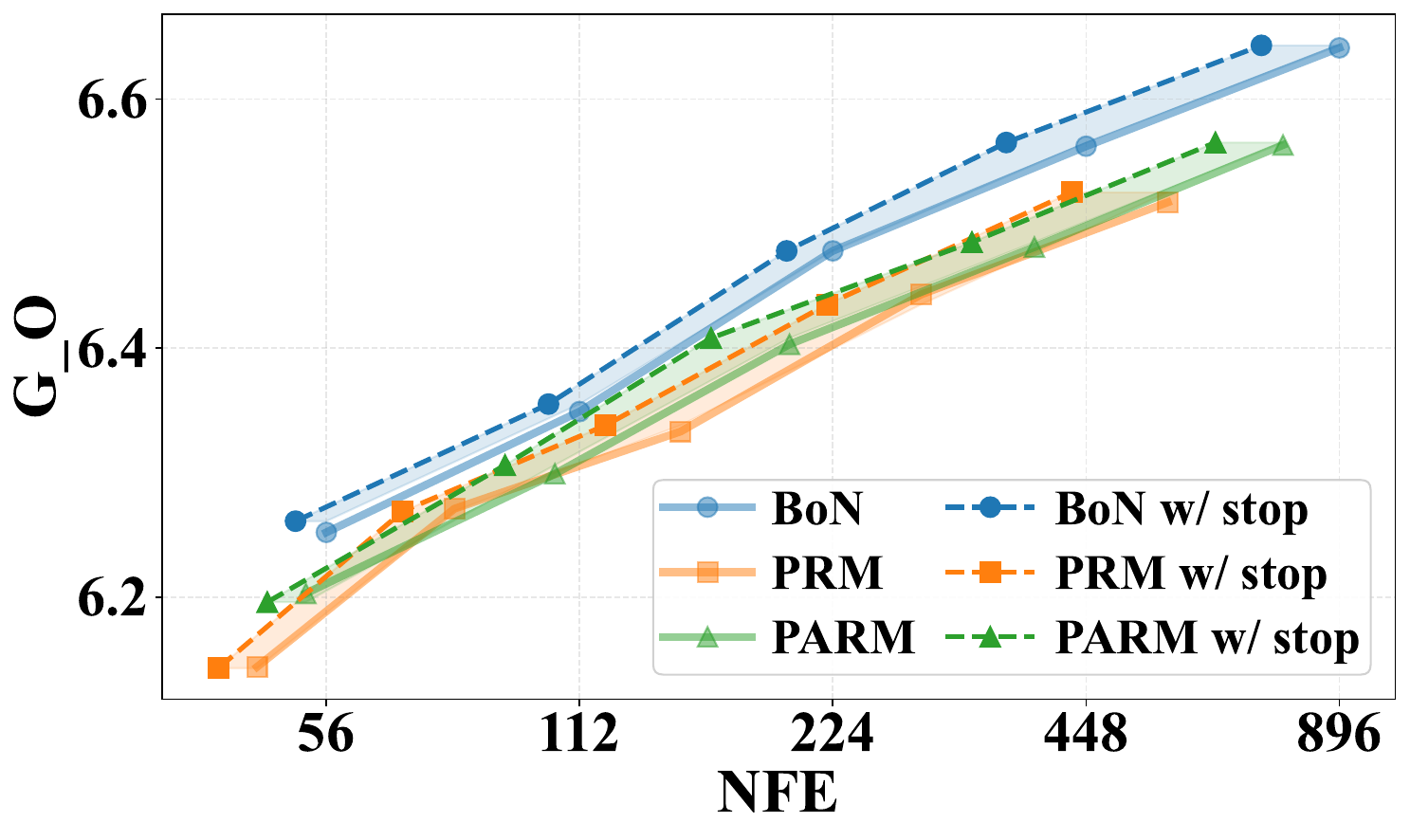}}
    \hfill
    \subfloat[Effect of $N_{\text{high}}$.]{\includegraphics[height=0.165\textwidth]{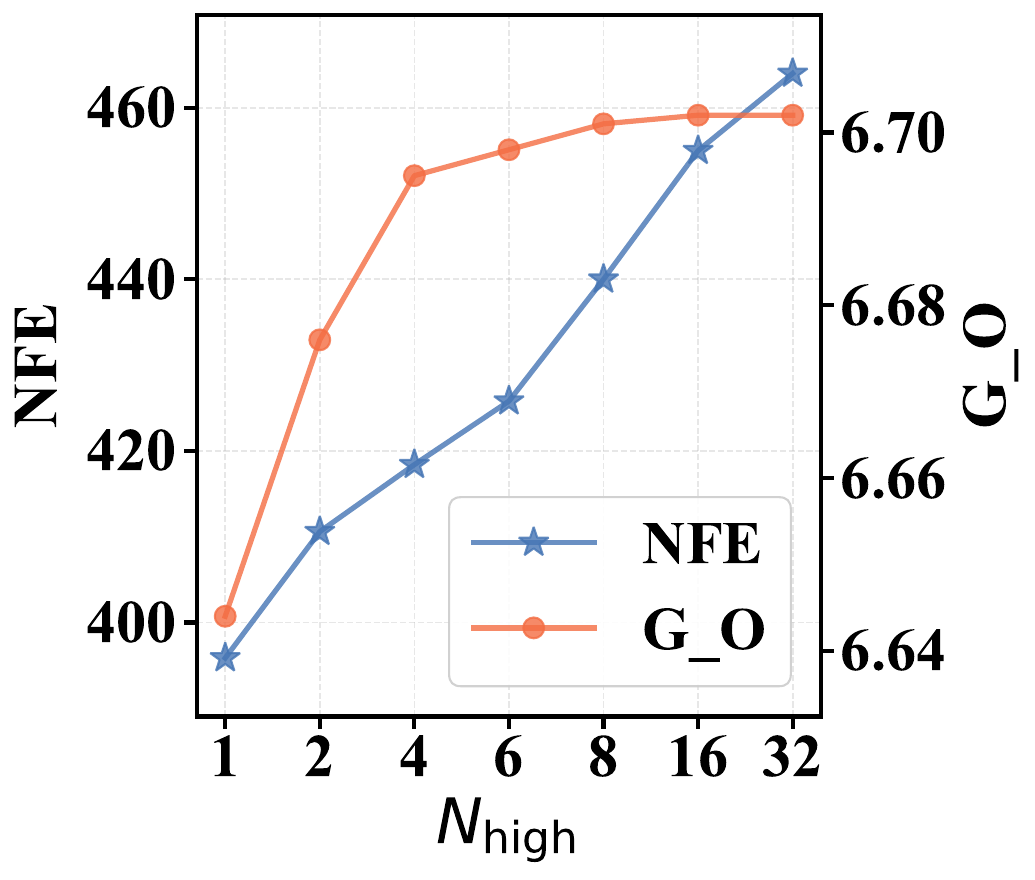}}
}
\vspace{-9pt}
\caption{\textbf{Impact of opportunistic stopping strategy.} 
}
\label{fig: adapt_stop_analysis}
\vspace{-1.5em}
\end{figure}
}


\textbf{Opportunistic stopping reduces redundancy in late denoising}.  
We analyze its performance and efficiency gains from three aspects:
\ding{182} Instance-specific verification improves final selection. As shown in~\cref{fig: visualization_edit_specific}, general scores often fail to distinguish correct from flawed outputs, assigning similarly high values to both. In contrast, our verifier generates targeted questions (\textit{e.g.}, ``Are shoulders aligned?'', ``Is the head facing forward?'') that correctly identify errors. Row f) in~\cref{tab: efficiency_improve} quantitatively confirms this improvement across all models.
\ding{183} Depth-first opportunistic stopping reduces redundant computation. Our strategy exploits inherent sampling redundancy in image editing (\cref{fig: motivation}(c)). When integrated with breadth-first pruning methods (BoN, PRM, PARM), it consistently reduces their NFE while maintaining comparable performance, as shown in \cref{fig: adapt_stop_analysis}(a).  
Row g) in~\cref{tab: efficiency_improve} confirms at least a 10\% NFE reduction with minimal quality drop.
\ding{184} Optimal $N_{\text{high}}$ balances performance and efficiency. \cref{fig: adapt_stop_analysis}(b) shows that performance (orange) saturates when $N_{\text{high}} \geq 4$, while NFE (blue) increases linearly. To balance these factors, we set $N_\text{high} =4$ as the default. 
Although stopping after the first intent-aligned image could reduce computation, collecting four high‑quality candidates improves robustness across diverse edits.

\subsection{Ablation Study}


{
\setlength{\tabcolsep}{3pt}
\renewcommand{\arraystretch}{1.01} 
\begin{table}[t]
    \caption{\textbf{Effect of the three proposed strategies on efficiency and performance}. We evaluate our method on GEdit-Bench~\cite{Step1X_Edit_arxiv2025}.}
    \label{tab: efficiency_improve}
    \vspace{-0.6em}
    \centering
    \resizebox{1.0\linewidth}{!}
    {
    \scriptsize
\begin{tabular}{l | cc | cc | cc }
\toprule
\multirow{2}{*}{\textbf{Model}} & \multicolumn{2}{c|}{\textbf{Kontext}} & \multicolumn{2}{c|}{\textbf{BAGEL}} & \multicolumn{2}{c}{\textbf{Step1X-Edit}}  \\ 
\cmidrule(lr){2-3} \cmidrule(lr){4-5} \cmidrule(lr){6-7}
 & G\_O$\uparrow$ & NFE$\downarrow$ & G\_O$\uparrow$ & NFE$\downarrow$ & G\_O$\uparrow$ & NFE$\downarrow$ \\ 
\midrule 
Baseline (BoN) & 6.641 & 896 & 6.908 & 1600 & 7.157 & 896  \\
\midrule 
\HighTableOrange{\texttt{a)} $+$difficulty-aware budgets} & \HighTableOrange{6.641} & \HighTableOrange{797} & \HighTableOrange{6.909} & \HighTableOrange{1391} & \HighTableOrange{7.157} & \HighTableOrange{778}  \\ 
\midrule 
\HighTableBlue{\texttt{b)} $+$early pruning (w/ $S_\text{gen}$)} & \HighTableBlue{6.642} & \HighTableBlue{719} & \HighTableBlue{6.912} & \HighTableBlue{1351} & \HighTableBlue{7.157} & \HighTableBlue{719} \\ 
\HighTableBlue{\texttt{c)} $+$early pruning (w/ $S$)} & \HighTableBlue{6.647} & \HighTableBlue{673} & \HighTableBlue{6.916} & \HighTableBlue{1290} & \HighTableBlue{7.161} & \HighTableBlue{638}\\ 
\HighTableBlue{\texttt{d)} $+$filtering similar samples} & \HighTableBlue{6.651} & \HighTableBlue{508} & \HighTableBlue{6.915} & \HighTableBlue{1087} & \HighTableBlue{7.162} & \HighTableBlue{522}  \\  
\midrule 
\HighTableGreen{\texttt{e)} $+$late retaining} & \HighTableGreen{6.652} & \HighTableGreen{464} & \HighTableGreen{6.935} & \HighTableGreen{972} & \HighTableGreen{7.163}  & \HighTableGreen{462}  \\ 

\HighTableGreen{\texttt{f)} $+$instance-specific verifier} & \HighTableGreen{\textbf{6.702}} & \HighTableGreen{464} & \HighTableGreen{\textbf{6.984}} & \HighTableGreen{972} & \HighTableGreen{\textbf{7.206}}  & \HighTableGreen{462} \\  

\HighTableGreen{\texttt{g)} $+$opportunistic stopping (\textbf{full})} & \HighTableGreen{6.695} & \HighTableGreen{\textbf{418}} & \HighTableGreen{6.972} & \HighTableGreen{\textbf{882}} & \HighTableGreen{7.196} & \HighTableGreen{\textbf{434}} \\
\bottomrule
\end{tabular}
}
\vspace{-0.8 em}
\end{table}
}

\textbf{Does filtering visually redundant images degrade performance?}
Row d) of~\cref{tab: efficiency_improve} shows that removing similar candidates reduces NFE by $24\%, 16\%, 18\%$ for Kontext, BAGEL, and Step1X-Edit, respectively, with almost no change in G\_O. This indicates that many candidates are redundant and can be discarded without harming quality. 

\noindent \textbf{What is the optimal way to obtain early previews?}
We compare three methods in~\cref{tab: ablation_obtain_early_preview}: (1) extra denoising steps (as in TTS-EF~\cite{ICEdit_arxiv2025}), (2) directly decoding noisy latents, and (3) our one-step preview. 
Additional denoising steps produce high-quality previews but significantly increase NFE. 
Decoding noisy latents provides clear previews only at later stages, leading to the highest NFE. 
In contrast, our one-step preview generates clear images in early steps with no extra iterations, achieving comparable G\_O with the lowest NFE across all models.
Detailed analysis is in Supp. C.3.

{
\setlength{\tabcolsep}{4.8pt}
\renewcommand{\arraystretch}{1.01} 
\begin{table}[t]
    \caption{\textbf{Ablation of obtaining early preview images}.}
    \label{tab: ablation_obtain_early_preview}
    \vspace{-0.6em}
    \centering
    \resizebox{1.0\linewidth}{!}
    {
    \scriptsize
\begin{tabular}{l | cc | cc | cc }
\toprule
\multirow{2}{*}{\textbf{Model}} & \multicolumn{2}{c|}{\textbf{Kontext}} & \multicolumn{2}{c|}{\textbf{BAGEL}} & \multicolumn{2}{c}{\textbf{Step1X-Edit}}  \\ 
\cmidrule(lr){2-3} \cmidrule(lr){4-5} \cmidrule(lr){6-7}
 & G\_O$\uparrow$ & NFE$\downarrow$ & G\_O$\uparrow$ & NFE$\downarrow$ & G\_O$\uparrow$ & NFE$\downarrow$ \\ 
\midrule 

\texttt{a)} w/ additional steps & 6.678 & 523 & 6.952 & 1008 & 7.188 & 525  \\
\texttt{b)} w/ noisy latents & 6.648 & 790 & 6.945 & 1334 & 7.153 & 765 \\
\midrule 
\HighTableBlue{\textbf{ADE-CoT} (full)} & \HighTableBlue{\textbf{6.695}} & \HighTableBlue{\textbf{418}} & \HighTableBlue{\textbf{6.972}} & \HighTableBlue{\textbf{882}} & \HighTableBlue{\textbf{7.196}} & \HighTableBlue{\textbf{434}} \\

\bottomrule 
\end{tabular}
}
\vspace{-0.8 em}
\end{table}
}

{
\setlength{\tabcolsep}{3.1pt}
\renewcommand{\arraystretch}{1.02} 
\begin{table}[t]
    \caption{\textbf{Ablation of different search ways for pruning strategy}.}
    \label{tab: ablation_search_way}
    \vspace{-0.6em}
    \centering
    \resizebox{1.0\linewidth}{!}
    {
    \scriptsize
\begin{tabular}{l | ccc | ccc | ccc }
\toprule
\multirow{2}{*}{\textbf{Model}} & \multicolumn{3}{c|}{\textbf{Kontext}} & \multicolumn{3}{c|}{\textbf{BAGEL}} & \multicolumn{3}{c}{\textbf{Step1X-Edit}}  \\ 
\cmidrule(lr){2-4} \cmidrule(lr){5-7} \cmidrule(lr){8-10}
 & G\_O$\uparrow$ & NFE$\downarrow$ & $\eta$ $\uparrow$ & G\_O$\uparrow$ & NFE$\downarrow$ & $\eta$ $\uparrow$ & G\_O$\uparrow$ & NFE$\downarrow$ & $\eta$ $\uparrow$ \\ 
\midrule 
\texttt{a)} w/ BFS~\cite{BFS} & \textbf{6.702} & 464 & 1.37 & \textbf{6.984} & 972 & 1.12 & \textbf{7.206} & 462  & 1.36 \\
\texttt{b)} w/ DFS~\cite{DFS} & 6.644 & 574 & 1.32 & 6.966 & 1073 & 1.08 & 7.162 & 554 & 1.29 \\
\texttt{c)} w/o sorting & 6.694 & 433 & 1.42 & 6.972 & 905 & 1.23 & 7.194 & 445 & 1.41 \\
\midrule 
\HighTableGreen{\textbf{ADE-CoT} (full)} & \HighTableGreen{{6.695}} & \HighTableGreen{\textbf{418}} & 
\HighTableGreen{\textbf{1.47}} & \HighTableGreen{{6.972}} & \HighTableGreen{\textbf{882}} & \HighTableGreen{\textbf{1.27}} & \HighTableGreen{{7.196}} & \HighTableGreen{\textbf{434}} & \HighTableGreen{\textbf{1.45}} \\

\bottomrule 
\end{tabular}
}
\vspace{-0.9 em}
\end{table}
}

\noindent \textbf{Which search strategy is most effective for pruning?} 
\cref{tab: ablation_search_way} compares Breadth-First Search (BFS) and Depth-First Search (DFS) with our hybrid search: BFS in the early denoising stage and DFS in the late denoising stage. While BFS (row a) yields the highest G\_O, our ADE-CoT achieves a superior performance-efficiency balance with the highest $\eta$. 
This validates that early BFS preserves high-potential candidates, while late DFS avoids unnecessary computation once aligned results are found.
Row c) shows that sequential generation further reduces NFE without quality drop.

\noindent \textbf{What are the optimal timesteps for early pruning and late retaining?} 
We analyze the impact of $t_e$ and $t_l$ for FLUX.1 Kontext in~\cref{fig: hyper_result}. Increasing the early step $t_e$ improves quality up to $t_e = 8$, but higher values only increase cost. This is because preview images become clearer and enable more accurate pruning. For the retain step $t_l$, performance increases until $16$ and then saturates, since previews are already high-fidelity. We set $t_e = 8$ and $t_l = 16$ as the optimal trade-off. More results are in Supp. Sec.C.3.

\noindent \textbf{How do MLLM verifiers affect performance?} 
We evaluate ADE-CoT with three MLLMs in~\cref{tab: ablation_mllm}, including two open-source models (Qwen2.5-VL-72B~\cite{qwen25vl} and Qwen3-VL-32B~\cite{qwen3}) and one proprietary model (Qwen-VL-MAX~\cite{qwen25vl}). We demostrate that ADE-CoT is robust to different verifiers.
\ding{182} It achieves over \textbf{2}$\times$ speedup across all MLLMs compared to BoN.  
\ding{183} The benefits also scale with MLLM capability. 
Stronger MLLMs, such as Qwen3-VL, yield higher G\_O and larger efficiency gains. 
Additional discussion on the effect of MLLMs are in Supp. Sec.C.7.





{
\begin{figure}[t]
\centering
{
    \hfill
    \subfloat[early step $t_e$]{\includegraphics[width=0.235\textwidth]{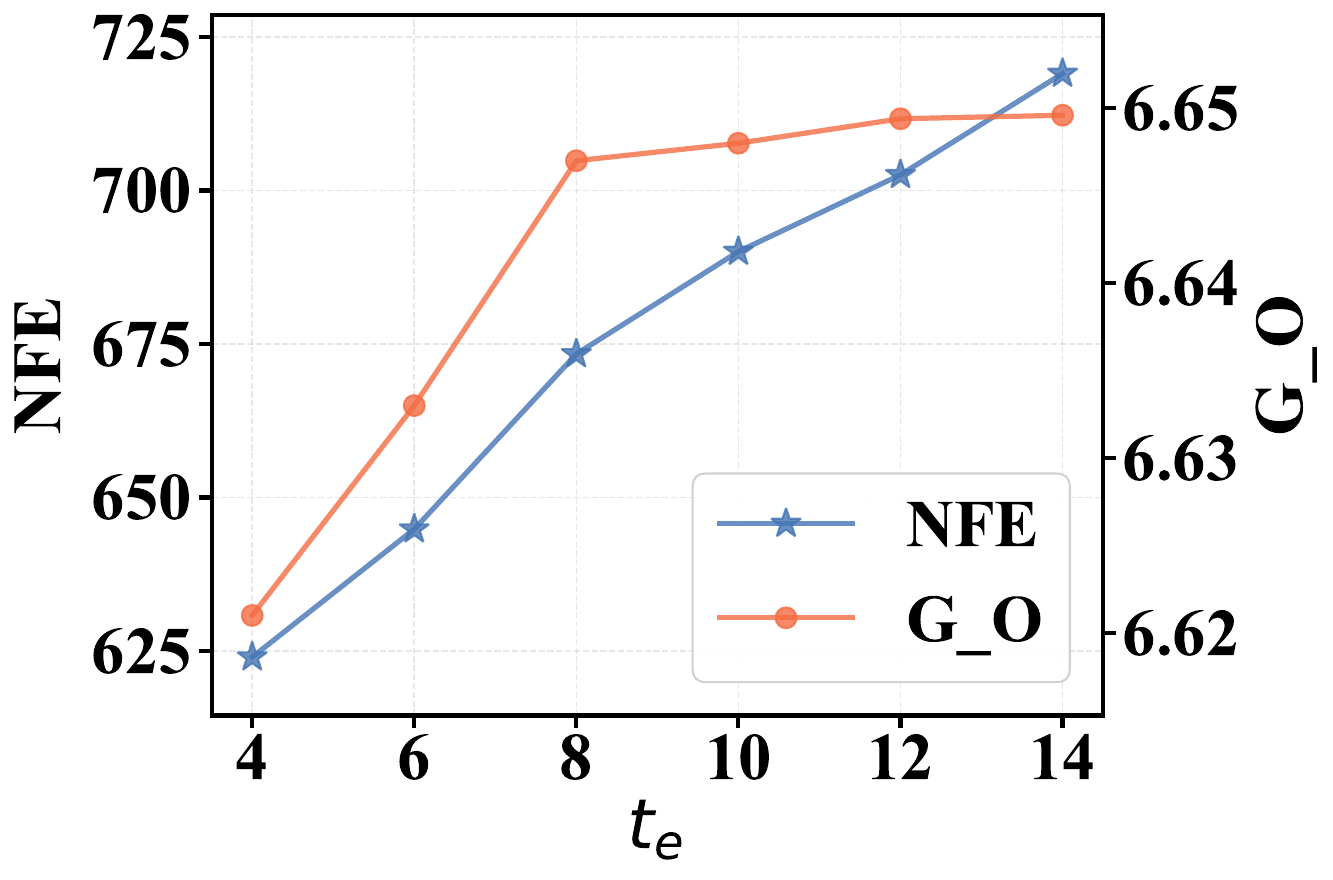}}
    \hfill
    \subfloat[retain step $t_l$]{\includegraphics[width=0.235\textwidth]{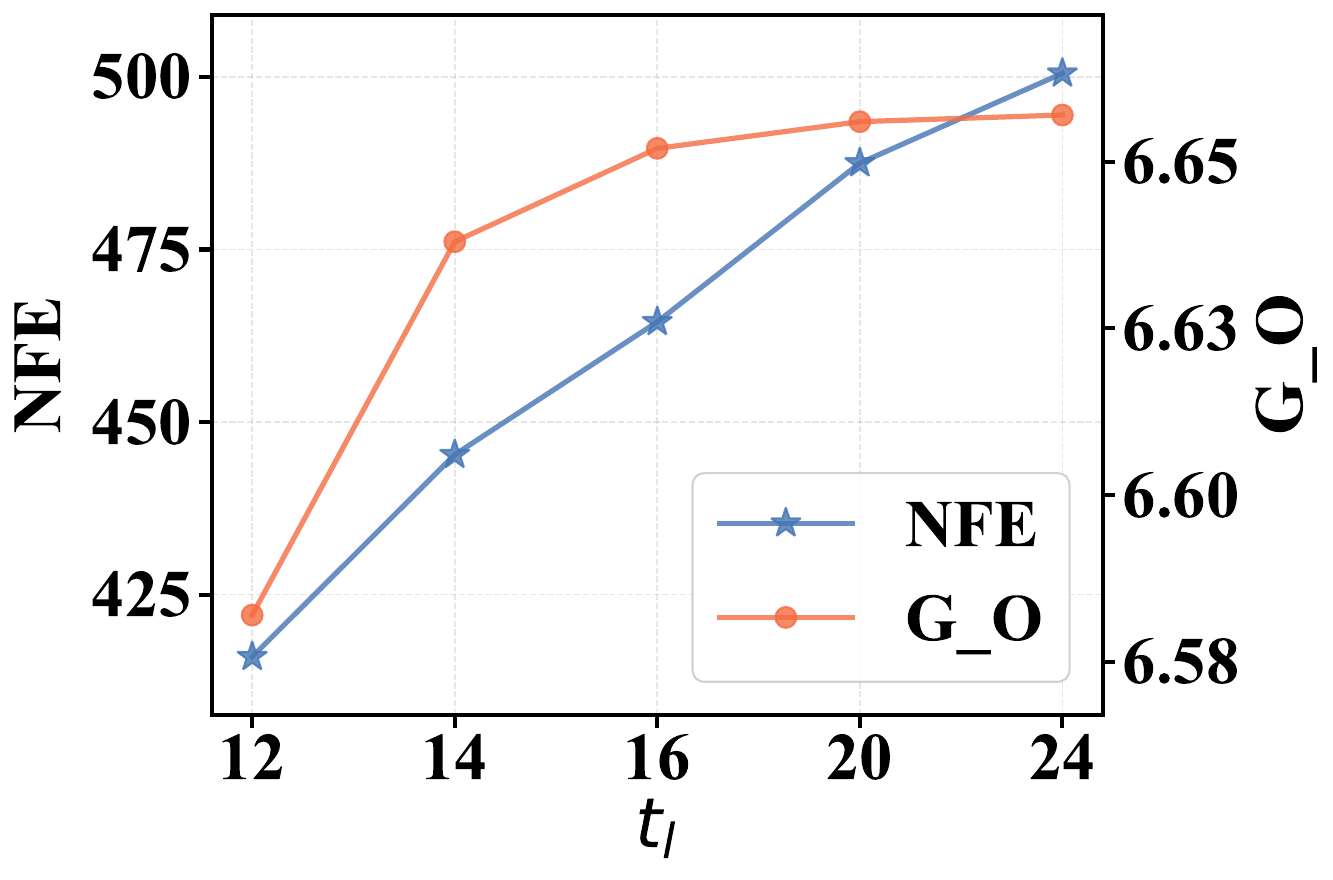}}
}
\vspace{-7pt}
\caption{\textbf{Effect of timesteps in early prune and late retain}.
}
\label{fig: hyper_result}
\vspace{-11pt}
\end{figure}
}

{
\setlength{\tabcolsep}{3.5 pt}
\renewcommand{\arraystretch}{1.01} 
\begin{table}[t]
    \caption{\textbf{Effect of MLLMs on performance and efficiency}.}
    \label{tab: ablation_mllm}
    \vspace{-0.6em}
    \centering
    \resizebox{1.0\linewidth}{!}
    {
    \scriptsize
\begin{tabular}{l | c | cc | cc | cc }
\toprule
\multirow{2}{*}{\textbf{Model}} & \multirow{2}{*}{\textbf{MLLM}} & \multicolumn{2}{c|}{\textbf{Kontext}} & \multicolumn{2}{c|}{\textbf{BAGEL}} & \multicolumn{2}{c}{\textbf{Step1X-Edit}}  \\ 
\cmidrule(lr){3-4} \cmidrule(lr){5-6} \cmidrule(lr){7-8}
 & & G\_O$\uparrow$ & NFE$\downarrow$ & G\_O$\uparrow$ & NFE$\downarrow$ & G\_O$\uparrow$ & NFE$\downarrow$ \\ 
\midrule 
BoN & \multirow{2}{*}{Qwen2.5-VL-72B~\cite{qwen25vl}} & 6.568 & 896 & 7.001 & 1600 & 7.155 & 896 \\
\textbf{ADE-CoT} & & \textbf{6.637} & \textbf{436} & \textbf{7.042} & \textbf{897} & \textbf{7.193} & \textbf{446}  \\
\midrule 
BoN & \multirow{2}{*}{Qwen-VL-MAX~\cite{qwen25vl}} & 6.641 & 896 & 6.908 & 1600 & 7.157 & 896 \\
\textbf{ADE-CoT}  &  & {\textbf{6.695}} & {\textbf{418}} & \textbf{6.972} & \textbf{882} & \textbf{7.196} & \textbf{434} \\
\midrule 
BoN & \multirow{2}{*}{Qwen3-VL-32B~\cite{qwen3}} & 6.691 & 896 & 7.034 & 1600 & 7.158 & 896  \\
\textbf{ADE-CoT} & & \textbf{6.719} & \textbf{403} & \textbf{7.109} & \textbf{806} & \textbf{7.240} & \textbf{414} \\
\bottomrule 
\end{tabular}
}
\vspace{-1.4 em}
\end{table}
}




\section{Conclusion}
\label{sec: conclusion}


In this work, we propose ADE-CoT, an on-demand test-time scaling algorithm to enhance quality and efficiency in image editing. 
Our difficulty-aware resource allocation adjusts computational budget based on edit difficulty, avoiding waste on simple edits.
To improve selection accuracy, we introduce edit-specific verification for early pruning, effectively finding high-potential candidates.
Finally, our depth-first opportunistic stopping mechanism terminates generation once intent-aligned results are found, reducing redundancy without quality drop. 
Extensive experiments on three SOTA editing models and three benchmarks show ADE-CoT achieves over a 2$\times$ speedup while maintaining performance.
We hope our work provides new insights into efficient test-time scaling for goal-directed generation.

{
    \small
    \bibliographystyle{ieeenat_fullname}
    \bibliography{main}
}

\newpage
\appendix

In this appendix, we provide comprehensive supplementary material to offer a more complete understanding of our ADE-CoT framework. The appendix is organized as follows: 
\begin{itemize}
    \item \cref{supp_sec: discuss_motivation} revisits and expands upon the motivation behind our work. 
    \item \cref{supp_sec: details_ADE_CoT} details the technical components of ADE-CoT. 
    \item \cref{supp_sec: extra_results} presents additional experimental results and ablation studies.    \item \cref{supp_sec: limitation_and_future} outlines the limitations and directions for future research. 
    \item \cref{supp_sec: extended_related_work} reviews extended related work in the field of test-time scaling and image editing.
\end{itemize}
Detailed contents are listed as follows:

\setlength{\cftbeforesecskip}{0.5em}
\cftsetindents{section}{0em}{1.8em}
\cftsetindents{subsection}{1em}{2.5em}
\etoctoccontentsline{part}{Appendix}
{
  \etocsettocstyle{}{}
  \localtableofcontents
}

\section{Further Discussion on Motivation}
\label{supp_sec: discuss_motivation}

\subsection{Limitations of SOTA Models on Complex Edits}
\label{supp_sec: limitation_on_complex_edit}

Recent image editing models~\cite{Step1X_Edit_arxiv2025, Bagel_2025_arxiv, FLUX_Kontext_arxiv2025, qwen_image_arxiv2025} based on latent-level fusion between MLLMs and diffusion decoders have demonstrated impressive capabilities on standard editing tasks. However, their performance remains challenging when faced with complex editing scenarios: 

\textbf{Large pose changes}. As shown in~\cref{supp_fig: baselin_w_ADE_CoT}, baseline models (\ie, default single-pass inference setting) struggle with edits requiring significant pose or action modifications. For instance, when asked to ``\texttt{change the man's gesture to raising his hands}'', the baseline unintentionally alters the surrounding context, including the position of the chair and the man's location in the scene. Similarly, the instruction ``\texttt{make the action of the plane to taking off}'' results in the baseline replacing the original plane with a completely different aircraft model and color scheme. The instruction ``\texttt{change the bird's action to flapping its wings and flying}'' often yields anatomically incorrect wing positions. 

\textbf{Multi-object modification}. Complex edits involving multiple objects pose additional challenges for baseline models. In~\cref{supp_fig: baselin_w_ADE_CoT}, we observe failures in instructions such as ``\texttt{remove the woman standing next to the lady in white}'', where the model fails to remove the correct person and leaves visible artifacts. Similarly, ``\texttt{remove the foreground snow-covered trees}'' results in incomplete removal, with only some trees being eliminated while others remain.

\textbf{Fine-grained regional edits}. 
Baseline models often fail to perform precise localized modifications. As shown in~\cref{supp_fig: baselin_w_ADE_CoT}, instructions such as ``\texttt{change clothes of the person in lower right corner to green}'' frequently result in incorrect region selection and color bleeding to adjacent areas. The edit ``\texttt{change the hair of the green-eyed toy figure to blonde}'' shows similar issues with precise attribute modification. Fine-grained color changes such as ``\texttt{alter the color of the front bus to lime}'' or ``\texttt{change the color of couch to yellow}'' commonly affect unintended areas.

\textbf{Multi-turn editing}. 
As illustrated in~\cref{supp_fig: baselin_w_ADE_CoT_multi_turn}, multi-turn edits are particularly susceptible to cascading errors, where mistakes in early turns propagate and accumulate through subsequent editing steps. For instance, in the first example, the initial instruction is ``\texttt{Turn 1: Change the pants color to black.}'' However, the model fails to execute this correctly, leaving the pants unchanged. This initial mistake cascades: subsequent edits are performed on an already flawed image. Consequently, the final result fails to reflect the cumulative user intent.

These limitations demonstrate that single-pass inference is insufficient for complex editing scenarios.

{
\begin{figure*}[htbp]
\begin{center}
\includegraphics[width=0.95\linewidth]{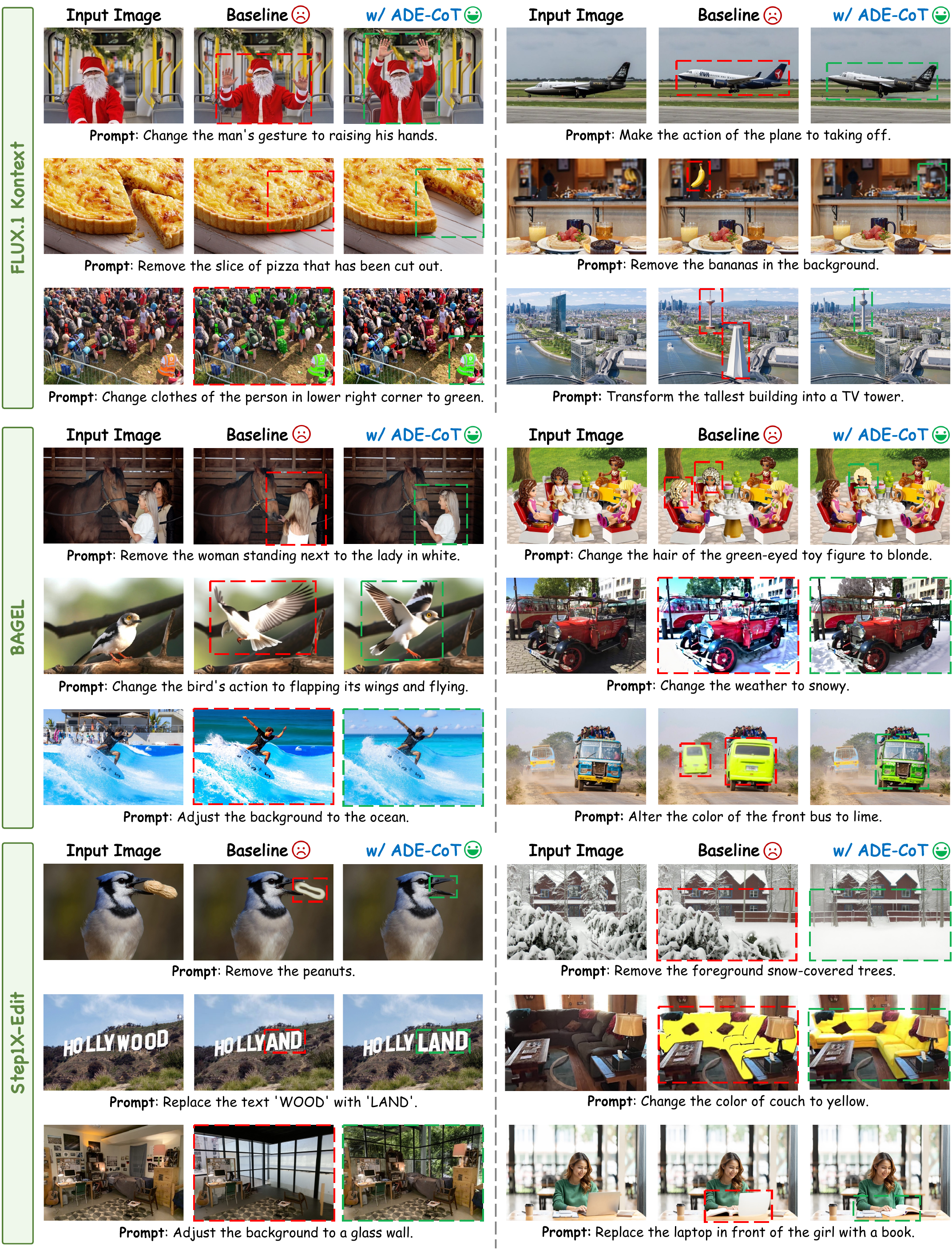}
\end{center}
\vspace{-1.7em}
\caption{
\textbf{Qualitative comparison on complex edits.}
We compare three SOTA editing models (FLUX.1 Kontext~\cite{FLUX_Kontext_arxiv2025}, BAGEL~\cite{Bagel_2025_arxiv}, and Step1X-Edit~\cite{Step1X_Edit_arxiv2025}) on challenging edits (large pose changes, multi-object modifications, and fine-grained regional edits). 
Baseline models often fail, while our ADE-CoT produces correct results via adaptive test-time scaling. \faSearch~\textbf{Zoom in} for detailed view.
}
\label{supp_fig: baselin_w_ADE_CoT}
\vspace{-1.6em}
\end{figure*}
}

{
\begin{figure*}[htbp]
\begin{center}
\includegraphics[width=0.99\linewidth]{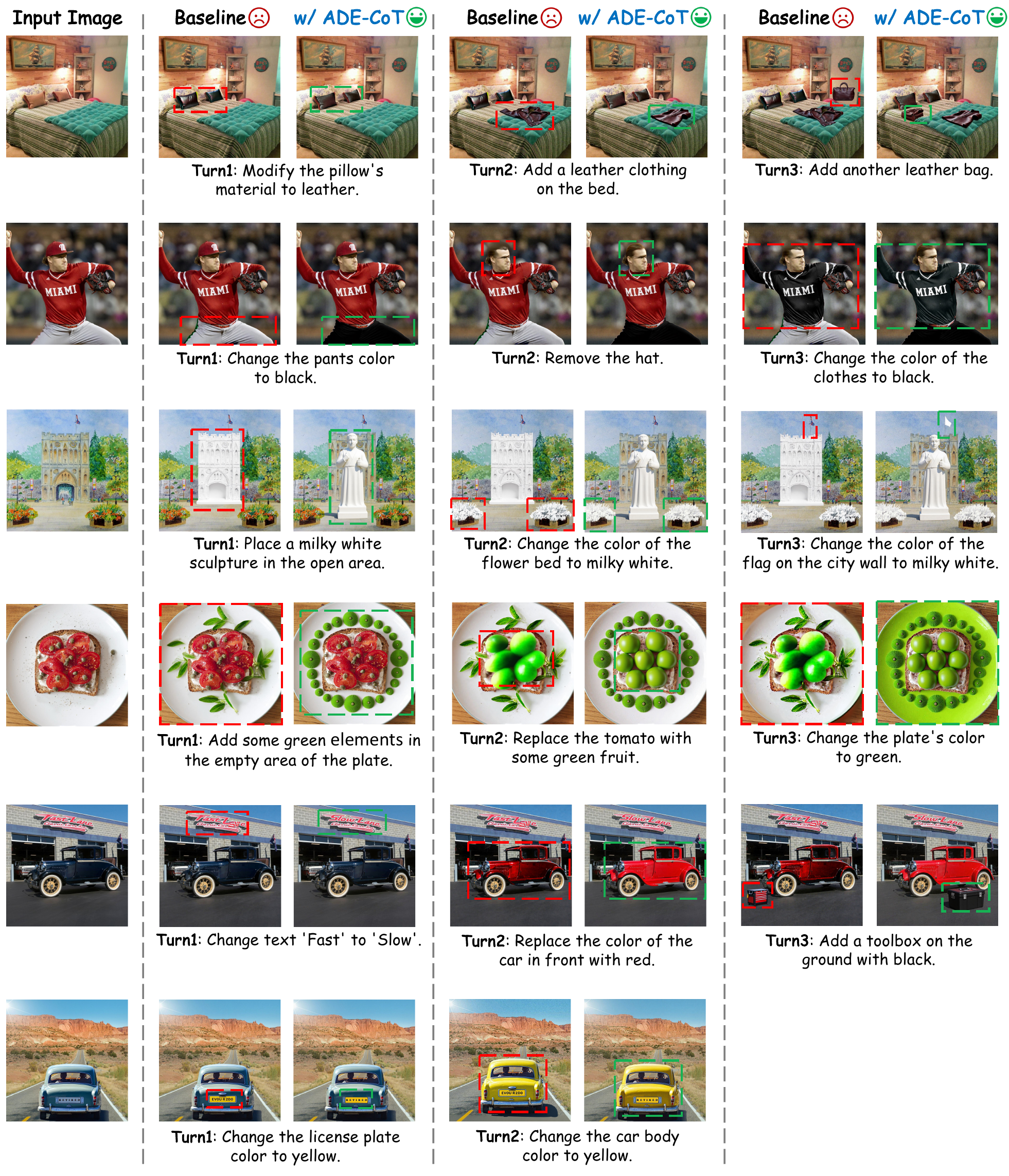}
\end{center}
\vspace{-1.7em}
\caption{\textbf{Qualitative comparison on multi-turn edits}. The baseline model often fails to preserve the context from previous edits, leading to accumulated errors in subsequent turns. Our ADE-CoT maintains consistency across multiple sequential instructions, producing correct final images that reflect all requested changes. \faSearch~\textbf{Zoom in} for detailed view. 
}
\label{supp_fig: baselin_w_ADE_CoT_multi_turn}
\end{figure*}
}

\subsection{Image Chain-of-Thought Methods}

Image Chain-of-Thought (Image-CoT)~\cite{Image_CoT, TTS_SANA, TTS_Baseline} offers a promising approach to address these challenges. As a test-time scaling strategy, Image-CoT generates multiple candidates through extended inference time. By selecting the best one from diverse candidates, it improves editing quality on complex scenarios without requiring additional training.

\begin{algorithm*}[htbp]
\caption{\textit{Best-of-N} (\texttt{BoN}) algorithm for image editing.}
\label{alg: BoN}
\begin{algorithmic}[1]
\REQUIRE source image $I_{\text{src}}$, text prompt $c$, number of samples $N$, and total steps $T$
\STATE $\mathcal{U} \leftarrow \{ \}$ \quad $\triangleright$ Initialize empty set for $(I, S)$ pairs
\FOR{$i = 1$ to $N$} 
    \STATE $x^{(i)}_T \sim \mathcal{N}(0, \mathrm{I})$ \quad $\triangleright$  Sample initial noise
    \STATE $c^{(i)} \leftarrow \texttt{Rewrite}(c)$ \quad $\triangleright$ (optional) Rewrite prompt 
    \STATE $x_0^{(i)} \leftarrow \texttt{Sampler}(I_{\text{src}}, x^{(i)}_T, c^{(i)}, T, 0)$   \quad \textcolor{method_purple}{$\triangleright$ {Sample from $T$ to $0$, \ie, full denoising process}}
    \STATE $I^{(i)} \leftarrow \texttt{VAE\_Decoder}(x_0^{(i)})$ 
    \STATE $S^{(i)} \leftarrow \texttt{Vrf}_g(I_{\text{src}}, I^{(i)}, c)$ \quad $\triangleright$ Compute general MLLM score
    \STATE $\mathcal{U} \leftarrow \mathcal{U} \cup \{ (I^{(i)}, S^{(i)}) \}$ \quad $\triangleright$ Update set
\ENDFOR
\STATE $I^* \leftarrow \argmax_{(I, S) \in \mathcal{U}} S $ 
\RETURN $I^*$
\end{algorithmic}
\end{algorithm*}

\textbf{Best-of-N} (BoN)~\cite{TTS_Baseline, TTS_SANA} is the standard method for Image-CoT. As shown in Fig. 3(a) and summarized in Alg.~\ref{alg: BoN}, it consists of two primary stages: 
\ding{182} \textbf{Generation}. This stage produces a diverse set of $N$ candidate images. This is achieved through a loop that iterates $N$ times (Line 2). Within each iteration, diversity is introduced by sampling a unique initial noise $x_T^{(i)}$ (Line 3) and optionally rewriting the text prompt $c$ into a variant $c^{(i)}$ (Line 4). The core of this stage is the \texttt{Sampler} function (Line 5), which performs a complete denoising process from timestep $T$ down to $0$ to generate a clean latent representation $x_0^{(i)}$. This latent is then decoded into the final image $I^{(i)}$ (Line 6).
\ding{183} \textbf{Selection}. For each image $I^{(i)}$, a general MLLM verifier $\texttt{Vrf}_g$ computes a score $S^{(i)}$ that reflects its quality and adherence (Line 7). The $(I^{(i)}, S^{(i)})$ pair is added to the set $\mathcal{U}$ (Line 8). After all $N$ candidates are generated, the candidate with the highest score is chosen as the final output $I^*$ (Line 10). 
\textbf{Computational cost}. While BoN improves generation quality, its computational cost scales linearly with the number of samples $N$. Since every candidate must complete the full $T$ denoising steps before selection, the total cost is $N \times T$ function evaluations (NFE). This inefficiency makes BoN impractical for large-scale sampling.

\begin{algorithm*}[htbp]
\caption{\textit{Early Pruning} algorithm for image editing.}
\label{alg: BoN_w_early_filter}
\begin{algorithmic}[1]
\REQUIRE source image $I_{\text{src}}$, text prompt $c$, number of samples $N$ and steps $T$, early step $t_e$, reject threshold $S_{\mathrm{rj}}$, \texttt{mode} $\in$ \{`additional\_steps', `intermediate\_state'\}
\STATE $\mathcal{U} \leftarrow \{ \}$ \quad $\triangleright$ Initialize empty set for $(I, S)$ pairs
\FOR{$i = 1$ to $N$} 
    \STATE $x^{(i)}_T \sim \mathcal{N}(0, \mathrm{I})$ \quad $\triangleright$  Sample initial noise
    \STATE $c^{(i)} \leftarrow \texttt{Rewrite}(c)$ \quad $\triangleright$ (optional) Rewrite prompt 
    \IF{\texttt{mode} == `additional\_steps'}
        \STATE \textcolor{method_purple}{$\triangleright$ TTS-EF~\cite{ICEdit_arxiv2025} method }
        \STATE $x^{(i)}_{t_e} \leftarrow \texttt{Sampler} (I_{\text{src}}, x^{(i)}_T, c^{(i)}, t_e, 0 )$ \quad \textcolor{method_purple}{$\triangleright$ Sample from $t_e$ to $0$, \ie, early preview by additional denoising steps}
    \ELSIF{\texttt{mode} == `intermediate\_state'}
        \STATE \textcolor{method_purple}{$\triangleright$ PRM~\cite{Image_CoT} and PARM~\cite{Image_CoT} methods } 
         \STATE $x^{(i)}_{t_e} \leftarrow \texttt{Sampler} (I_{\text{src}}, x^{(i)}_T, c^{(i)}, T, t_e )$ \quad \textcolor{method_purple}{$\triangleright$ Sample from $T$ to $t_e$, \ie, partially denoising process}
    \ENDIF
    \STATE $I^{(i)}_{t_e} \leftarrow \texttt{VAE\_Decoder} (x^{(i)}_{t_e})$
    \STATE $S_{t_e}^{(i)} \leftarrow \texttt{Vrf}_g(I_{\text{src}}, I^{(i)}_{t_e}, c)$ \quad $\triangleright$ Compute general MLLM score 
    \STATE \textcolor{method_purple}{$\triangleright$ Prune sample below $S_{\mathrm{rj}}$}
    \IF{$S^{(i)}_{t_e} >= S_{\mathrm{rj}}$} 
        \IF{\texttt{mode} == `additional\_steps'}
            \STATE $x^{(i)}_0 \leftarrow \texttt{Sampler}(I_{\text{src}}, x_T^{(i)}, c^{(i)}, T, 0)$  \quad \textcolor{method_purple}{$\triangleright$ Sample from $T$ to $0$, \ie, full denoising process}
        \ELSIF{\texttt{mode} == `intermediate\_state'} 
            \STATE $x^{(i)}_0 \leftarrow \texttt{Sampler}(I_{\text{src}}, x_{t_e}^{(i)}, c^{(i)}, t_e, 0)$  \quad \textcolor{method_purple}{$\triangleright$ Sample from $t_e$ to $0$, \ie, resume denoising process}
        \ENDIF
        \STATE $I^{(i)} \leftarrow \texttt{VAE\_Decoder} (x_0^{(i)})$ 
        \STATE $S^{(i)} \leftarrow \texttt{Vrf}_g(I_{\text{src}}, I^{(i)}, c)$ \quad $\triangleright$ Compute general MLLM score 
        \STATE $\mathcal{U} \leftarrow \mathcal{U} \cup \{ (I^{(i)}, S^{(i)})  \}$ \quad $\triangleright$ Update set
    \ENDIF 
\ENDFOR
\STATE $I^* \leftarrow \argmax_{(I, S) \in \mathcal{U}} S $ 
\RETURN $I^*$
\end{algorithmic}
\end{algorithm*}

\textbf{Early pruning strategies in Image-CoT}. 
To address the inefficiency of BoN, early pruning~\cite{ICEdit_arxiv2025, Image_CoT, Video_TTS} is a standard approach. The core idea is to identify and discard low-potential candidates at an early stage, avoiding the cost of their full generation. A unified framework for two common strategies is shown in Alg.~\ref{alg: BoN_w_early_filter}, controlled by the \texttt{mode}.
\ding{182} \textbf{Early preview via additional denoising steps}. This variant, proposed by TTS-EF~\cite{ICEdit_arxiv2025}, generates preview images by performing additional denoising from $t_e$ to $0$ (Line 7). The \texttt{Sampler} function produces a complete denoised latent $x_{t_e}^{(i)}$ that is decoded into a clear preview image $I_{t_e}^{(i)}$ (Line 12). This provides high-quality previews for reliable verification. However, it introduces extra computational cost of $t_e$ denoising steps per candidate. After pruning, passing candidates resume denoising denoising from $T$ to $0$ to generate the final output (Line 17).
\ding{183} \textbf{Early pruning on intermediate states}.  This variant, used by PRM~\cite{Image_CoT} and PARM~\cite{Image_CoT}, samples from $T$ to $t_e$ to obtain an intermediate latent $x_{t_e}^{(i)}$ (Line 10). The latent is directly decoded into a preview image. This approach requires no extra steps for preview generation. After pruning, passing candidates resume denoising from $t_e$ to $0$ to complete generation (Line 19). However, the preview may be noisy due to incomplete denoising, which may affect verification accuracy.

\subsection{Issues of Image-CoT Methods for Editing}

Most Image-CoT methods~\cite{TTS_Baseline, Image_CoT, ICEdit_arxiv2025, Reflection_DiT} are developed for text-to-image generation. However, directly applying them to editing is suboptimal. As discussed in the Introduction, this mismatch causes three issues: 


\textbf{Inefficient resource allocation}. 
Image-CoT methods typically use a fixed sampling budget for all edits, which may be inefficient. To validate this, we measure the performance gain relative to edit difficulty. First, we defined edit difficulty by generating a single image for each editing instance and using its MLLM score as an initial score. We then grouped the edit tasks into bins based on this initial score. For all tasks, we applied a standard Best-of-$32$ (BoN) sampling strategy. We calculated the average score gain between the initial score and the final score after adding Image-CoT for each bin. As shown in~\cref{supp_fig: score_gain_after_image_cot}, edits with high initial scores (simple edits) showed minimal improvement from large-scale sampling. In contrast, edits with low initial scores (complex edits) benefited significantly. This confirms that a fixed budget wastes computational resources on simple edits that do not require extensive sampling.

{
\begin{figure*}[htbp]
\begin{center}
\includegraphics[width=0.85\linewidth]{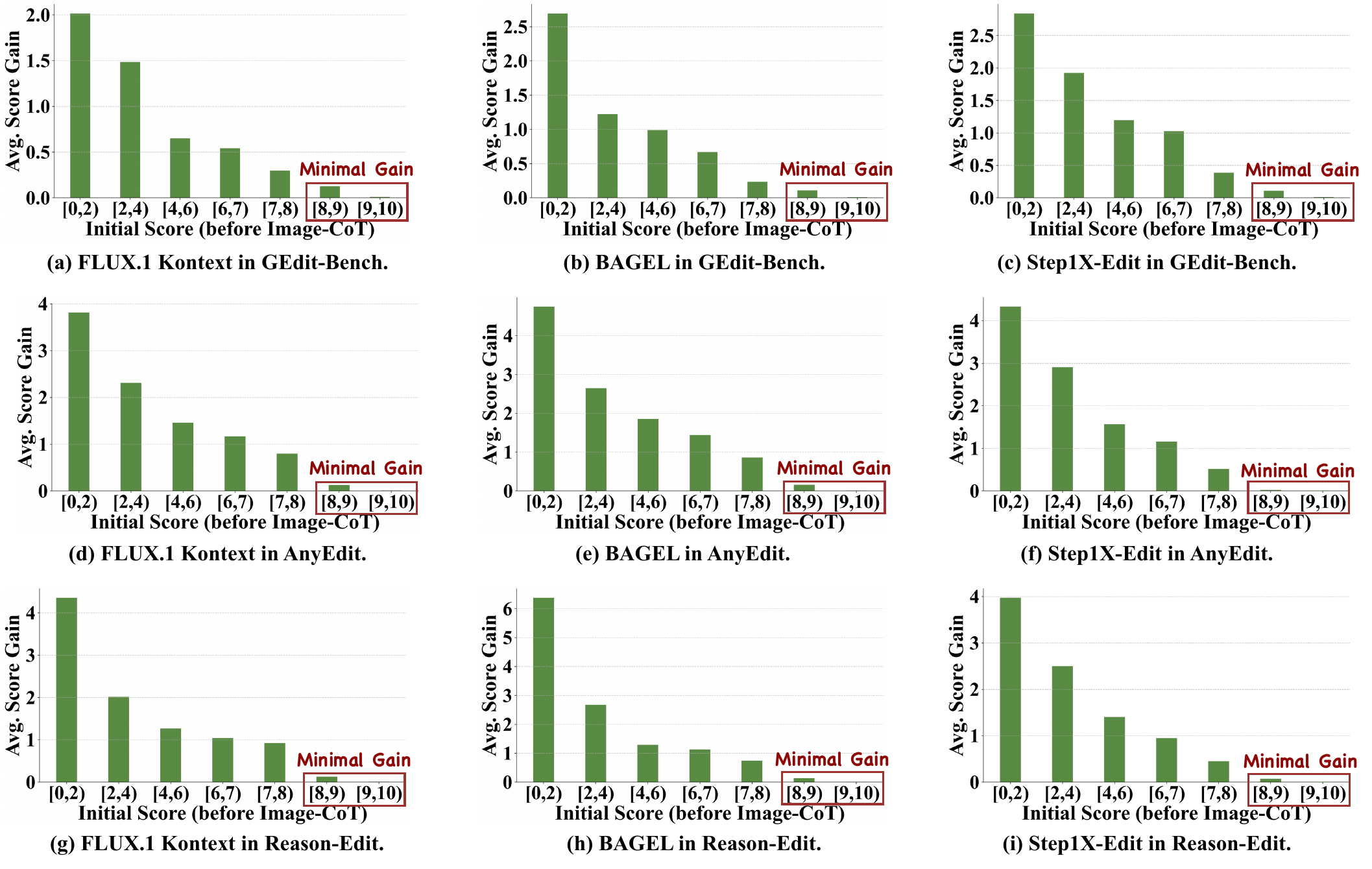}
\end{center}
\vspace{-1.7em}
\caption{\textbf{Inefficient resource allocation with fixed sampling budgets}. This figure extends the analysis from Fig.~2(a) to three SOTA models (FLUX.1 Kontext, BAGEL, and Step1X-Edit) and three benchmarks (GEdit-Bench, AnyEdit, and Reason-Edit). Edits with high initial scores (\textbf{\textcolor{removed}{red boxes}}, [8,9) and [9,10)) show minimal improvement across all three models and three benchmarks. Edits with low initial scores (indicating complex tasks) benefit significantly from Image-CoT, achieving substantial performance gains. This demonstrates that fixed sampling budgets waste computation on simple edits, motivating our difficulty-aware resource allocation strategy.}
\vspace{-1.0em}
\label{supp_fig: score_gain_after_image_cot}
\end{figure*}
}

\textbf{Unreliable early-stage verification}. 
Current Image-CoT methods~\cite{ICEdit_arxiv2025, Image_CoT, Video_TTS} use general MLLM scores to prune candidates at early denoising stages. We examine whether these scores correctly identify high-potential candidates. In our experiment, we generate $N=32$ candidates per edit case and evaluate each at an early timestep $t_e = 8$ using VIE-Score~\cite{VIE_Score}. Candidates scoring below a rejection threshold $S_{\text{rj}}$ are \textit{pruned}. We then complete the full denoising process for all candidates to obtain final scores. As shown in~\cref{supp_fig: misjudgement_in_pruned_samples}, this misjudgement occurs consistently across all tested models and datasets. On average, 40\% of the pruned samples ultimately achieve high final scores ($\geq 6$). This indicates that general scores incorrectly discard many high-potential candidates during early pruning. This misjudgement leads to degraded final performance.

{
\begin{figure*}[htbp]
\begin{center}
\includegraphics[width=0.9\linewidth]{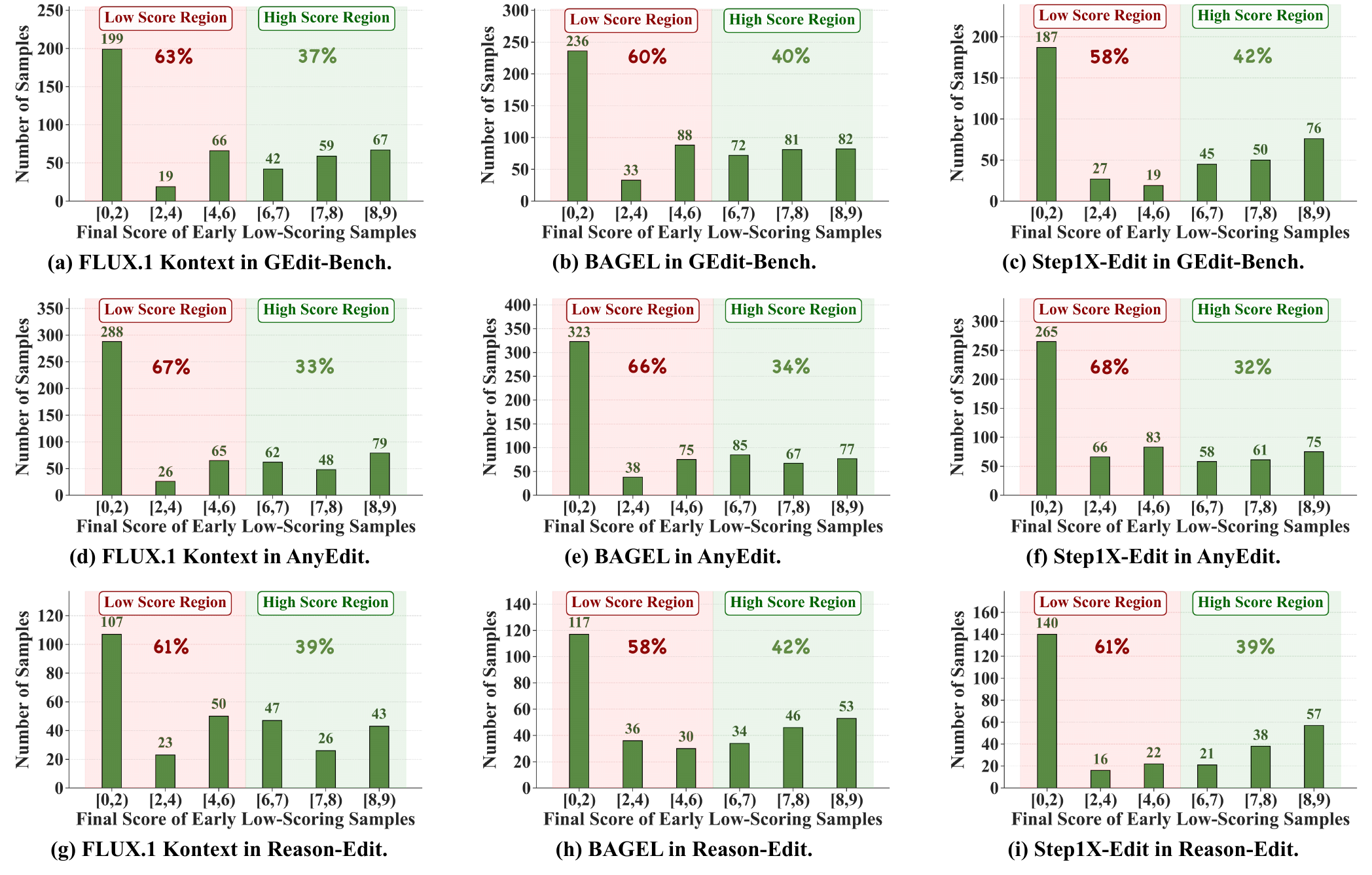}
\end{center}
\vspace{-1.7em}
\caption{\textbf{Misjudgement by general MLLM scores in early pruning.} This figure extends the analysis from Fig.~2(b) to three SOTA editing models (FLUX.1 Kontext, BAGEL, and Step1X-Edit) evaluated on three benchmarks (GEdit-Bench, AnyEdit-Test, and Reason-Edit).  Samples with low early scores are categorized by their final scores (x-axis): low score region (\textbf{\textcolor{removed}{red}}, [0,6)) and high score region (\textcolor{method_green}{\textbf{green}}, [6,9)). On average, 37\% of early low-scoring samples eventually achieve high final scores, yet would be incorrectly discarded by general MLLM scores. This demonstrates unreliable early-stage verification in editing, motivating our edit-specific verification strategy.}
\vspace{-0.3em}
\label{supp_fig: misjudgement_in_pruned_samples}
\end{figure*}
}

\textbf{Redundant edited results}. 
The goal-directed nature of image editing suggests that large-scale sampling may produce many similar, correct results. To quantify this redundancy, we apply a Best-of-$32$ (BoN) strategy to each editing instance. For each case, we identify the best score achieved among the 32 candidates and then count how many candidates share the same best score. As shown in~\cref{supp_fig: high_redundancy}, we observe this redundancy across three models and three datasets. For edit cases with high final scores (\eg, in the range $[7, 9)$), a large number of candidates, often more than 8, achieve the identical best score. Since only one intent-aligned result is sufficient for editing, this redundancy reflects unnecessary computation. Existing breadth-first search strategies~\cite{TTS_Baseline, Image_CoT, ICEdit_arxiv2025} generate all candidates before selection. This leads to wasted denoising steps on redundant correct outputs.


{
\begin{figure*}[htbp]
\begin{center}
\includegraphics[width=0.99\linewidth]{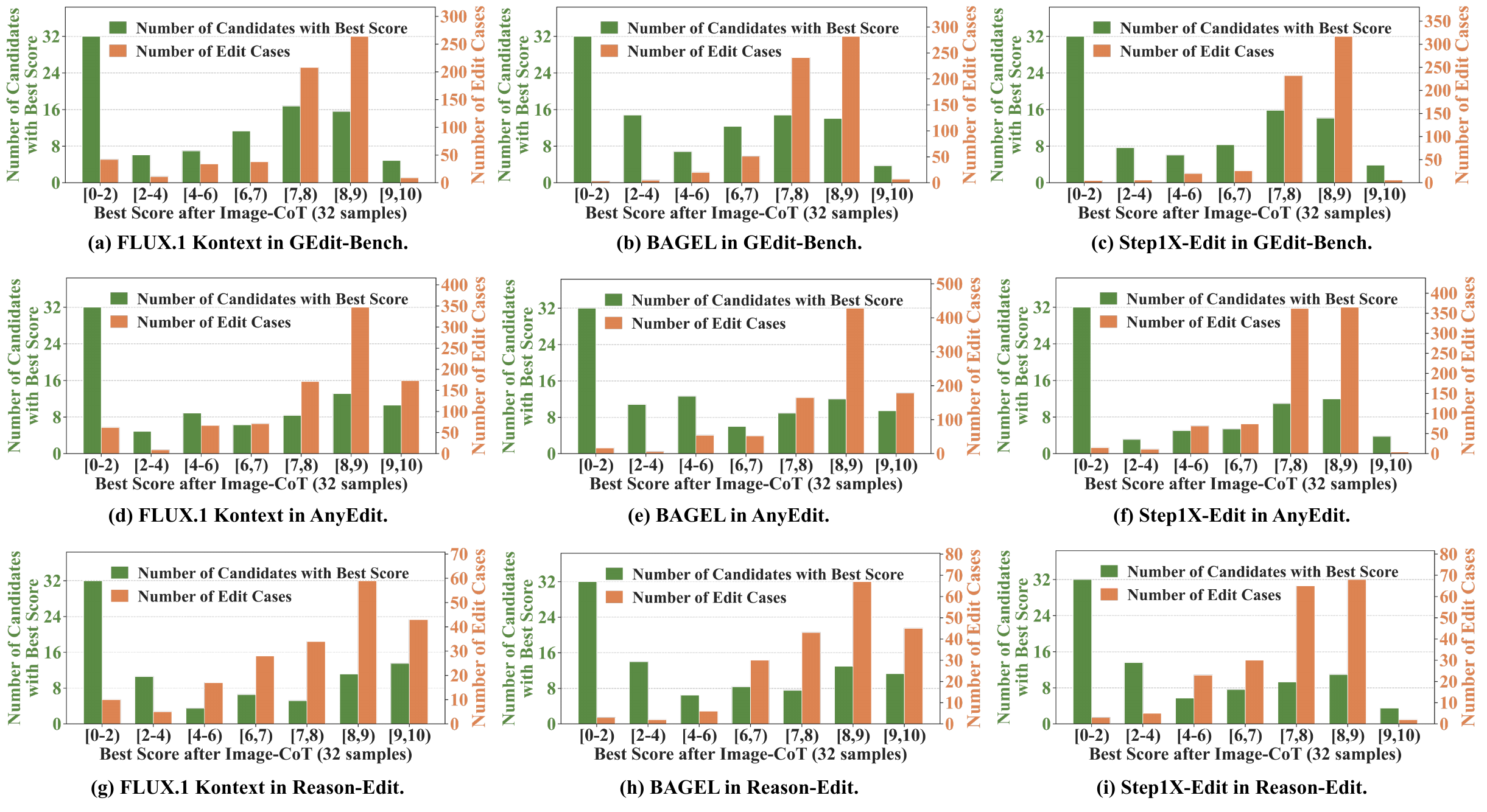}
\end{center}
\vspace{-1.7em}
\caption{\textbf{Redundant edited results in large-scale sampling.} This figure extends the analysis from Fig.~2(c) to three SOTA editing models (FLUX.1 Kontext, BAGEL, and Step1X-Edit) evaluated on three benchmarks (GEdit-Bench, AnyEdit-Test, and Reason-Edit). For most edit cases, a large number of candidates (green bars, left y-axis) share identical best scores. The number of edit cases exhibiting such redundancy (orange bars, right y-axis) increases significantly in high score regions, particularly in [7,8) and [8,9) (x-axis). This demonstrates that Image-CoT produces redundant correct outputs in goal-directed editing, motivating our opportunistic stopping strategy.}
\vspace{-0.5em}
\label{supp_fig: high_redundancy}
\end{figure*}
}

To address these issues, we propose ADE-CoT, an adaptive test-time scaling framework for image editing. Our method improves editing performance while maintaining computational efficiency. Specifically, we introduce three key strategies: (1) difficulty-aware resource allocation to dynamically adjust sampling budgets based on edit difficulty, (2) edit-specific verification to accurately identify high-potential candidates during early pruning, and (3) depth-first opportunistic stopping to reduce redundant computation on correct results. As shown in~\cref{supp_fig: baselin_w_ADE_CoT} and~\cref{supp_fig: baselin_w_ADE_CoT_multi_turn}, our method significantly enhances baseline performance on complex editing scenarios discussed in~\cref{supp_sec: limitation_on_complex_edit}.

\newpage

\section{Details of the Proposed Method ADE-CoT}
\label{supp_sec: details_ADE_CoT}

In this section, we provide implementation details of three components in ADE-CoT: difficulty-aware resource allocation (\cref{supp_sec: adapt_sample}), edit-specific verification in early pruning (\cref{supp_sec: prune_and_rank}), and depth-first opportunistic stopping (\cref{supp_sec: adapt_stop}).


\begin{algorithm*}[t]
\caption{Our \textit{\textbf{AD}aptive \textbf{E}dit-\textbf{CoT}} (\texttt{ADE-CoT}) algorithm for image editing.}
\label{alg: ADE_CoT}
\begin{algorithmic}[1]
\REQUIRE source image $I_{\text{src}}$, text prompt $c$, number of samples $N$ and steps $T$, early step $t_e$ and retain step $t_l$
\STATE \textcolor{method_orange}{$\triangleright$ \textbf{Adapt $N$ by edit difficulty}} 
\STATE {$N_{a} \leftarrow \texttt{Adapt\_Num}(I_{\text{src}}, c, N, T)$} 
\STATE \textcolor{method_blue}{$\triangleright$  \textbf{Sample from $T$ to $t_e$ and prune samples}} 
\STATE $\mathcal{X}_{t_e}, \mathcal{C} \leftarrow \texttt{Early\_Prune}(I_{\text{src}}, c, N_{a}, T, t_e)$ 
\STATE \textcolor{method_green}{$\triangleright$ \textbf{Sample from $t_e$ to $t_l$ and retain top results; Sample from $t_l$ to $0$ and select intent-aligned results}} 
\STATE $\mathcal{U} \leftarrow \texttt{Adaptive\_Stop} (I_{\text{src}}, c, \mathcal{X}_{t_e}, \mathcal{C}, t_e, t_l ) $
\STATE $I^* \leftarrow \argmax_{(I, S) \in \mathcal{U}} S $
\RETURN $I^*$
\end{algorithmic}
\end{algorithm*}

\subsection{Difficulty-aware Resource Allocation}
\label{supp_sec: adapt_sample}

As described in Sec.~3.1 of the main paper, our difficulty-aware resource allocation strategy dynamically adjusts the sampling budget to improve computational efficiency. The process is summarized in Alg.~\ref{alg: AdaptNum}. It first generates a single candidate to estimate the edit difficulty, which then determines the final sampling budget, $N_a$.

\ding{182} The process begins by generating one preliminary-estimation image (Lines 1-3). First, we sample an initial noise vector $x_T$ from a standard normal distribution (Line 1). A diffusion \texttt{Sampler} then generates a clean latent $x_0$ from this noise, conditioned on the source image $I_{\text{src}}$ and instruction $c$ (Line 2). A VAE decoder subsequently converts the latent $x_0$ into a pixel-space image $I$ (Line 3). 

\ding{183} Next, we estimate the edit difficulty using this initial image (Line 4). A general MLLM verifier, \texttt{Vrf$_g$}, evaluates the image $I$ to produce an initial score $S$. This score acts as a proxy for difficulty, where a high score suggests an easy edit and a low score indicates a difficult one. 

\ding{184} The adaptive budget $N_a$ is then calculated based on this score (Line 5), following Eq.~3 from the main text. For an easy edit where the score $S$ is high, the budget $N_a$ is reduced towards the minimum budget $N_{\text{min}}$. Conversely, for a difficult edit where $S$ is low, the budget $N_a$ increases towards the original budget $N$. The hyperparameter $\gamma$ controls the sensitivity of this adjustment, and the ceiling function $\lceil\cdot\rceil$ ensures the budget is an integer. Finally, the algorithm returns the calculated budget $N_a$ (Line 6). This strategy effectively allocates more computational resources to difficult cases and saves them on easy ones.

\begin{algorithm*}[htbp]
\caption{\texttt{AdaptNum} algorithm - Lines 1-2 of Alg.~\ref{alg: ADE_CoT}.}
\label{alg: AdaptNum}
\begin{algorithmic}[1]
\REQUIRE source image $I_{\text{src}}$, text prompt $c$, number of samples $N$ and steps $T$
\REQUIRE number of minimum samples $N_{\mathrm{min}}$, maximum possible score $S_{\mathrm{max}}$, sensitivity factor $\gamma$
\STATE $x_T \sim \mathcal{N}(0, \mathrm{I})$ 
\STATE $x_0 \leftarrow \texttt{Sampler}(I_{\text{src}}, x_T, c, T, 0 ) $  \quad $\triangleright$ Sample from $T$ to 0
\STATE $I \leftarrow \texttt{VAE\_Decoder} (x_0)$ 
\STATE $S \leftarrow \texttt{Vrf}_{g}(I_{\text{src}}, I, c)$ \quad $\triangleright$ Compute general MLLM score 
\STATE $N_a \leftarrow N_{\mathrm{min}} + \lceil(N - N_{\mathrm{min}}) \times (1 -  S / S_{\mathrm{max}})^\gamma \rceil $  \quad \textcolor{method_orange}{$\triangleright$ Adapt $N$ based on $S$ (\cf Sec~3.1 Eq.~3)}
\RETURN $N_a$
\end{algorithmic}
\end{algorithm*}

\subsection{Edit-specific Verification in Early Pruning}
\label{supp_sec: prune_and_rank}

As described in Sec.~3.2 of the main paper, our edit-specific verification strategy addresses the misjudgement issue of general MLLM scores in early denoising stages. The detailed process is summarized in Alg.~\ref{alg: Early_Prune}. 

The algorithm operates as follows. 
\ding{182} We first initialize empty sets for intermediate latents $\mathcal{X}_{t_e}$, prompts $\mathcal{C}$, and scores $\mathcal{S}_{t_e}$ (Line 1). 
\ding{183} For each of the $N_a$ samples (Lines 2-13), we sample random noise $x_T^{(i)} \sim \mathcal{N}(0, \mathrm{I})$ (Line 3) and optionally rewrite the prompt $c^{(i)}$ (Line 4). We then perform denoising from timestep $T$ to early timestep $t_e$ (Line 5) and apply the one-step preview mechanism to obtain $x_{0|t_e}^{(i)}$ (Line 6), which is decoded into preview image $I_{0|t_e}^{(i)}$ (Line 7). The unified score $S_{0|t_e}^{(i)}$ is computed by combining general MLLM score, edited-region correctness, and instruction-caption consistency (Line 8). Candidates with scores below the rejection threshold $S_{\text{rj}}$ are pruned (Lines 10-12), while others are retained. 
\ding{184} After processing all samples, we remove visually similar candidates using DINOv2 features and the threshold $\tau_{\text{sim}}$ (Lines 14-15). 
\ding{185} Finally, the remaining candidates are sorted by their unified scores in descending order (Lines 16-17), which guides the subsequent depth-first generation stage.

\begin{algorithm*}[htbp]
\caption{\texttt{Early\_Prune} algorithm - Lines 3-4 of Alg.~\ref{alg: ADE_CoT}.}
\label{alg: Early_Prune}
\begin{algorithmic}[1]
\REQUIRE source image $I_{\text{src}}$, text prompt $c$, number of samples $N_a$ and steps $T$, early step $t_e$ 
\REQUIRE reject threshold $S_{\mathrm{rj}}$, similarity threshold $\tau_{\mathrm{sim}}$
\STATE $\mathcal{X}_{t_e} \leftarrow \{\}, \mathcal{C} \leftarrow \{\}, \mathcal{S}_{t_e} \leftarrow \{\}$ \quad $\triangleright$ Initialize empty set 
\FOR{$i=1$ to $N_a$}
    \STATE $x^{(i)}_T \sim \mathcal{N}(0, \mathrm{I})$
    \STATE $c^{(i)} \leftarrow \texttt{Rewrite}(c)$ \quad $\triangleright$ (optional) Rewrite prompt
    \STATE $x^{(i)}_{t_e} \leftarrow \texttt{Sampler}(I_{\text{src}}, x^{(i)}_T, c^{(i)}, T, t_e) $ \quad \textcolor{method_blue}{$\triangleright$ Sample from $T$ to $t_e$}
    \STATE $x^{(i)}_{0 | t_e} \leftarrow \texttt{One\_Step\_Preview}(x^{(i)}_{t_e}, t_e) $   \quad \textcolor{method_blue}{$\triangleright$ Preview image from intermediate latent (\cf Sec~3.2 Eq.~4)}
    \STATE $I^{(i)}_{0|{t_e}} \leftarrow \texttt{VAE\_Decoder}(x^{(i)}_{0 | {t_e}})$
    \STATE $S^{(i)}_{0|{t_e}} \leftarrow \texttt{Vrf}(I_{\text{src}}, I^{(i)}_{0| {t_e} }, c )$ \quad $\triangleright$ Compute unified score (\cf Sec~3.2, Eq.~8)
    \STATE \textcolor{method_blue}{$\triangleright$ \textbf{Filter error by evaluated score}}
    \IF{$S^{(i)}_{0|{t_e}} >= S_{\mathrm{rj}}$}
        \STATE $\mathcal{X}_{t_e} \leftarrow \mathcal{X}_{t_e} \cup \{x_{t_e}^{(i)}\}, \mathcal{C}  \leftarrow \mathcal{C} \cup \{c^{(i)}\}, \mathcal{S}_{t_e} \leftarrow \mathcal{S}_{t_e} \cup \{ S^{(i)}_{0|{t_e}} \} $ \quad $\triangleright$ Update set
    \ENDIF
\ENDFOR
\STATE \textcolor{method_blue}{$\triangleright$ \textbf{Filter visually similar candidates}}
\STATE $\mathcal{X}_{t_e}, \mathcal{C}, \mathcal{S}_{t_e} \leftarrow \texttt{Remove\_Similar}(\mathcal{X}_{t_e}, \mathcal{C}, \mathcal{S}_{t_e}, \tau_{\mathrm{sim}}) $        
\STATE \textcolor{method_blue}{$\triangleright$ \textbf{Sort by evaluated score}}
\STATE $\mathcal{X}_{t_e}, \mathcal{C} \leftarrow \texttt{Sort\_by\_Score}(\mathcal{X}_{t_e}, \mathcal{C}, \mathrm{key}=\mathcal{S}_{t_e})$ \quad \textcolor{method_blue}{$\triangleright$ {Sort $\mathcal{X}_{t_e}, \mathcal{C}$ by $\mathcal{S}_{t_e}$ in descending order}}
\RETURN $\mathcal{X}_{t_e}, \mathcal{C}$
\end{algorithmic}
\end{algorithm*}

\subsubsection{One-Step Preview Mechanism}
\label{supp_sec: one_step_preview}

Lines 6-7 of Alg.~\ref{alg: Early_Prune} implement the one-step preview mechanism, which obtains approximate previews without additional denoising steps. To validate that these early previews reliably reflect the final output, we visualize the process in \cref{supp_fig: preview_image}. The figure compares the noisy latent $I_t$ with our corresponding one-step preview $I_{0|t}$ across various timesteps for three models. As shown, our preview generates a clear and high-fidelity approximation of the final result, even at very early stages where the noisy latent is uninterpretable (\eg, $t=8$). This observation confirms that our one-step preview provides a sufficiently clear signal for early-stage evaluation, enabling accurate pruning.

{
\begin{figure*}[htbp]
\begin{center}
\includegraphics[width=0.85\linewidth]{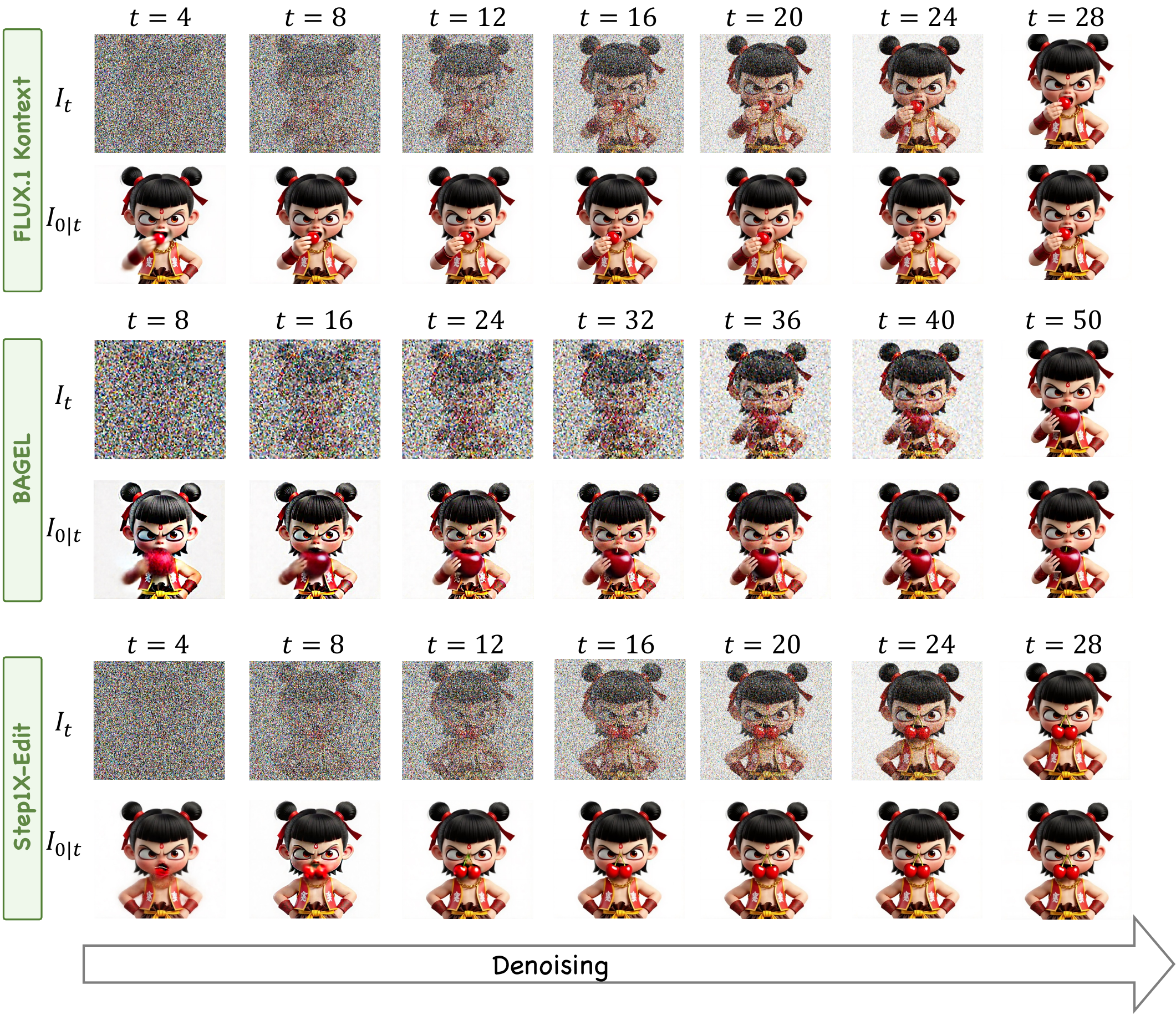}
\end{center}
\vspace{-1.7em}
\caption{\textbf{Effectiveness of the one-step preview mechanism}. We compare the noisy latent $I_t$ (top row for each model) with our corresponding one-step preview image $I_{0|t}$ (bottom row) at various denoising timesteps. Across all three models, our one-step preview generates a clear and high-fidelity approximation of the final output, even at very early stages (\eg, $t=8$). This demonstrates that the preview accurately reflects the final image's content and quality, providing a solid basis for our edit-specific verifiers.}
\vspace{-0.5em}
\label{supp_fig: preview_image}
\end{figure*}
}

\subsubsection{General Score by MLLM}
\label{supp_sec: verifier_general_score}

Following prior work~\cite{ICEdit_arxiv2025}, we use VIE-Score~\cite{VIE_Score} as our general MLLM verifier. VIE-Score evaluates an edited image based on two criteria: Semantic Consistency (SC), which measures instruction adherence and preservation of unedited regions, and Perceptual Quality (PQ), which assesses visual realism and aesthetics. The final overall score is calculated as the geometric mean of these two components: $S_{\text{gen}} = \sqrt{S_{\text{SC}} \times S_{\text{PQ}}}$.
While this general score provides a coarse-grained assessment, it struggles to detect subtle errors such as mislocalized edits or semantic misalignment in early denoising stages. To address this, we complement it with edit-specific metrics.

\subsubsection{Edited-Region Correctness}
\label{supp_sec: verifier_edited_region}

In~\cref{supp_prompt:region_object}, we present the prompt $P_{\text{reg}}$ used to identify the edited or kept objects. When the MLLM successfully identifies the edited object, the mask $M$ corresponds to that object's region. When the MLLM identifies the kept object, the mask $M$ is set to the inverted region (\ie, the complement of the kept object). If the MLLM cannot accurately determine either the edited or kept objects, we skip this verifier. Since directly computing RGB differences across the entire image is computationally expensive, we employ a sliding window approach to aggregate the change map $\Delta$. However, the obtained mask may not perfectly align with the true editing region. To address this, we apply an adaptive mask refinement strategy. If all early preview candidates yield $S_{\text{reg}} = 0$, we iteratively expand the mask $M$ by padding additional pixels around its boundary. The expansion process continues until at least one candidate achieves $S_{\text{reg}} > 0$. This ensures that the mask adequately covers the actual edited region. This refinement step improves the robustness of region correctness evaluation.

\begin{figure*}[t]
\centering
\begin{promptbox_blue}{Prompt $P_{\text{reg}}$ for identifying edited and keep objects}

\textbf{System Role:} You are an assistant in an image-editing pipeline. Your sole task is to determine, from a textual edit instruction and the original image, (1) which object(s) in the image must be edited and (2) which object(s) are explicitly required to remain unchanged (``keep'').

\vspace{0.5em}
\hrule
\vspace{0.5em}
\noindent\textbf{INPUT} \\
The user will always provide both keys:
\begin{itemize}[leftmargin=*, nosep]
    \item \texttt{original image} -- the image to be edited.
    \item \texttt{edit instruction} -- the user's textual instruction that specifies how the image should be edited.
\end{itemize}

\vspace{0.5em}
\hrule
\vspace{0.5em}
\noindent\textbf{OUTPUT (must follow exactly)} \\
Return one of the two following JSON structures:

\vspace{0.3em}
\noindent\textbf{Case A.} The \texttt{edit\_instruction} clearly identifies at least one concrete visual object to be edited:
\begin{verbatim}
{"edit_object": ["<the object(s) in the image that must be edited>"],
 "keep_object": ["<the object(s) that must NOT be edited>"]}
\end{verbatim}
\begin{itemize}[leftmargin=*, nosep]
    \item Both keys must be present.
    \item If no ``keep'' object is mentioned, output an empty array for \texttt{keep\_object}.
    \item If multiple distinct objects are explicitly requested, list each object as a separate string in the array.
    \item Regardless of whether the edit instruction is in English or Chinese, ALWAYS return object names in English.
\end{itemize}

\vspace{0.3em}
\noindent\textbf{Case B.} No concrete visual object is to be edited (ambiguous, global change, object absent, or the instruction is about modifying, adding, or deleting TEXT in the image):
\begin{verbatim}
{"edit_object": null,
 "keep_object": null}
\end{verbatim}
\vspace{0.5em}
\hrule
\vspace{0.5em}

\noindent\textbf{DECISION RULES}
\begin{itemize}[leftmargin=*, nosep]
    \item ``Clearly identified'' means the instruction contains explicit nouns or noun phrases that unambiguously map to elements in the image description (e.g., ``cat,'' ``red cup,'' ``man's face,'' ``background sky'').
    \item If the instruction applies to the entire image without pointing to a specific object (e.g., ``add a vintage filter,'' ``increase brightness''), return Case B.
    \item If the instruction references an object absent from the image description, return Case B.
    \item If the instruction's main purpose is to change, remove, or add TEXT in the image, return Case B (both values null).
    \item Do NOT perform the edit and do NOT add explanations---output the JSON only.
\end{itemize}

\vspace{0.5em}
\hrule
\vspace{0.5em}




\noindent\textbf{User Input:} \\
\texttt{edit instruction:} $\langle$edit\_prompt$\rangle$ \\
\texttt{original image:} $\langle$image$\rangle$
\end{promptbox_blue}
\vspace{-1.1em}
\caption{\textbf{Prompt $P_{\text{reg}}$ for identifying edited and keep objects}.}
\vspace{-0.5em}
\label{supp_prompt:region_object}
\end{figure*}

\subsubsection{Instruction-Caption Consistency} 
\label{supp_sec: verifier_caption_consistency}

In~\cref{supp_prompt:caption_generation}, we present the prompt $P_{\text{cap}}$, which instructs an MLLM to generate a target caption for the ideally edited image. However, for subtle edits like local object modifications, this caption-based score may not be sufficiently discriminative. To ensure the reliability of the generated caption, we introduce a two-stage filtering process. First, we confirm that the MLLM correctly understands the source image. We check if its generated \texttt{original\_caption} has a CLIP score above a threshold (default 0.27) with the source image. Second, we verify that the \texttt{edited\_caption} actually reflects a change. We ensure its textual similarity to the \texttt{original\_caption} is below a threshold (default 0.9). Only if a caption passes both these checks is it deemed reliable.


\begin{figure*}[t]
\centering
\begin{promptbox_blue}{Prompt $P_{\text{cap}}$ for generating output caption}

\textbf{System Role:} You are ``Dual-Caption,'' an expert vision-language assistant. Your job is to describe (a) the user-supplied original image, and (b) the image that would exist after perfectly applying the user's editing instruction.

\vspace{0.5em}
\hrule
\vspace{0.5em}

\noindent\textbf{INPUT} \\
The user will always provide both keys:
\begin{itemize}[leftmargin=*, nosep]
    \item \texttt{edit instruction}:
    \item \texttt{original image}:
\end{itemize}

\vspace{0.5em}
\hrule
\vspace{0.5em}

\noindent\textbf{OUTPUT (must follow exactly)} \\
Return a single JSON object with two keys, in this order:
\begin{verbatim}
{"original_caption": "<1-2 sentences describing the original image>",
 "edited_caption": "<1-2 sentences describing the image after the edit>"}
\end{verbatim}

\vspace{0.5em}
\hrule
\vspace{0.5em}

\noindent\textbf{STYLE \& CONTENT RULES}
\begin{itemize}[leftmargin=*, nosep]
    \item Base captions solely on visible content; no unfounded guesses.
    \item Mention dominant objects, attributes, actions, and setting.
    \item In \texttt{edited\_caption}, describe the resulting scene only---do NOT mention the edit process or instruction.
    \begin{itemize}[leftmargin=*, nosep]
        \item (Bad: ``The image is edited to\ldots.'' Good: ``A red balloon now floats above the dog\ldots.'')
        \item (Bad: ``An apple, no pear\ldots.'' Good: ``An apple\ldots.'')
    \end{itemize}
    \item Each caption $\leq$ 40 English words.
    \item Do not quote the user's instruction verbatim; convey its visual effect instead.
    \item If the instruction is impossible or unsafe, produce the best policy-compliant visual outcome.
\end{itemize}

\vspace{0.5em}
\hrule
\vspace{0.5em}

\noindent\textbf{FINAL CHECKLIST}
\begin{itemize}[leftmargin=*, nosep]
    \item[$\checkmark$] Return exactly one JSON object---no extra text, blank lines, or comments.
    \item[$\checkmark$] Keys must be \texttt{original\_caption} and \texttt{edited\_caption}, in that order.
    \item[$\checkmark$] Follow length, tense, tone, and truthfulness requirements.
\end{itemize}

\vspace{0.5em}
\hrule
\vspace{0.5em}

\noindent\textbf{User Input:} \\
\texttt{edit instruction:} $\langle$edit\_prompt$\rangle$ \\
\texttt{original image:} $\langle$image$\rangle$

\end{promptbox_blue}
\vspace{-1.1em}
\caption{\textbf{Prompt $P_{\text{cap}}$ for generating output caption}.}
\vspace{-0.5em}
\label{supp_prompt:caption_generation}
\end{figure*}

\subsubsection{Filtering Visually Similar Candidates} 
\label{supp_sec: semantic_similarity_filter}

To eliminate redundancy, we filter out visually similar candidates. Goal-directed image editing often yields multiple redundant results during large-scale sampling. Notably, this redundancy is already apparent in the early preview images, as illustrated in \cref{supp_fig: semantic_similarity_filter}. To address this, we extract visual embeddings from each preview using DINOv2 and compute pairwise similarity. If the similarity between two candidates exceeds a threshold $\tau_{\text{sim}}$, the one with the lower unified score is discarded. This step ensures that only visually distinct and high-potential candidates are retained.


{
\begin{figure}[t]
\begin{center}
\includegraphics[width=0.99\linewidth]{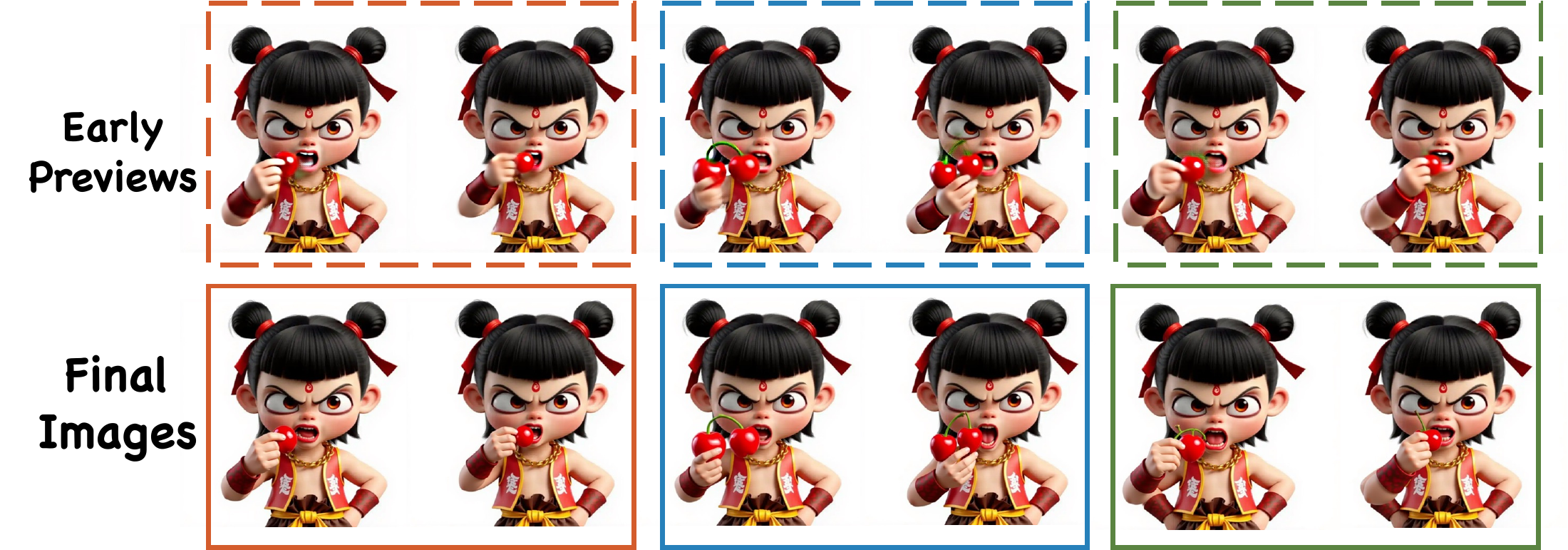}
\end{center}
\vspace{-1.7em}
\caption{\textbf{Redundancy appears in early preview images.} The figure displays clusters of visually similar images, grouped by colored boxes. We observe that candidates which are highly similar in the final output (bottom row) are also highly similar in their early previews (top row). This confirms that visual redundancy can be detected and filtered at an early stage.}
\vspace{-0.6em}
\label{supp_fig: semantic_similarity_filter}
\end{figure}
}

\subsection{Depth-first Opportunistic Stopping}
\label{supp_sec: adapt_stop}

As described in Sec.~3.3 of the main paper, our depth-first opportunistic stopping strategy avoids unnecessary computation on redundant yet correct results. It is summarized in Alg.~\ref{alg: adapt_stop}, which consists of two key components: (1) a late-stage filter to retain the most promising candidates; and (2) an instance-specific verifier to guide the stopping decision.

The algorithm operates as follows. 
\ding{182} We first initialize an empty set $\mathcal{U}$ to store image-score pairs $(I, S)$, a retain threshold $S_{\mathrm{rt}}$, and a high-score count $N_{\mathrm{cnt}}$ (Lines 1-3). 
\ding{183} For each candidate $(x_{t_e}, c_r)$ from the sorted sets $(\mathcal{X}_{t_e}, \mathcal{C})$ (Line 5), we perform depth-first sequential generation. We first sample from early timestep $t_e$ to late timestep $t_l$ (Line 6) and apply the one-step preview mechanism to obtain $x_{0|t_l}$ (Line 7), which is decoded into preview image $I_{0|t_l}$ (Line 8). The unified score $S_{0|t_l}$ is computed using the same verifiers as in early pruning (Line 9).
\ding{184} We apply an adaptive late-stage filter. If $S_{0|t_l} \geq S_{\mathrm{rt}} - \delta$, we update the retain threshold $S_{\mathrm{rt}}$ (Line 12) and continue sampling from $t_l$ to $0$ to generate the final image $I$ (Lines 13-14). Otherwise, we skip this candidate and proceed to the next one.
\ding{185} For each retained candidate, we compute the unified score $S$ (Line 15) and apply the instance-specific verifier to obtain $S_{\text{spec}}$ (Line 17). We update the evaluated score as $S \leftarrow S + S_{\text{spec}}$ (Line 18). If $S_{\text{spec}} \geq S_{\text{high}}$ (indicating all \textit{yes-no} questions are answered ``\textit{yes}''), we increment the high-score count $N_{\mathrm{cnt}}$ (Lines 19-21). The image-score pair $(I, S)$ is added to $\mathcal{U}$ (Line 22).
\ding{186} The search terminates when $N_{\mathrm{cnt}} = N_{\mathrm{high}}$ (Lines 24-27), meaning sufficient intent-aligned results have been found. Finally, we return the set $\mathcal{U}$ (Line 29), and the candidate with the highest score is selected as the output.

\begin{algorithm*}[t]
\caption{\texttt{Adaptive\_Stop} algorithm - Lines 5-6 of Alg.~\ref{alg: ADE_CoT}.} 
\label{alg: adapt_stop}
\begin{algorithmic}[1]
\REQUIRE source image $I_{\text{src}}$, text prompt $c$, intermediate latent set $\mathcal{X}_{t_e}$, prompt set $\mathcal{C}$, early step $t_e$ and retain step $t_l$
\REQUIRE high-score number $N_{\mathrm{high}}$, high-score threshold $S_{\mathrm{high}}$, tolerance factor $\delta$
\STATE $\mathcal{U} \leftarrow \{ \}$ \quad $\triangleright$ Initialize empty set for $(I, S)$ pairs
\STATE $S_{\mathrm{rt}} \leftarrow 0$ \quad $\triangleright$ Initialize retain threshold
\STATE $N_{\mathrm{cnt}} \leftarrow 0$ \quad $\triangleright$ Initialize high-score count
\STATE \textcolor{method_green}{$\triangleright$ \textbf{Depth-first generation}}
\FOR{$(x_{t_e}, c_r)$ in $(\mathcal{X}_{t_e}, \mathcal{C})$ }
    \STATE $x_{t_l} \leftarrow \texttt{Sampler}(I_{\text{src}}, x_{t_e}, c_r, t_e, t_l) $ \quad \textcolor{method_green}{$\triangleright$ Sample from $t_e$ to $t_l$}
    \STATE $x_{0 | t_l } \leftarrow \texttt{One\_Step\_Preview}(x_{t_l}, t_l)$ \quad \textcolor{method_green}{$\triangleright$ Preview image from intermediate latent (\cf Sec~3.2 Eq.~4)}
    \STATE $I_{0|{t_l}} \leftarrow \texttt{VAE\_Decoder} (x_{0 | t_l }) $
    \STATE $S_{0| t_l} \leftarrow \texttt{Vrf}(I_{\text{src}}, I_{0| t_l}, c ) $ \quad $\triangleright$ Compute unified score (\cf Sec~3.2, Eq.~8) 
    \STATE \textcolor{method_green}{$\triangleright$ \textbf{Retain top results by evaluated score} }
    \IF{$S_{0|t_l} >= S_{\mathrm{rt}} - \delta$}
        \STATE $S_{\mathrm{rt}} \leftarrow \texttt{max}(S_{\mathrm{rt}}, S_{0 | t_l})$ \quad \textcolor{method_green}{$\triangleright$ Update retain threshold}
        \STATE $x_0 \leftarrow \texttt{Sampler}(I_{\text{src}}, x_{t_l}, c_r, t_l, 0 ) $ \quad \textcolor{method_green}{$\triangleright$ Sample from $t_l$ to $0$}
        \STATE $I \leftarrow \texttt{VAE\_Decoder}(x_0)$ 
        \STATE $S \leftarrow \texttt{Vrf}(I_{\text{src}}, I, c) $ \quad $\triangleright$ Compute unified score 
        \STATE \textcolor{method_green}{$\triangleright$ \textbf{Instance-specific verification}}
        \STATE $S_{\text{spec}} \leftarrow \texttt{Specific\_Verifier}(I_{\text{src}}, I, c) $ \quad \textcolor{method_green}{$\triangleright$ Compute instance-specific score}
        \STATE $S \leftarrow S + S_{\text{spec}}$ \quad $\triangleright$ Update evaluated score
        \IF{$S_{\mathrm{spec}} \geq S_{\text{high}}$} 
            \STATE $N_{\mathrm{cnt}} \leftarrow N_{\mathrm{cnt}} + 1$ \quad \textcolor{method_green}{$\triangleright$ Update high-score count} 
        \ENDIF      
        \STATE $\mathcal{U} \leftarrow \mathcal{U} \cup \{ (I, S) \}$ 
    \ENDIF
    \STATE \textcolor{method_green}{$\triangleright$ \textbf{Stop when intent-aligned results suffice}}
    \IF{$N_{\mathrm{cnt}} == N_{\mathrm{high}} $}
        \STATE break 
    \ENDIF
\ENDFOR
\RETURN $\mathcal{U}$ 
\end{algorithmic}
\end{algorithm*}

\subsubsection{Details of Retaining Top Results}
\label{supp_sec: retain_top_results} 

Lines 10-23 of Alg.~\ref{alg: adapt_stop} implement the late-stage filter to retain the most promising candidates. Unlike the early pruning stage that uses a fixed rejection threshold $S_{\mathrm{rj}}$, we adopt an adaptive filtering strategy at the late timestep $t_l$. This is motivated by the observation that preview scores at later denoising stages exhibit stronger correlation with final image quality. For each candidate, we generate a preview image $I_{0|t_l}$ and compute its unified score $S_{0|t_l}$ (Lines 6-9). We maintain a retain threshold $S_{\mathrm{rt}}$ that dynamically updates to the maximum score observed so far (Line 12). A candidate is retained if its score $S_{0|t_l}$ is within a tolerance $\delta$ of the current threshold: $S_{0|t_l} \geq S_{\mathrm{rt}} - \delta$ (Line 11). This ensures that only candidates with scores comparable to the current best are fully denoised. This adaptive filter dynamically adjusts to the quality distribution of candidates. It avoids wasting computation on samples that are unlikely to be optimal.

\subsubsection{Details of Instance-Specific Verifier}
\label{supp_sec: instance_specific_verify}

\begin{figure*}[t]
\centering
\begin{promptbox_green}{Prompt $P_q$ for generating instance-specific questions}

\textbf{System Role:} You are an expert AI assistant specializing in Image Editing Quality Assurance (QA). Your primary role is to act as a meticulous verifier of digital image edits. Your task is to analyze an Original Image and a corresponding Edit Instruction, and then generate a set of exactly 5 specific, verifiable questions. The fundamental rule is this: if a human reviewer answers ``yes'' to all 5 of your questions, it must unequivocally confirm that the edit was successfully and perfectly executed according to the instruction.

\vspace{0.5em}
\hrule
\vspace{0.5em}

\noindent\textbf{INPUT}
\begin{itemize}[leftmargin=*, nosep]
    \item \texttt{Original Image}: [The initial image before editing]
    \item \texttt{Edit Instruction}: [A text description of the desired change]
\end{itemize}

\vspace{0.5em}
\hrule
\vspace{0.5em}

\noindent\textbf{OUTPUT (must follow exactly)}
\begin{itemize}[leftmargin=*, nosep]
    \item Every question MUST be phrased so that a ``yes'' answer confirms a positive outcome.
    \item Return a single JSON object with the following structure:
\end{itemize}
\begin{verbatim}
{"questions": [
    "Question 1",
    "Question 2",
    "Question 3",
    "Question 4",
    "Question 5" ]}
\end{verbatim}

\vspace{0.5em}
\hrule
\vspace{0.5em}

\noindent\textbf{CORE PRINCIPLES FOR QUESTION GENERATION}
\begin{itemize}[leftmargin=*, nosep]
    \item \textbf{Instruction-Centric:} Every question must directly derive from the Edit Instruction. Deconstruct the instruction into its core components (e.g., object, action, style, location).
    \item \textbf{Binary \& Objective:} Frame each question to be answerable with a simple and objective ``Yes'' or ``No''. Avoid subjective questions like ``Does it look better?'' and instead focus on verifiable facts, such as ``Has the color of the car been changed from blue to red?''.
    \item \textbf{Comprehensive Coverage:} Your 5 questions must collectively cover all aspects of the instruction. If the instruction is ``Make the man taller and add a hat,'' you must have questions that verify both his height and the presence of the hat.
    \item \textbf{Negative Verification (No Collateral Damage):} At least one question must check for unintended side effects. This includes verifying that parts of the image not mentioned in the instruction remain unchanged, and that no new artifacts, blurs, or distortions have been introduced.
    \item \textbf{Holistic Quality Check:} At least one question must assess the overall integration and realism of the edit. It should check if the edit blends seamlessly with the rest of the image, maintaining consistent lighting, shadows, and texture.
\end{itemize}

\vspace{0.5em}
\hrule
\vspace{0.5em}

\noindent\textbf{User Input:} \\
\texttt{Edit Instruction:} $\langle$edit\_instruction$\rangle$ \\
\texttt{Original Image:} $\langle$image$\rangle$

\end{promptbox_green}
\vspace{-1.1em}
\caption{\textbf{Prompt $P_q$ for generating instance-specific questions}.}
\vspace{-0.5em}
\label{supp_prompt:instance_specific_question}
\end{figure*}

\begin{figure*}[t]
\centering
\begin{promptbox_green}{Prompt $P_a$ for answering instance-specific questions}

\textbf{System Role:} You are an ``Image-Edit Compliance Judge.'' For every dialogue turn you will receive:
\begin{itemize}[leftmargin=*, nosep]
    \item \texttt{EDIT\_INSTRUCTION} -- A text description of the desired change.
    \item \texttt{QUESTION\_LIST} -- exactly five yes/no questions, each asking whether a certain visual condition is true after the edit.
    \item \texttt{ORIGINAL\_IMAGE} -- The initial image before editing.
    \item \texttt{EDITED\_IMAGE} -- an edited version of the initial image.
\end{itemize}

\vspace{0.5em}
\hrule
\vspace{0.5em}

\noindent\textbf{YOUR TASK}
\begin{itemize}[leftmargin=*, nosep]
    \item Imagine the edit is carried out exactly as written and reason about the resulting image.
    \item For each of the five questions decide ``yes'' (the condition is satisfied) or ``no'' (the condition is not satisfied or cannot be inferred). When unsure, answer ``no.''
    \item Return nothing except a JSON object with five keys: \texttt{"Q1"}, \texttt{"Q2"}, \texttt{"Q3"}, \texttt{"Q4"}, \texttt{"Q5"}.
    \item The value of every key must be the lowercase string ``yes'' or ``no''.
    \item Do not output any explanations, comments, or additional keys.
\end{itemize}

\vspace{0.5em}
\hrule
\vspace{0.5em}

\noindent\textbf{OUTPUT (must follow exactly)} \\
Return a single JSON object with the following structure:
\begin{verbatim}
{"Q1": "yes|no",
 "Q2": "yes|no",
 "Q3": "yes|no",
 "Q4": "yes|no",
 "Q5": "yes|no"}
\end{verbatim}

\vspace{0.5em}
\hrule
\vspace{0.5em}

\noindent\textbf{User Input:} \\
\texttt{EDIT\_INSTRUCTION:} $\langle$edit\_instruction$\rangle$ \\
\texttt{QUESTION\_LIST:} $\langle$instance-specific\_question$\rangle$ \\
\texttt{ORIGINAL\_IMAGE:} $\langle$original\_image$\rangle$ \\
\texttt{EDITED\_IMAGE:} $\langle$edited\_image$\rangle$

\end{promptbox_green}
\vspace{-1.1em}
\caption{\textbf{Prompt $P_a$ for answering instance-specific questions}.}
\vspace{-0.5em}
\label{supp_prompt:instance_specific_answer}
\end{figure*}

The instance-specific verifier provides a fine-grained assessment to distinguish high-quality results from subtly flawed ones. It employs a two-stage process. First, using prompt $P_q$ (shown in~\cref{supp_prompt:instance_specific_question}), the MLLM generates a set of five specific \textit{yes/no} questions tailored to the edit instruction. These questions cover aspects such as instruction adherence and aesthetics. Second, using prompt $P_a$ (shown in~\cref{supp_prompt:instance_specific_answer}), the MLLM answers these questions for the fully generated image. As implemented in Alg.~\ref{alg: adapt_stop}, the instance-specific score, $S_{\text{spec}}$, is calculated by the verifier (Line 17). This score is then added to the candidate's unified score to reward correctness (Line 18). We also track the number of intent-aligned candidates using a counter $N_{\text{cnt}}$, which is incremented if $S_{\text{spec}} \geq S_{\text{high}}$ (Lines 19--21). This counter is used to trigger the final opportunistic stopping condition.

\section{Additional Experimental Results}
\label{supp_sec: extra_results}

\subsection{Experimental Details}
\label{supp_sec: experimental_details}

Our method, ADE-CoT, is built upon three open-sourced, state-of-the-art image editing models: Step1X-Edit~\cite{Step1X_Edit_arxiv2025}, FLUX.1 Kontext~\cite{FLUX_Kontext_arxiv2025}, and BAGEL~\cite{Bagel_2025_arxiv}. We adhere to their default configurations for the total number of denoising steps, which are $T=28, 28, 50$, respectively. The early pruning timestep $t_e$ and the late retaining timestep $t_l$ are set based on the total steps for each model. The specific hyperparameter settings used in our experiments are detailed in~\cref{supp_tab: hyperparameters}. We use Qwen-VL-MAX~\cite{qwen25vl} as the MLLM for all queries and VIE-Score~\cite{VIE_Score} as our general verifier $S_{\text{gen}}$. To ensure the robustness of our findings, we conduct each scaling experiment three times with different random seeds for sampling noise. The results reported in the main paper are the average of these three runs. When multiple candidates achieve the same maximum score, we compute their pairwise visual similarity using DINOv2~\cite{DINO_v2} embeddings. 
We select the candidate with the highest average similarity to others as the centroid, representing the visual consensus among top-scoring outputs.

{
\setlength{\tabcolsep}{3.6pt}
\renewcommand{\arraystretch}{1.1} 
\begin{table}[h]
\centering
\caption{\textbf{Default hyperparameters used in our experiments.} }
\vspace{-0.5em}
\label{supp_tab: hyperparameters}
\resizebox{\columnwidth}{!}{%
\begin{tabular}{l|c|l}
\toprule
\textbf{Hyper.} & \textbf{Value} & \textbf{Description} \\
\midrule
$T$ & 28, 28, 50 & Total denoising steps for each model. \\
$t_e$ & 8, 8, 16 & Timestep for early preview and pruning. \\
$t_l$ & 16, 16, 36 & Timestep for late retaining. \\
$N_{\text{min}}$ & 1 & Minimum sampling budget. \\
$\gamma$ & 0.15 & Sensitivity for difficulty-aware allocation. \\
$S_{\text{max}}$ & 10 & Maximum score for normalization. \\
$\lambda_{\text{reg}}$ & 1 & Weight for the region correctness score. \\
$\lambda_{\text{cap}}$ & 3 & Weight for the caption consistency score. \\
$S_{\text{rj}}$ & 5 & Rejection threshold for early pruning. \\
$\tau_{\text{sim}}$ & 0.98 & Similarity threshold for filtering candidates. \\
$N_{\text{high}}$ & 4 & Number of intent-aligned results to stop. \\
\bottomrule
\end{tabular}
}
\end{table}
}

\subsection{Details of Evaluation setting}
\label{supp_sec: details_of_evaluation}

\textbf{Proposed efficiency metrics.}
We introduce two metrics to measure efficiency from different perspectives.
\ding{182} \textbf{Reasoning Efficiency ($\eta$):} This metric is designed to measure the overall trade-off between performance and computational cost. An effective pruning strategy must not only reduce the Number of Function Evaluations (NFE) but also maintain high image quality. The design of $\eta$ rewards methods that achieve high final scores with low NFE. The binary factor $\sigma_i$ ensures that only methods achieving non-degraded performance are considered, which prevents strategies from gaining high efficiency scores by producing poor results.
\ding{183} \textbf{Outcome Efficiency ($\xi$):} This metric is designed to quantify generation redundancy. Large-scale sampling in image editing often yields multiple correct outputs. An ideal method should find a satisfactory result quickly. $\xi$ measures this by comparing the total NFE spent against the minimum NFE required to generate the first acceptable image. A higher $\xi$ indicates that the method wastes less computation on producing redundant correct images.

\textbf{Two comparison settings}. 
We analyze the performance of all methods from two different settings to provide a comprehensive comparison.
\ding{182} \textbf{Results under fixed sampling budget.} In this setting, all methods start with the same initial sampling budget (e.g., $N=32$). This comparison aims to identify which method achieves the best performance and efficiency when allocated a fixed amount of computational resources. It directly shows the effectiveness of different pruning and search strategies.
\ding{183} \textbf{Results under comparable performance.} In this setting, we compare the computational cost of methods that achieve a similar quality level, specifically a non-degraded performance relative to the Best-of-N (BoN) baseline. The goal is to measure the actual speedup a method provides while maintaining a target performance. This highlights the practical value of a strategy in terms of saving time and resources.





\subsection{More Ablation Studies}
\label{supp_sec: ablation}

We present a comprehensive ablation study of our key components in~\cref{tab: supp_efficiency_improve}. Most of these results have been analyzed in the main paper. Here, we provide additional analyses to further validate the effectiveness of our design choices.

{
\begin{table*}[t]
    \caption{\textbf{Effect of the three proposed strategies on efficiency and performance}. We evaluate our method on GEdit-Bench~\cite{Step1X_Edit_arxiv2025}.}
    \label{tab: supp_efficiency_improve}
    \vspace{-0.6em}
    \centering
    \resizebox{1.0\linewidth}{!}
    {
    \scriptsize
\begin{tabular}{l | c | llll | llll | llll }
\toprule
\multirow{2}{*}{\textbf{Model}} & \multirow{2}{*}{$N$} & \multicolumn{4}{c|}{\textbf{FLUX.1 Kontext~\cite{FLUX_Kontext_arxiv2025}}} & \multicolumn{4}{c|}{\textbf{BAGEL~\cite{Bagel_2025_arxiv}}} & \multicolumn{4}{c}{\textbf{Step1X-Edit}~\cite{Step1X_Edit_arxiv2025}}  \\ 
\cmidrule(lr){3-6} \cmidrule(lr){7-10} \cmidrule(lr){11-14}
 & & G\_O $\uparrow$ & NFE $\downarrow$ & $\eta$ $\uparrow$ & $\xi$ $\uparrow$ & G\_O $\uparrow$ & NFE $\downarrow$ & $\eta$ $\uparrow$ & $\xi$ $\uparrow$ & G\_O $\uparrow$ & NFE $\downarrow$ & $\eta$ $\uparrow$ & $\xi$ $\uparrow$ \\ 
\midrule 
Baseline (BoN) & 32 & 6.641 & 896 & 0.66 & 0.11 & 6.908 & 1600 & 0.69 & 0.14 & 7.157 & 896 & 0.72 & 0.13   \\
\midrule 
\HighTableOrange{\texttt{a)} $+$ adaptive sampling} & \HighTableOrange{32} & \HighTableOrange{6.641} & \HighTableOrange{797} & \HighTableOrange{0.74} & \HighTableOrange{0.26} & \HighTableOrange{6.909} & \HighTableOrange{1391} & \HighTableOrange{0.76} & \HighTableOrange{0.23} & \HighTableOrange{7.157} & \HighTableOrange{778} & \HighTableOrange{0.81} & \HighTableOrange{0.27}  \\ 
\midrule 
\HighTableBlue{\texttt{b)} $+$ early filter by general verifier} & \HighTableBlue{32} & \HighTableBlue{6.642} & \HighTableBlue{719} & \HighTableBlue{0.81} & \HighTableBlue{0.40} & \HighTableBlue{6.912} & \HighTableBlue{1351} & \HighTableBlue{0.79} & \HighTableBlue{0.39} & \HighTableBlue{7.157} & \HighTableBlue{719} & \HighTableBlue{0.89} & \HighTableBlue{0.42}    \\ 
\HighTableBlue{\texttt{c)} $+$ early filter (+ $S_\text{reg}$)} & \HighTableBlue{32} & \HighTableBlue{6.645} & \HighTableBlue{687} & \HighTableBlue{0.84} & \HighTableBlue{0.42} & \HighTableBlue{6.915} & \HighTableBlue{1321} & \HighTableBlue{0.84} & \HighTableBlue{0.43} & \HighTableBlue{7.158} & \HighTableBlue{678} & \HighTableBlue{0.95} & \HighTableBlue{0.45} \\ 
\HighTableBlue{\texttt{d)} $+$ early filter (+ $S_\text{cap}$)} & \HighTableBlue{32} & \HighTableBlue{6.647} & \HighTableBlue{673} & \HighTableBlue{0.87} & \HighTableBlue{0.44} & \HighTableBlue{6.916} & \HighTableBlue{1290} & \HighTableBlue{0.88} & \HighTableBlue{0.45} & \HighTableBlue{7.161} & \HighTableBlue{638} & \HighTableBlue{1.02} & \HighTableBlue{0.47} \\ 
\HighTableBlue{\texttt{e)} $+$ removing visually similar samples} & \HighTableBlue{32} & \HighTableBlue{6.651} & \HighTableBlue{508} & \HighTableBlue{1.26} & \HighTableBlue{0.58} & \HighTableBlue{6.915} & \HighTableBlue{1087} & \HighTableBlue{1.02} & \HighTableBlue{0.50} & \HighTableBlue{7.162} & \HighTableBlue{522} & \HighTableBlue{1.22} & \HighTableBlue{0.54}  \\ 
\midrule 
\HighTableGreen{\texttt{f)} $+$ retaining top results at late stage} & \HighTableGreen{32} & \HighTableGreen{6.652} & \HighTableGreen{464} & \HighTableGreen{1.34} & \HighTableGreen{0.61} & \HighTableGreen{6.935} & \HighTableGreen{972} & \HighTableGreen{1.08} & \HighTableGreen{0.54} & \HighTableGreen{7.163}  & \HighTableGreen{462} & \HighTableGreen{1.34} & \HighTableGreen{0.58}  \\ 

\HighTableGreen{\texttt{g)} $+$ instance-specific verifier} & \HighTableGreen{32} & \HighTableGreen{\textbf{6.702}} & \HighTableGreen{464} & \HighTableGreen{1.37} & \HighTableGreen{0.63} & \HighTableGreen{\textbf{6.984}} & \HighTableGreen{972} & \HighTableGreen{1.12} & \HighTableGreen{0.57} & \HighTableGreen{\textbf{7.206}}  & \HighTableGreen{462} & \HighTableGreen{1.36} & \HighTableGreen{0.60} \\  

\HighTableGreen{\texttt{h)} $+$ opportunistic stopping (\textbf{full model}) } & \HighTableGreen{32} & \HighTableGreen{6.695} & \HighTableGreen{\textbf{418}} & \HighTableGreen{\textbf{1.47}} & \HighTableGreen{\textbf{0.66}} & \HighTableGreen{6.972} & \HighTableGreen{\textbf{882}} & \HighTableGreen{\textbf{1.27}} & \HighTableGreen{\textbf{0.62}}  & \HighTableGreen{7.196} & \HighTableGreen{\textbf{434}} & \HighTableGreen{\textbf{1.45}} & \HighTableGreen{\textbf{0.62}} \\
\bottomrule
\end{tabular}
}
\vspace{-0.4 em}
\end{table*}
}

\textbf{Are both edited-region correctness and instruction-caption consistency effective?}
To isolate the impact of our edit-specific verifiers, we start from a baseline that uses only early filtering with a general verifier ($S_\text{gen}$) (row b). As shown in \cref{tab: supp_efficiency_improve}, adding the edited-region correctness score ($S_\text{reg}$) (row c) consistently reduces NFE across all models while maintaining performance. This indicates that $S_\text{reg}$ is effective at pruning candidates with incorrect edit localization. We then further incorporate the instruction-caption consistency score ($S_\text{cap}$) (row d). This step yields additional efficiency gains, reducing the NFE for Step1X-Edit from 678 to 638 and for FLUX.1 Kontext from 687 to 673. These results confirm that both $S_\text{reg}$ and $S_\text{cap}$ are effective and complementary. They target distinct failure modes—localization and semantic alignment, respectively—and their combined use significantly enhances the precision of our early pruning strategy.

\textbf{Why fixed thresholds for early pruning but dynamic thresholds for late retaining?}  
Our choice of pruning thresholds is based on the correlation between intermediate and final scores at different denoising stages, as shown in \cref{supp_fig: denoised_to_final}. In the early stage, the correlation between preview scores and final scores is moderate. A low preview score does not guarantee a low final score. An aggressive pruning strategy is therefore risky, as it could discard high-potential candidates. We use a fixed, conservative threshold to safely remove only clear failures. Conversely, the correlation becomes much stronger in the late stage. \cref{supp_fig: denoised_to_final}(b) shows that late-stage scores are highly predictive of the final quality. This strong correlation allows for a more aggressive strategy. We apply a dynamic threshold to filter out candidates that are unlikely to surpass the current best, which improves efficiency without performance loss.

{
\begin{figure}[t]
\centering
{
    \hfill
    \subfloat[Early denoising stage.]{\includegraphics[height=0.165\textwidth]{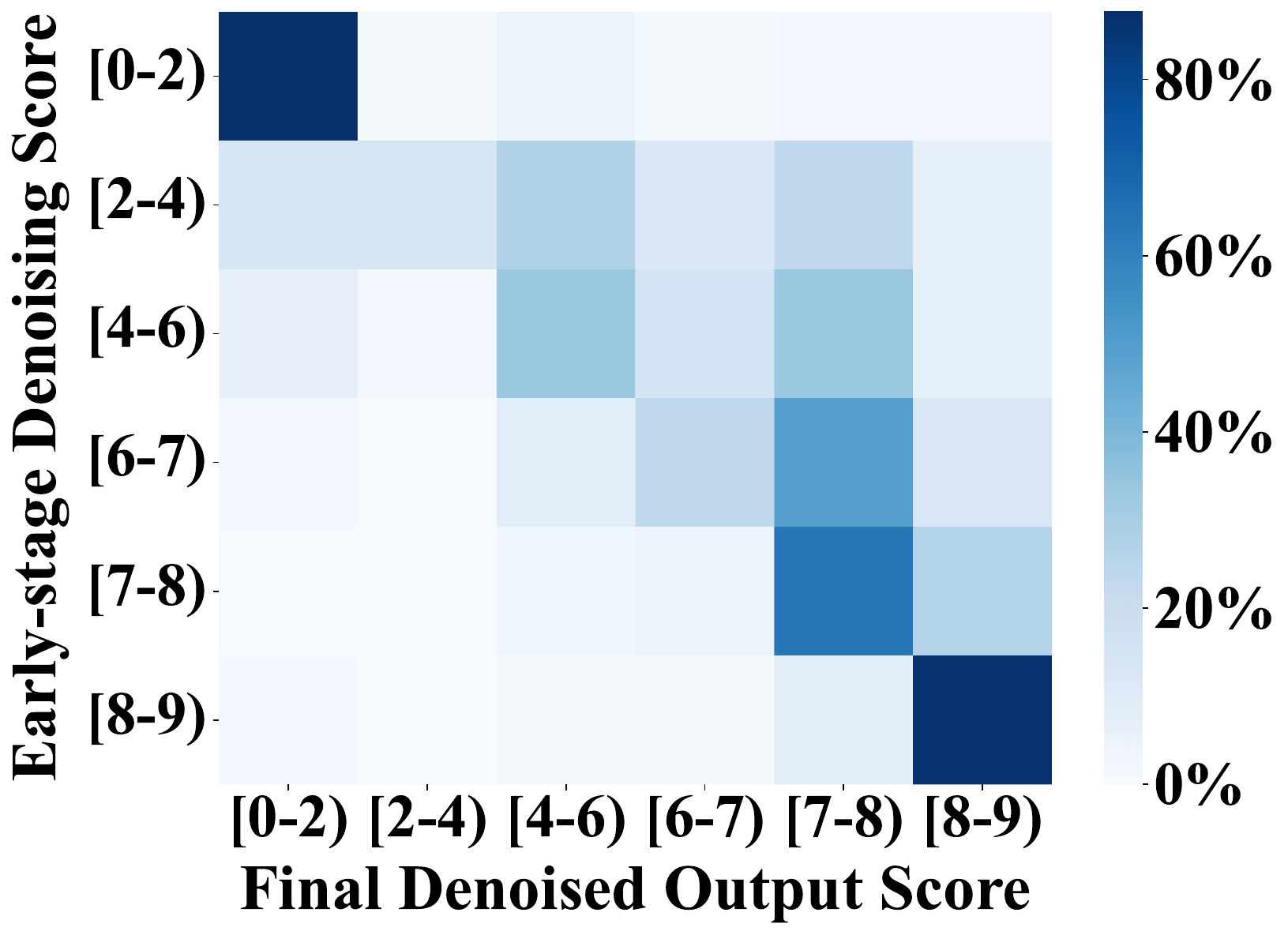}}
    \hfill
    \subfloat[Late denoising stage.]{\includegraphics[height=0.165\textwidth]{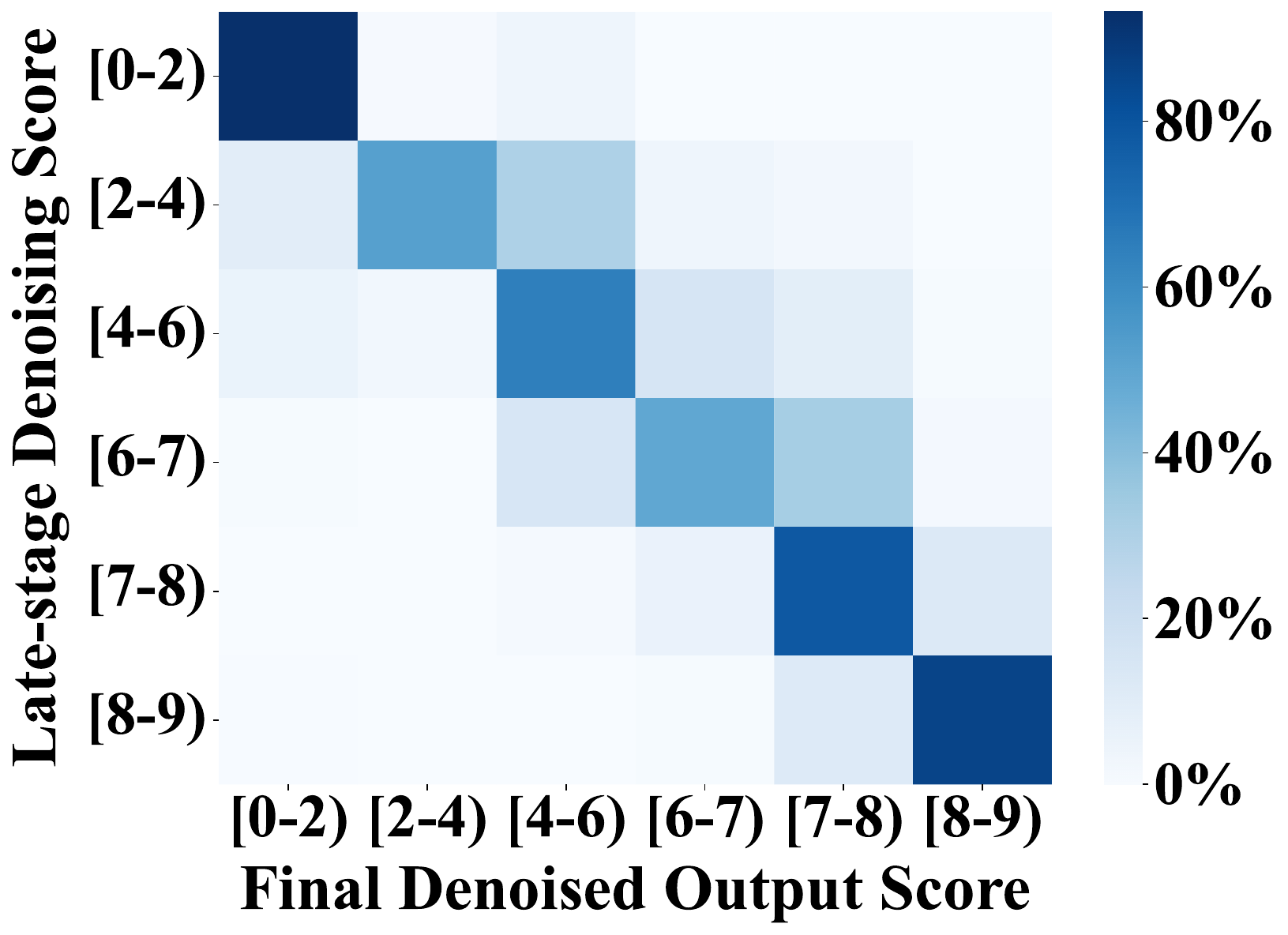}}
}
\vspace{-7pt}
\caption{\textbf{Correlation between preview scores and final performance across denoising stages.} We show correlation matrixs between intermediate scores (y-axis) and final scores (x-axis) at \textbf{(a)} the early and \textbf{(b)} late denoising stages.}
\label{supp_fig: denoised_to_final}
\vspace{-5pt}
\end{figure}
}

\subsection{More Hyperparameter Analysis}
\label{supp_sec: hyper_analysis}

\noindent \textbf{What are the optimal hyperparameters $S_{rj}$ and $\tau_{sim}$?}
We analyze the impact of the rejection threshold $S_{rj}$ and the similarity threshold $\tau_{sim}$ in~\cref{supp_fig: hyper_result}. Increasing $S_{rj}$ improves reasoning efficiency by pruning low-potential samples, while performance (G\_O) remains stable up to a threshold of 5. However, a higher $S_{rj}$ may remove potentially correct candidates, causing both metrics to decline. For $\tau_{sim}$, decreasing it removes redundant images, which improves both performance and reasoning efficiency, peaking at $\tau_{sim} = 0.98$. This may be because discarded similar images also tend to have low potential scores. A further decrease causes a sharp drop in both metrics. Thus, we set $S_{rj} = 5$ and $\tau_{sim} = 0.98$ as our default values.

{
\begin{figure}[t]
\centering
{
    \hfill
    \subfloat[$S_\text{rj}$]{\includegraphics[width=0.235\textwidth]{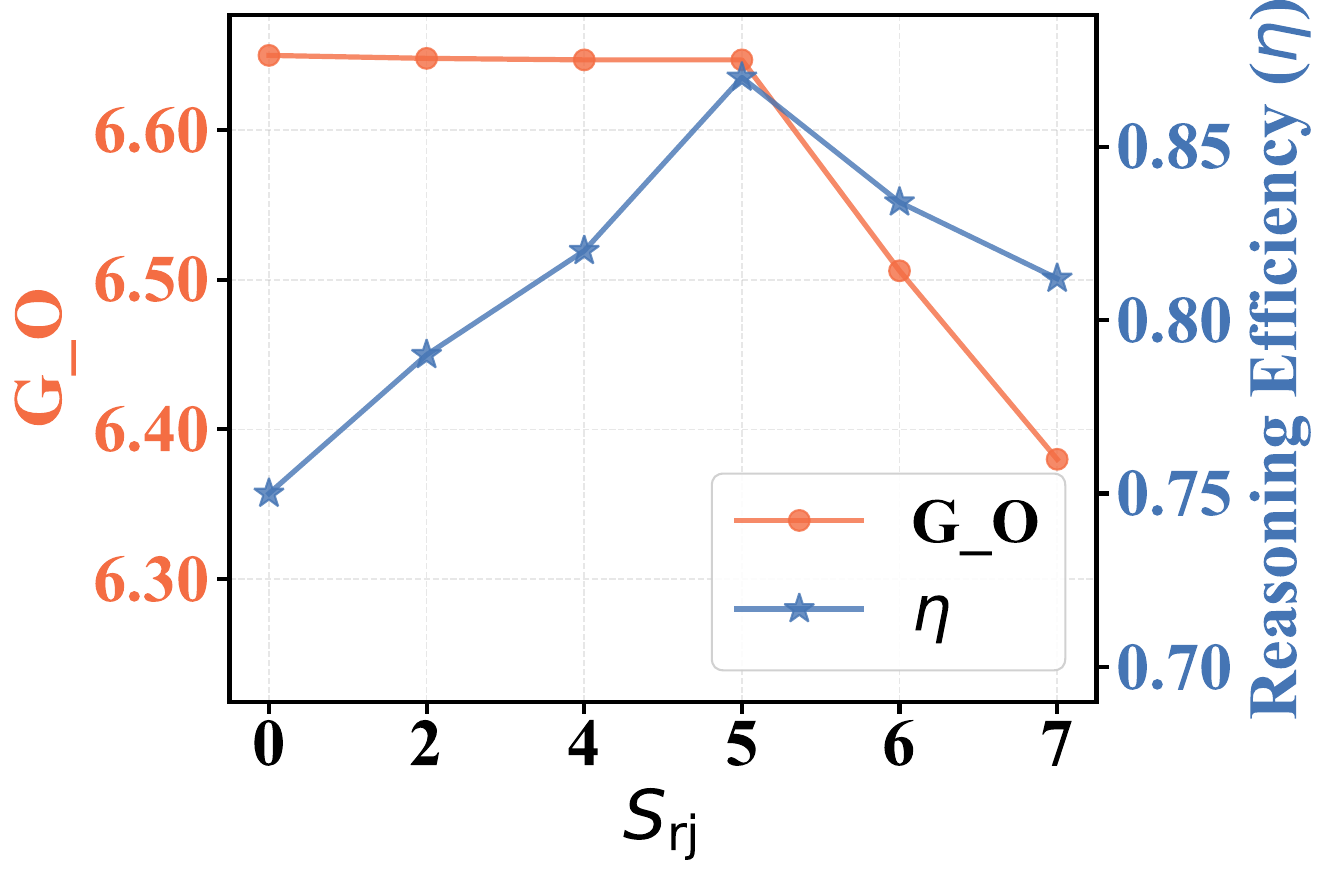}}
    \hfill
    \subfloat[$\tau_\text{sim}$]{\includegraphics[width=0.235\textwidth]{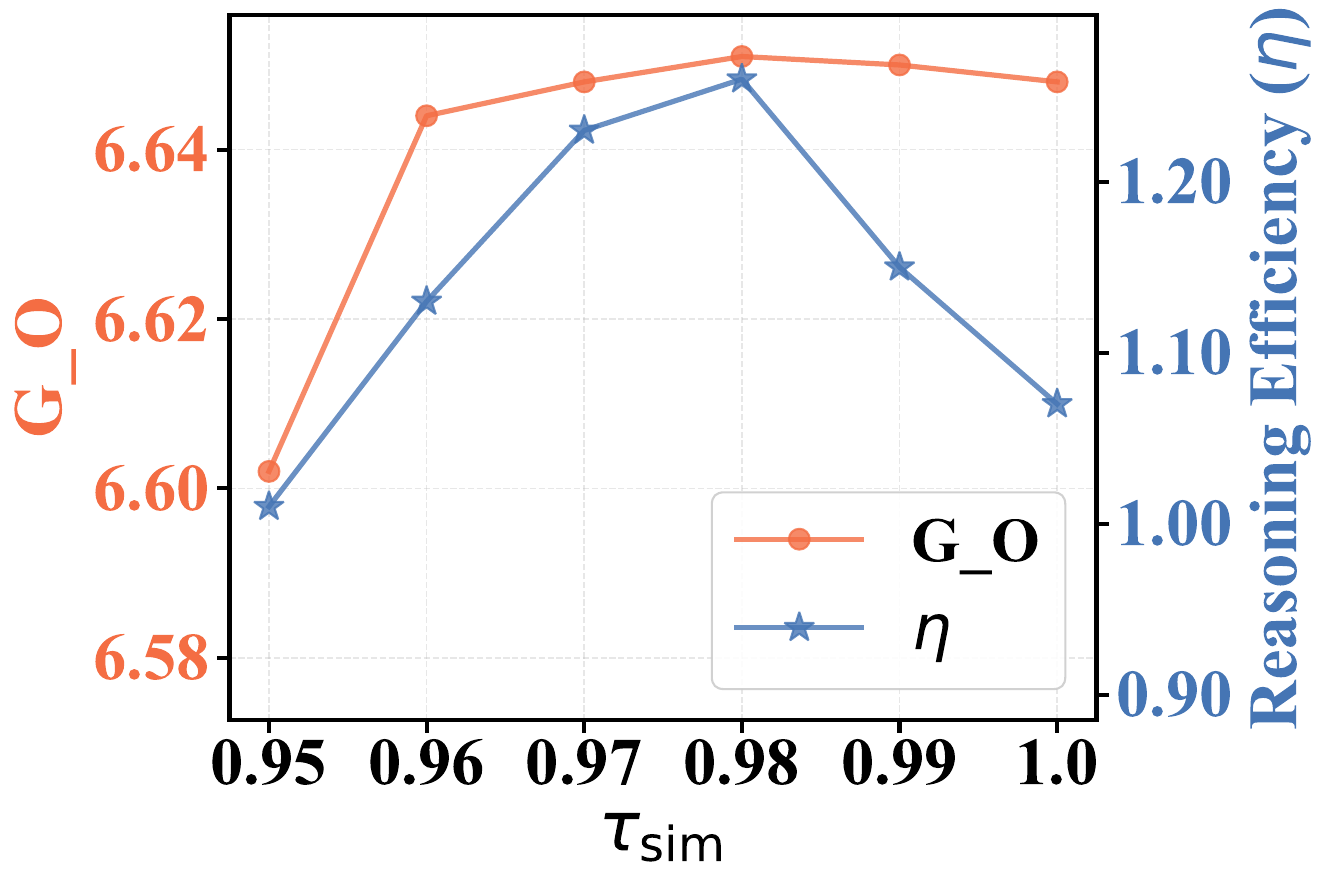}}
}
\vspace{-7pt}
\caption{\textbf{More hyperparameter analysis}.
}
\label{supp_fig: hyper_result}
\vspace{-11pt}
\end{figure}
}

\subsection{More Analysis of Cost Computation}
\label{supp_sec: cost_computation}

Beyond the Number of Function Evaluations (NFE), we analyze the computational cost of MLLM queries in our framework. \cref{supp_tab: mllm_cost} reports the average number of MLLM calls per editing case on GEdit-Bench under a sampling budget of $N=32$ across three models. We observe that the general MLLM verifier, which is also required by all prior Image-CoT methods~\cite{ICEdit_arxiv2025, TTS_Baseline, Image_CoT}, accounts for the majority of queries in our framework. This component evaluates multiple candidates during early pruning and late retaining stages. In contrast, verifiers introduced by our method, including edited-region localization (via prompt $P_{\text{reg}}$), instruction-caption consistency (via prompt $P_{\text{cap}}$), and the instance-specific verifier, contribute minimal overhead. Specifically, region and caption generation require only 1.0 query per case. The instance-specific verifier generates questions once per case and answers them for a small number of top candidates, resulting in an average of 14.9, 17.6, and 15.5 queries for Kontext, BAGEL, and Step1X-Edit, respectively. The total average MLLM calls per case range from 83.0 to 92.0 across different models. This demonstrates that our edit-specific verification strategies introduce limited additional MLLM overhead while achieving significant NFE reduction compared to baseline methods.

{
\setlength{\tabcolsep}{3.5 pt}
\renewcommand{\arraystretch}{1.15} 
\begin{table}[t]
    \caption{\textbf{Average MLLM queries per editing case on GEdit-Bench}. We report the average number of calls for each MLLM component under a sampling budget of $N=32$.}
    \label{supp_tab: mllm_cost}
    \vspace{-0.6em}
    \centering
    \resizebox{1.0\linewidth}{!}
    {
    \scriptsize
\begin{tabular}{l | c | c | c }
\toprule
\textbf{MLLM Component} & \textbf{Kontext} & \textbf{BAGEL} & \textbf{Step1X-Edit} \\
\midrule
General Verifier & 65.1 & 71.4 & 66.1 \\
Region Generation & 1.0 & 1.0 & 1.0 \\
Caption Generation & 1.0 & 1.0 & 1.0 \\
Instance-Specific Questions (Generate) & 1.0 & 1.0 & 1.0 \\
Instance-Specific Questions (Answer) & 14.9 & 17.6 & 15.5 \\
\midrule
\textbf{Total Avg.} & \textbf{83.0} & \textbf{92.0} & \textbf{84.6} \\
\bottomrule 
\end{tabular}
}
\vspace{-0.6 em}
\end{table}
}

\subsection{More Qualitative Results}
\label{supp_sec: qualitative_result}

We present additional qualitative results to demonstrate the effectiveness of ADE-CoT across different editing scenarios. \cref{supp_fig: baselin_w_ADE_CoT} and \cref{supp_fig: baselin_w_ADE_CoT_multi_turn} show that our ADE-CoT significantly improves baseline model performance on complex edits and multi-turn edits. In~\cref{supp_fig: baselin_w_ADE_CoT}, baseline models often fail on challenging tasks such as large pose changes, multi-object modifications, and fine-grained regional edits. Our method successfully handles these cases through adaptive test-time scaling. \cref{supp_fig: baselin_w_ADE_CoT_multi_turn} demonstrates that baseline models struggle to maintain consistency across multiple sequential instructions, leading to accumulated errors in subsequent turns. In contrast, ADE-CoT preserves context from previous edits and produces correct final images that reflect all requested changes. \cref{supp_fig: compare_w_BoN} illustrates the advantage of our instance-specific verifier in final selection. Both baseline and Best-of-N methods fail to detect subtle errors in edited results. Our ADE-CoT generates targeted questions to examine critical details, enabling more accurate identification of correct outputs and producing superior final results compared to Best-of-N selection.

{
\begin{figure*}[htbp]
\begin{center}
\includegraphics[width=0.99\linewidth]{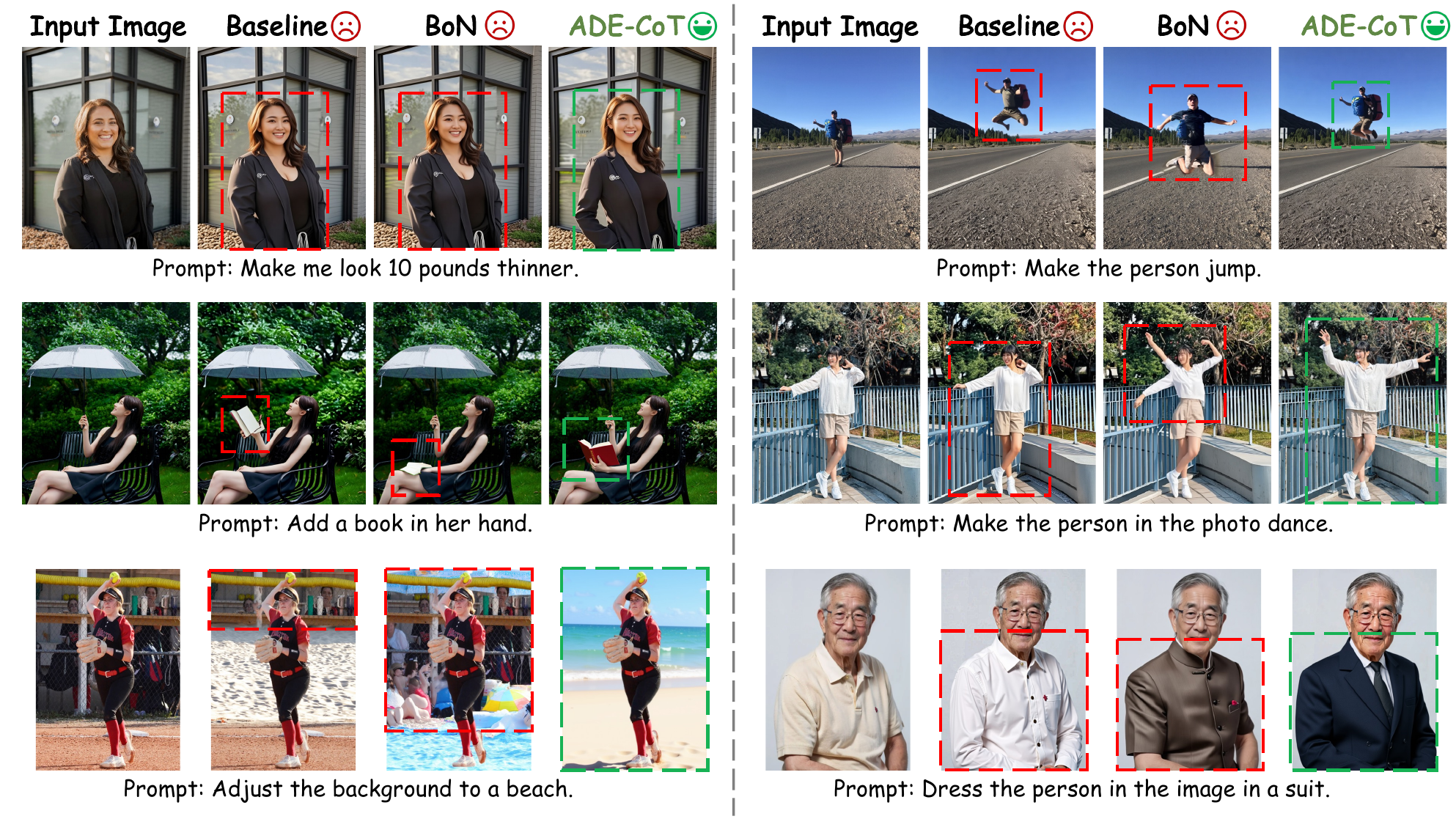}
\end{center}
\vspace{-1.7em}
\caption{\textbf{Instance-specific verifier improves fine-grained selection}. We compare the baseline model, Best-of-N (BoN), and our ADE-CoT across challenging editing scenarios. Baseline models and BoN often produce results with subtle errors (marked in red boxes), such as incorrect pose changes, background adjustments, and clothing modifications. In contrast, our instance-specific verifier generates targeted questions to examine critical details, enabling accurate detection of editing errors and selecting correct results (marked in green boxes). }
\vspace{-0.5em}
\label{supp_fig: compare_w_BoN}
\end{figure*}
}


\subsection{Critical Analysis and Discussion}
\label{supp_sec: critical_analysis}

\noindent \textbf{How much do MLLMs impact our framework?} Our framework, ADE-CoT, relies on MLLMs for key verification steps. It uses MLLMs to identify the edit region, generate a target caption, and compute the general score. Due to hallucinations, these models inevitably introduce incorrect judgments. To verify the impact of MLLM capability on our framework, we conduct experiments with three different MLLMs. We find that ADE-CoT demonstrates strong robustness across varying MLLM capacities. 
\ding{182} As shown in Tab.~5 of the main paper, our method consistently achieves over 2× speedup compared to BoN across all tested MLLMs. 
\ding{183} We observe that even when using a weaker MLLM such as Qwen2.5-VL-72B, replacing specific components with stronger MLLMs leads to consistent improvements. 
\cref{supp_tab: ablation_mllm_on_region} shows that replacing the MLLM for region localization improves edited-region correctness and overall performance. 
Similarly, \cref{supp_tab: ablation_mllm_on_caption} demonstrates that using a stronger MLLM for caption generation enhances instruction-caption consistency. These results show that more accurate predictions from better MLLMs lead to consistent gains in both performance and efficiency.

\noindent \textbf{Can Image-CoT improve all editing cases?} 
As illustrated in~\cref{supp_fig: high_redundancy}, we find that some samples still receive low scores even after applying Image-CoT. These cases typically represent scenarios where the model inherently lacks editing capability.  Even after extensive sampling or prompt modification, their performance shows minimal improvement. It also demonstrates that Image-CoT can serve as a diagnostic method to identify model capability boundaries. By incorporating these cases into training data, we can further enhance the capabilities of existing models.

{
\setlength{\tabcolsep}{3.5 pt}
\renewcommand{\arraystretch}{1.15} 
\begin{table}[t]
    \caption{\textbf{Effect of MLLMs on edited-region correctness}. We compare three MLLMs for edited-region localization (via prompt $P_{\text{reg}}$), while keeping Qwen-VL-72B for other components.}
    \label{supp_tab: ablation_mllm_on_region}
    \vspace{-0.6em}
    \centering
    \resizebox{1.0\linewidth}{!}
    {
    \scriptsize
\begin{tabular}{l | c | cc | cc | cc }
\toprule
\multirow{2}{*}{\textbf{Model}} & \multirow{2}{*}{\textbf{MLLM for Region}} & \multicolumn{2}{c|}{\textbf{Kontext}} & \multicolumn{2}{c|}{\textbf{BAGEL}} & \multicolumn{2}{c}{\textbf{Step1X-Edit}}  \\ 
\cmidrule(lr){3-4} \cmidrule(lr){5-6} \cmidrule(lr){7-8}
 & & G\_O$\uparrow$ & NFE$\downarrow$ & G\_O$\uparrow$ & NFE$\downarrow$ & G\_O$\uparrow$ & NFE$\downarrow$ \\ 
\midrule 
BoN & - & 6.641 & 896 & 6.908 & 1600 & 7.157 & 896 \\
\midrule
\multirow{3}{*}{\textbf{ADE-CoT}}  & Qwen2.5-VL-72B~\cite{qwen25vl} & {{6.637}} & {{436}} & {7.042} & {897} & {7.193} & {446} \\
 & Qwen-VL-MAX~\cite{qwen25vl} & 6.661 & 428 & 6.993 & 901 & 7.194 & 445  \\

 & Qwen3-VL-32B~\cite{qwen3} & \textbf{6.673} & \textbf{424} & \textbf{7.048} & \textbf{869} & \textbf{7.198} &  \textbf{436} \\

\bottomrule 
\end{tabular}
}
\vspace{-0.6 em}
\end{table}
}

{
\setlength{\tabcolsep}{3.5 pt}
\renewcommand{\arraystretch}{1.15} 
\begin{table}[t]
    \caption{\textbf{Effect of MLLMs on instruction-caption consistency}. We compare three MLLMs for instruction-caption consistency (via prompt $P_{\text{cap}}$), while keeping Qwen-VL-72B for others.}
    \label{supp_tab: ablation_mllm_on_caption}
    \vspace{-0.6em}
    \centering
    \resizebox{1.0\linewidth}{!}
    {
    \scriptsize
\begin{tabular}{l | c | cc | cc | cc }
\toprule
\multirow{2}{*}{\textbf{Model}} & \multirow{2}{*}{\textbf{MLLM for Caption}} & \multicolumn{2}{c|}{\textbf{Kontext}} & \multicolumn{2}{c|}{\textbf{BAGEL}} & \multicolumn{2}{c}{\textbf{Step1X-Edit}}  \\ 
\cmidrule(lr){3-4} \cmidrule(lr){5-6} \cmidrule(lr){7-8}
 & & G\_O$\uparrow$ & NFE$\downarrow$ & G\_O$\uparrow$ & NFE$\downarrow$ & G\_O$\uparrow$ & NFE$\downarrow$ \\ 
\midrule 
BoN & - & 6.641 & 896 & 6.908 & 1600 & 7.157 & 896 \\
\midrule
\multirow{3}{*}{\textbf{ADE-CoT}}  & Qwen2.5-VL-72B~\cite{qwen25vl} & {{6.637}} & {{436}} & {7.042} & {897} & {7.193} & {446} \\
 & Qwen-VL-MAX~\cite{qwen25vl} & 6.651 & \textbf{419}  & 7.021 & 899 & 7.186 & 450  \\

 & Qwen3-VL-32B~\cite{qwen3} & \textbf{6.664} & 423 & \textbf{7.052} & \textbf{883} & \textbf{7.195} & \textbf{440} \\

\bottomrule 
\end{tabular}
}
\vspace{-0.6 em}
\end{table}
}

\section{Limitations and Future Work}
\label{supp_sec: limitation_and_future}

\noindent \textbf{Limitations}. Despite achieving superior trade-offs between performance and efficiency, our method suffers from the following limitations: 
\ding{182} \textbf{MLLM computational overhead}.
Our verification relies on large-scale MLLMs (\eg, Qwen-VL 72B), which increases inference latency and limits in resource-constrained scenarios.
\ding{183} \textbf{Hallucination in verification}. 
MLLMs may generate hallucinations during the verification process. This can compromise the accuracy of generated captions, region masks, and instance-specific questions, leading to incorrect quality assessments. While our experiments demonstrate that different MLLMs show minimal variance when ranking multiple candidates (Sec.~4.3, Tab.~5), accurately determining whether the edited image fully satisfies user intent remains challenging.

\noindent \textbf{Future work}. 
We identify two primary directions for future research based on these limitations. 
\ding{182} \textbf{Efficient and accurate verification models}. A key direction is to utilize smaller, specialized models for evaluation. Lightweight models (\eg, 7B parameters) could be trained to provide fast and accurate assessments of edited images. Additionally, an accurate evaluation model would be particularly effective for evaluating intermediate preview images during the denoising process. This could enhance the proposed edit-specific verification and opportunistic stopping strategies, further improving overall efficiency.
\ding{183} \textbf{Broader applications}. 
Our ADE-CoT framework could be extended to other goal-directed generation tasks. 1
Its core strategies of difficulty-aware resource allocation and opportunistic stopping are applicable to domains like video editing and multi-turn conversational generation. These strategies may also be beneficial for standard text-to-image and text-to-video generation, enabling a more efficient Image-CoT process.
 
\section{Extended Related Work}
\label{supp_sec: extended_related_work}

\textbf{Test-time scaling in image generation}. 
Recent years have seen rapid development in generative models~\cite{tang2024crs, tang2025aerogen, Wang_2024, wang2025jasmine, wang2025editor, Bagel_2025_arxiv, FLUX_Kontext_arxiv2025, Step1X_Edit_arxiv2025, qwen_image_arxiv2025, li2025ld, xu2025scalar, wang2026geometry, jin2026semanticcontextmattersimproving, lan2025flux, chen2025s, mao2025omni, yu2025frequency}.
Test-time scaling, as a training-free approach, aims to improve generation quality by extending the inference time. 
Early work on diffusion-based models investigates this by scaling the number of denoising steps~\cite{TTS_add_steps_1, TTS_add_steps_2, TTS_add_steps_3, TTS_add_steps_efficiency_1, TTS_add_steps_efficiency_2, TTS_add_steps_efficiency_3}. 
More recently, following the success of Chain-of-Thought (CoT)~\cite{NLP_CoT, NLP_CoT_ORM, NLP_CoT_PRM_1, NLP_CoT_PRM_2, NLP_TTS_SETS, NLP_TTS_MC, bai2025univg, NLP_TTS_MCTS, ProAPO, yuan2025autodrive, chen2025finger, li2025next, yuan2025image, yuan2025video}, Image-CoT has emerged as a promising paradigm. 
The standard Image-CoT method is noise scaling~\cite{TTS_Baseline}, which generates multiple candidates by perturbing the noise and selects the best image as the final output. 
While effective, its computational cost scales linearly with the number of candidates. Subsequent work aims to generate higher-quality candidates within a fixed budget. Some methods enhance candidate diversity through prompt-level intervention, which enhances prompts by rewriting~\cite{prompt_TTS_Yoonjin, T2IR1} or reflective updates for iterative refinement~\cite{Reflection_DiT, GenRef_CoT_ICCV2025, Image_CoT}. 
Another line of work adapts search algorithms to the diffusion process. 
These methods treat the reverse diffusion chain as a search trajectory~\cite{TTS_Baseline, classical_search_TTS, Tree_Sample_TTS, Video_TTS_EVO, TTS_search_ICML}.
They change the noise based on verifier scores to select promising sampling directions.
Recent methods~\cite{Video_TTS, Image_CoT, wu2025imagerysearch, UniGen_TTS, Video_TTS_beam_search, ICEdit_arxiv2025, TTS_SANA} utilize MLLMs as a verifier to prune low-potential trajectories at early denoising stages. However, most Image-CoT methods focus on text-to-image generation. They are inefficient for image editing due to three key issues: (1) fixed sampling budgets waste computation on simple edits, (2) general MLLM scores cause misjudgement during early pruning, and (3) large-scale sampling produces redundant correct outputs. To address these challenges, we propose ADE-CoT, an edit-specific test-time scaling method that incorporates difficulty-aware resource allocation, edit-specific verification, and depth-first opportunistic stopping.


\noindent \textbf{Verifiers for image editing} can be divided into two primary approaches. 
The first approach comprises metrics requiring ground-truth edited images and corresponding output captions. These methods typically utilize CLIP Score~\cite{CLIPScore} for image-text alignment, CLIP~\cite{CLIP} and DINO~\cite{DINO} similarity for image-image comparison, and L1/L2 distance for pixel-level similarity. However, such metrics are difficult to apply in test-time scaling (TTS) scenarios, as ground-truth data are unavailable during inference. 
The second approach leverages MLLMs as verifiers. Existing methods, such as VIEScore~\cite{VIE_Score} and HQScore~\cite{fine_tune_dataset_hq_edit}, use general verifiers that use instance-agnostic prompts to evaluate aesthetic quality and semantic consistency for each instance. However, general verifiers face two limitations in editing Image-CoT. In early denoising stages, they may incorrectly pruning high-potential candidates that exhibit low preview quality. In the final selection, they struggle to distinguish subtle differences between candidates, assigning identical high scores to images with minor errors. To address these limitations, we introduce two complementary verification strategies. For early-stage misjudgement, we propose edit-specific verification that queries MLLMs to generate ground-truth captions and expected edit regions, enabling more accurate assessment of caption consistency and region localization. For final-stage selection, we introduce instance-specific verification that first generates targeted \textit{yes-no} questions about specific editing aspects, then provides answers based on these questions. This two-stage inquiry guides MLLMs to notice critical details and provides the decision for opportunistic stopping, improving fine-grained selection accuracy, and reducing redundant outputs.


\end{document}